%% file: main.tex
\documentclass[10pt,twocolumn,letterpaper]{article}
\input{preamble}
\title{\textbf{From Augmentation to Synthetic Evidence: Evaluating Generated Chest Radiographs at Dataset Intake}}
\author{Shaikh Akib Shahriyar \quad Zinedine Sehili \quad Adam Arthur \quad Christopher Schwartz\\[0.4em]
\normalsize Rochester Institute of Technology\\
\normalsize Rochester, NY, USA\\
\small\texttt{as8751@rit.edu} \quad \texttt{zs6496@rit.edu} \quad \texttt{aia5138@rit.edu} \quad \texttt{ccsics@rit.edu}}
\date{}
\begin{document}
\maketitle
\input{sections/paper}
\clearpage
{\small\input{main.bbl}}
\clearpage
\setcounter{section}{0}
\setcounter{table}{0}
\setcounter{figure}{0}
\setcounter{equation}{0}
\renewcommand{\thesection}{S\arabic{section}}
\renewcommand{\thetable}{S\arabic{table}}
\renewcommand{\thefigure}{S\arabic{figure}}
\renewcommand{\theequation}{S\arabic{equation}}
\renewcommand{\theHsection}{supp.\arabic{section}}
\renewcommand{\theHtable}{supp.\arabic{table}}
\renewcommand{\theHfigure}{supp.\arabic{figure}}
\renewcommand{\theHequation}{supp.\arabic{equation}}
\twocolumn[\begin{center}\Large\bfseries Supplementary Material\end{center}\vspace{1em}]
\input{sections/supplement}
\end{document}

%% file: preamble.tex
\usepackage[T1]{fontenc}
\usepackage{times}
\usepackage[letterpaper,textwidth=6.875in,textheight=8.875in,top=1in]{geometry}
\usepackage{graphicx}
\usepackage{amsmath,amssymb}
\usepackage{booktabs}
\usepackage{xcolor}
\usepackage[numbers,sort&compress]{natbib}
\usepackage[format=plain,labelformat=simple,labelsep=period,font=small]{caption}
\usepackage[hyphens]{url}
\usepackage{tikz}
\usetikzlibrary{arrows.meta,positioning,fit,backgrounds}
\usepackage{placeins}
\usepackage{listings}
\usepackage[breaklinks,colorlinks,allcolors=blue]{hyperref}
\hypersetup{pdftitle={\textbf{From Augmentation to Synthetic Evidence: Evaluating Generated Chest Radiographs at Dataset Intake}},pdfauthor={Shaikh Akib Shahriyar, Zinedine Sehili, Adam Arthur, Christopher Schwartz}}

%% file: sections/paper.tex
\begin{abstract}
While generated chest radiographs can serve as valuable data augmentation when explicitly declared, the exact same pixels become unsupported evidence when submitted as records of real clinical examinations. We study this dataset-intake vulnerability by disentangling three distinct verification questions: (1) whether an image passes an automated technical quality check, (2) whether an origin classifier flags it as synthetic, and (3) whether a trusted provenance record supports the claimed clinical acquisition. In a controlled intake study, a submitter uses Qwen3.8-27B to select up to 20 candidate radiographs from pools of 1,000 Cheff-identified or SynthCheX images, while a receiver evaluates image quality via MedGemma 1.5 and predicts origin using a linear probe over frozen DINOv3 features. Although quality-guided selection increases MedGemma quality passes from 12/20 to 19/20 for Cheff-identified images and from 8/20 to 17/20 for SynthCheX, we find no evidence that candidate selection increases passage through both checks simultaneously: zero selected Cheff-identified images pass both gates, and uniform, quality-guided, and surrogate-guided selection yield 2, 1, and 2 SynthCheX passes out of 20, respectively. On unprocessed test cohorts, the origin classifier rejects 4,677 of 5,000 SynthCheX images at a nominal 5\% false-alarm rate (flagging 110 of 2,297 genuine NIH radiographs), yet leaves 152 SynthCheX images that pass both pixel-level checks. Moreover, pixel-level screening is fragile to routine distribution shifts: JPEG compression reduces SynthCheX-versus-NIH AUROC from 0.987 to 0.841, and applying the NIH-calibrated threshold to an external genuine cohort (Open-I) triggers false alarms on 568 of 957 authentic radiographs (59.4\%). Because pixel-level checks can filter submissions but cannot verify clinical acquisition claims, we argue that dataset intake pipelines must record and verify source provenance and intended use independently of image pixels.
\end{abstract}

\section{Introduction}
Modern generative models can synthesize high-resolution chest radiographs with striking anatomical fidelity~\cite{weber2023cascaded,dutt2026chexgenbench}. When explicitly declared as \textit{synthetic}, for instance, to rebalance underrepresented disease classes during model training, such images provide legitimate utility. However, they pose a severe evidentiary risk when submitted to medical research repositories as records of actual patient examinations. An automated dataset-intake pipeline may screen incoming radiographs along two pixel-level axes: whether the image satisfies technical quality standards, and whether a forensic classifier flags it as synthetic. Crucially, neither check establishes that a physical clinical examination ever took place. If a generated radiograph passes both pixel checks and enters a repository without provenance metadata, subsequent downstream users may treat synthetic pixels as empirical clinical evidence.

\begin{figure}[t]
\centering
\includegraphics[width=\columnwidth]{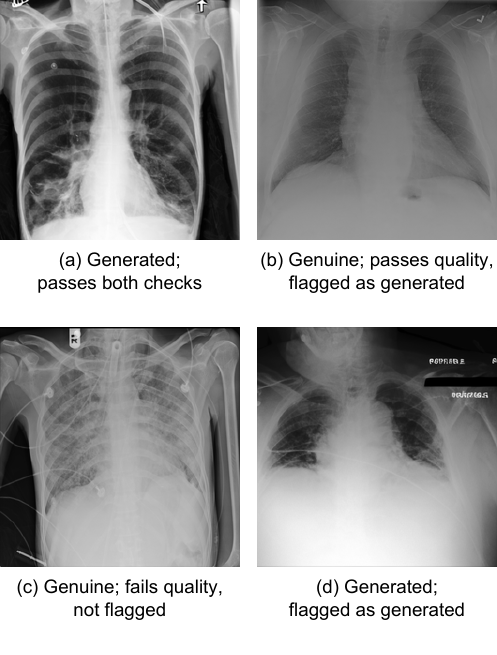}
\caption{Representative test radiographs for each joint outcome of the quality and origin checks, selected post hoc via a deterministic hash rule. Images (a) and (d) are synthetic (SynthCheX), whereas (b) and (c) are genuine (NIH ChestX-ray14). Although Image (a) passes both automated pixel checks, its pixels provide no evidentiary warrant that a real clinical examination occurred.}
\label{fig:main-deterministic-atlas}
\end{figure}

We define \emph{synthetic evidence} as a generated artifact offered to support a factual claim of real-world acquisition without the evidentiary warrant that claim requires, extending recent formulations of ground-truth drift and evidentiary grounding~\cite{arthurschwartz2026groundtruth,schwartz2026warrant}. Under this definition, whether an image constitutes synthetic evidence is a property of the tuple comprising the pixels, the provenance claim, and the declared use—not a property of the pixels alone. Figure~\ref{fig:main-deterministic-atlas}(a) illustrates this distinction using a test case from our study: a generated SynthCheX radiograph that passes our automated quality check and evades our origin classifier. Submitted as declared training augmentation, these pixels are benign; submitted as a clinical acquisition record, the same pixels represent unsubstantiated synthetic evidence.

Maintaining evidentiary integrity at dataset intake therefore requires separating three questions that pixel-only screening conflates:
\begin{itemize}
\item \textbf{Quality:} Does the image satisfy the receiving repository's technical adequacy criteria?
\item \textbf{Predicted origin:} Does a statistical classifier score the pixels as genuine or generated?
\item \textbf{Claim support:} Does an independent, verifiable record substantiate the claimed clinical acquisition and permit the declared use?
\end{itemize}
Satisfying the first two criteria leaves the third entirely unresolved.

Prior work has benchmarked the visual fidelity of synthetic radiographs, trained detectors for generated or locally edited medical imagery, and documented detector sensitivity to compression and resizing artifacts (Section~\ref{sec:related}). Here, we ask how these failure modes interact in an end-to-end dataset intake pipeline, including when a submitter actively curates generated candidates prior to submission. In a controlled laboratory evaluation, a submitter uses Qwen3.8-27B to select up to $K=20$ radiographs from pools of existing synthetic candidates, while a receiver enforces a MedGemma quality gate, a DINOv3-based origin classifier, and a separate provenance registry recording source and intended use. Our experiments reveal three principal findings:
\begin{itemize}
\item \textbf{Candidate selection boosts quality passage but not joint evasion.} Filtering candidates via Qwen's quality rubric substantially increases the fraction of submissions that pass MedGemma's quality gate, but provides no measurable advantage in passing both the quality and origin checks simultaneously.
\item \textbf{Routine image processing induces both threshold drift and ranking loss.} Post-processing operations such as JPEG compression and downsampling degrade origin screening through two distinct mechanisms: \emph{threshold drift}, where global score shifts invalidate a fixed decision threshold, and \emph{ranking loss}, where genuine and generated score distributions overlap even after condition-matched recalibration.
\item \textbf{Cross-collection shift breaks calibrated false-alarm guarantees.} Holding the NIH-calibrated 5\% false-alarm threshold fixed causes the primary origin classifier to falsely flag 568 of 957 genuine Open-I radiographs (59.4\%).
\end{itemize}
In summary, our contributions are: (1)~a formal dataset-intake evaluation framework that decouples technical quality, predicted origin, and acquisition-claim support; (2)~an empirical characterization of how candidate selection, post-processing, and genuine domain shift affect automated quality and origin gates; and (3)~a controlled registry demonstration showing how tracking source verification and intended use separately from pixel checks prevents synthetic evidence admission.

\section{Related Work}
\label{sec:related}
\paragraph{Synthetic chest radiographs.}
Cascaded latent diffusion models such as Cheff generate high-resolution thoracic radiographs~\cite{weber2023cascaded}, while CheXGenBench provides a unified benchmark evaluating synthetic fidelity, patient privacy, and downstream training utility across generators, including the SynthCheX cohort audited here~\cite{dutt2025chexgenbenchv2,dutt2026chexgenbench,synthchex2026card}. Whereas recent reader studies evaluate whether human radiologists can visually distinguish synthetic radiographs from clinical acquisitions~\cite{wang2026detectability,tordjman2026deepfake}, we evaluate automated dataset-intake screening with decoupled quality, origin, and provenance gates.

\paragraph{Medical forgery detection and shortcut learning.}
Adversarial tampering in medical imaging ranges from localized lesion injection or removal in CT volumes~\cite{mirsky2019ctgan} to full-image synthesis benchmarked in DSKI/MedForensics~\cite{li2025medforensics} and localized editing in MedForge~\cite{chen2026medforge}. Notably, MedForge-RSI documents high false-rejection rates on authentic clinical images, brittle cross-generator transfer, and vulnerability to transmission distortions~\cite{chen2026medforgersi}. In natural imagery, forensic classifiers frequently latch onto compression and resolution shortcuts~\cite{grommelt2025fakejpeg} or high-frequency resampling sensitivities~\cite{choi2025warpad}. Rather than proposing a new detector architecture, we audit how these statistical vulnerabilities interact with upstream candidate selection and conformal threshold calibration at dataset intake.

\paragraph{Statistical calibration and cryptographic provenance.}
Outlier testing via conformal $p$-values guarantees finite-sample false-alarm control on genuine inputs provided that calibration and test samples are exchangeable and the scoring function is fixed independently of the calibration set~\cite{bates2023conformal}. However, a conformal $p$-value quantifies only how atypical a score is relative to a reference distribution—not whether an image reflects a physical acquisition. Conversely, digital image signatures~\cite{smith1995authentication}, C2PA Content Credentials~\cite{c2pa2026spec}, and W3C PROV lineage graphs~\cite{moreau2013prov} preserve file integrity and custody chains without judging visual quality. Our intake registry (Section~\ref{sec:source}) bridges this gap by pairing pixel screening with an explicit tri-state verification record that separates known provenance from unknown status without requiring novel cryptographic primitives.

\section{Task and Threat Model}
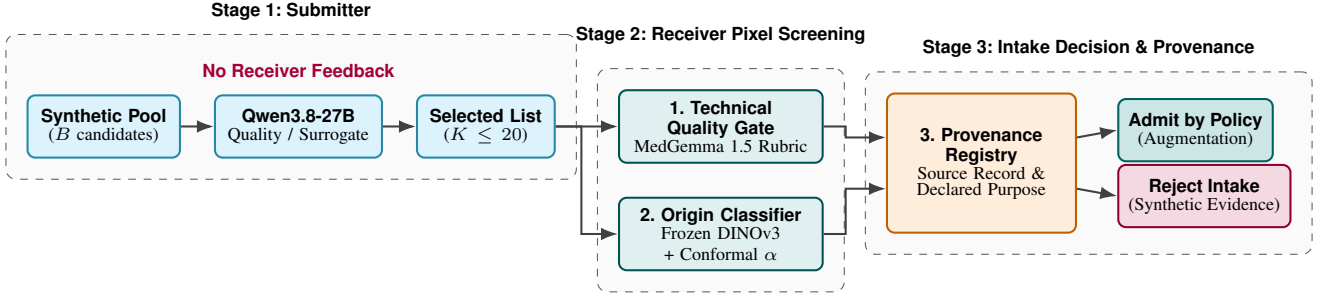
\begin{figure*}[t]
\centering
\resizebox{\textwidth}{!}{
\begin{tikzpicture}[
  font=\sffamily\footnotesize,
  >=Latex,
  stage/.style={draw=black!65, rounded corners=4pt, inner sep=8pt, dashed, fill=black!2},
  box/.style={draw=black!75, rounded corners=3pt, minimum height=0.85cm, align=center, inner sep=4pt, font=\sffamily\scriptsize\bfseries},
  sub/.style={fill=cyan!12, draw=cyan!60!black, thick},
  rec/.style={fill=teal!12, draw=teal!65!black, thick},
  reg/.style={fill=orange!14, draw=orange!75!black, thick},
  outpass/.style={fill=teal!20, draw=teal!75!black, thick, minimum width=2.1cm},
  outfail/.style={fill=purple!15, draw=purple!75!black, thick, minimum width=2.1cm},
  arr/.style={->, thick, draw=black!75}
]

\node[box, sub, text width=1.8cm] (pool) {Synthetic Pool\\[-1pt]{\normalfont\scriptsize ($B$ candidates)}};
\node[box, sub, right=0.45cm of pool, text width=2.0cm] (qwen) {Qwen3.8-27B\\[-1pt]{\normalfont\scriptsize Quality / Surrogate}};
\node[box, sub, right=0.45cm of qwen, text width=1.6cm] (topk) {Selected List\\[-1pt]{\normalfont\scriptsize ($K \le 20$)}};

\draw[arr] (pool) -- (qwen);
\draw[arr] (qwen) -- (topk);

\node[box, rec, right=0.85cm of topk, text width=2.5cm] (qual) {1. Technical Quality Gate\\[-1pt]{\normalfont\scriptsize MedGemma 1.5 Rubric}};
\node[box, rec, below=0.45cm of qual, text width=2.5cm] (orig) {2. Origin Classifier\\[-1pt]{\normalfont\scriptsize Frozen DINOv3 + Conformal $\alpha$}};

\node[box, reg, right=0.85cm of qual, yshift=-0.5cm, text width=2.3cm, minimum height=1.9cm] (registry) {3. Provenance\\Registry\\[-1pt]{\normalfont\scriptsize Source Record \&}\\[-2pt]{\normalfont\scriptsize Declared Purpose}};

\node[box, outpass, right=0.55cm of registry, yshift=0.45cm] (admit) {Admit by Policy\\[-1pt]{\normalfont\scriptsize (Augmentation)}};
\node[box, outfail, right=0.55cm of registry, yshift=-0.45cm] (reject) {Reject Intake\\[-1pt]{\normalfont\scriptsize (Synthetic Evidence)}};

\draw[arr] (topk.east) -- ++(0.35,0) |- (qual.west);
\draw[arr] (topk.east) -- ++(0.35,0) |- (orig.west);

\draw[arr] (qual.east) -- ++(0.3,0) |- ([yshift=0.35cm]registry.west);
\draw[arr] (orig.east) -- ++(0.3,0) |- ([yshift=-0.35cm]registry.west);

\draw[arr] ([yshift=0.35cm]registry.east) -- (admit.west);
\draw[arr] ([yshift=-0.35cm]registry.east) -- (reject.west);

\node[font=\sffamily\scriptsize\bfseries, text=purple!85!black, above=3pt of qwen] (feedback) {No Receiver Feedback};

\begin{scope}[on background layer]
  \node[stage, fit=(pool)(topk)(feedback), label={[font=\sffamily\scriptsize\bfseries, yshift=2pt]above:Stage 1: Submitter}] {};
  \node[stage, fit=(qual)(orig), label={[font=\sffamily\scriptsize\bfseries, yshift=6pt]above:Stage 2: Receiver Pixel Screening}] {};
  \node[stage, fit=(registry)(admit)(reject), label={[font=\sffamily\scriptsize\bfseries, yshift=2pt]above:Stage 3: Intake Decision \& Provenance}] {};
\end{scope}

\end{tikzpicture}
}
\caption{Overview of the dataset-intake evaluation pipeline. Operating without receiver feedback, the submitter uses Qwen3.8-27B to select at most $K$ candidates from a synthetic pool of size $B$. The receiver evaluates technical quality (MedGemma 1.5) and predicted origin (frozen DINOv3 probe with conformal threshold calibration), while an independent registry verifies claim support and intended use.}
\vspace{-3mm}
\label{fig:workflow}
\end{figure*}

We model an automated intake gate for a medical imaging repository (Figure~\ref{fig:workflow}). Each submission comprises an image $x$, a provenance claim asserting how $x$ was acquired, and a declared purpose (e.g., clinical examination record vs.\ synthetic training augmentation). We focus on the critical case where a submitter claims that $x$ records a real clinical acquisition, but provides no verifiable acquisition record, meaning that external claim support is absent by construction (Section~\ref{sec:source}). While dataset source labels (NIH vs.\ Cheff/SynthCheX) serve as our experimental ground truth for origin, they do not constitute clinical acquisition records.

In our threat model, the submitter operates over a fixed candidate pool of $B$ pre-existing synthetic radiographs and selects a subset of at most $K$ images (Section~\ref{sec:selection}) aimed at passing the receiver's intake checks. Crucially, the submitter operates in a strict black-box setting: it relies solely on candidate ratings from Qwen3.8-27B and has zero access to the receiver's MedGemma quality verdicts, DINOv3 origin scores, MedForge-Reasoner outputs, or calibration thresholds. Once the submitter locks its $K$-image submission list, the receiver evaluates each image through two independent pixel gates, a MedGemma technical quality check and a DINOv3-based origin check, followed by a provenance registry audit.

Formally, an image \emph{passes the quality check} if MedGemma judges it technically adequate under a locked rubric (Section~\ref{sec:quality-rule}). The origin check \emph{flags} an image when its predicted origin score exceeds the calibrated decision threshold (Section~\ref{sec:origin-head}); a flagged synthetic image is a \emph{rejection}, whereas a flagged genuine image is a \emph{false alarm}. An image \emph{passes the origin check} if it is unflagged, and \emph{passes both checks} if it satisfies both pixel gates simultaneously. Any input that fails image decoding or elicits an unparseable model response is recorded as \emph{unavailable}, tracked explicitly as a third outcome rather than dropped or counted as a pass or flag. Crucially, passing both pixel checks leaves the clinical acquisition claim unverified until adjudicated by the provenance registry (Section~\ref{sec:source}).

\section{Data and Methods}
\subsection{Datasets and Partitioning}
To prevent data leakage across model fitting, hyperparameter selection, threshold calibration, and final evaluation, we partition all cohorts into four strictly disjoint splits (Table~\ref{tab:splits}): \emph{fit} (trains linear classifier weights), \emph{development} (selects regularization strength and quality rubrics), \emph{calibration} (sets conformal false-alarm thresholds on genuine data), and \emph{test} (reports final held-out metrics).

\begin{table}[t]
\centering
\begin{tabular}{lrrrr}
\toprule
Collection & Fit & Dev. & Cal. & Test\\
\midrule
NIH genuine & 1,500 & 1,000 & 3,000 & 2,297\\
Cheff-identified & 1,500 & 500 & 0 & 1,000\\
SynthCheX & 1,000 & 500 & 0 & 5,000\\
\midrule
Total & 4,000 & 2,000 & 3,000 & 8,297\\
\bottomrule
\end{tabular}
\caption{Sample allocation across strictly disjoint splits. The calibration split contains genuine NIH radiographs exclusively. Primary origin probes are fitted on NIH and Cheff-identified splits, leaving SynthCheX as an unseen generator at test time (SynthCheX fit/dev splits are reserved for secondary transfer ablations).}
\label{tab:splits}
\end{table}

Genuine reference radiographs are drawn from the NIH ChestX-ray14 subset of MaCheX~\cite{wang2017chestxray,machex2026repo}, restricted to one radiograph per patient while respecting upstream train/test partitions. Disjoint fit, development, and calibration cohorts are sampled from the upstream train/validation pool, while the test split comprises all 2,297 upstream test patients remaining after excluding 500 patients used in our exploratory pilot. View projection—anteroposterior (AP) versus posteroanterior (PA)—is logged solely for observational stratification.

We evaluate two synthetic cohorts. The \emph{Cheff-identified} cohort comprises 3,000 legacy radiographs attributed to the Cheff cascaded latent diffusion model~\cite{weber2023cascaded}. Because exact generation seeds, conditioning prompts, and checkpoint hashes are unavailable for this legacy archive, we treat it as an empirical sample of observed synthetic artifacts rather than a characterization of the upstream Cheff sampler; 1,500 images (1,250 pilot plus 250 additional) are allocated to fitting, 500 to development, and 1,000 to testing. The \emph{SynthCheX} cohort is drawn from the fixed 75,649-image SynthCheX-75K-v2 release generated via Sana 0.6B~\cite{dutt2025chexgenbenchv2,synthchex2026card}, from which we randomly sample 6,500 images (1,000 fit, 500 development, and 5,000 test, strictly excluding 500 pilot images).

Finally, to evaluate out-of-distribution false alarms on an external clinical repository, we curate 957 frontal chest radiographs from the Indiana University Open-I collection, grouped by radiology report~\cite{demnerfushman2016indiana}. Open-I is used exclusively at test time under the frozen NIH-calibrated threshold.

\subsection{Model Roles and Backbones}
Each model in our pipeline serves a strictly segregated role:
\begin{itemize}
\item \textbf{Submitter selector (Qwen):} Unquantized \texttt{Qwen/\allowbreak Qwen3.8-\allowbreak 27B} evaluated in BF16~\cite{qwen38card}, used exclusively by the submitter to score and rank candidate pools.
\item \textbf{Receiver quality gate (MedGemma):} Instruction-tuned \texttt{google/\allowbreak medgemma-\allowbreak 1.5-\allowbreak 4b-\allowbreak it}~\cite{sellergren2026medgemma15,medgemma15card}, used by the receiver to assess technical image adequacy.
\item \textbf{Receiver origin backbones (DINOv3 \& MedGemma SigLIP):} Our primary origin classifier probes frozen CLS or mean-patch representations from web-pretrained DINOv3 ViT-7B/16 (\texttt{facebook/\allowbreak dinov3-\allowbreak vit7b16-\allowbreak pretrain-\allowbreak lvd1689m})~\cite{simeoni2025dinov3}. As a domain-specific comparison, our \emph{medical head} probes the mean of the 4,096 patch tokens from MedGemma's internal SigLIP vision encoder.
\item \textbf{External forensic baseline (MedForge-Reasoner):} \texttt{RichardChenZH/\allowbreak MedForge-\allowbreak Reasoner}~\cite{medforge2026checkpoint,chen2026medforgersi}, a Qwen-family vision-language detector trained to identify localized medical image tampering, evaluated zero-shot for categorical real/fake classification.
\end{itemize}

\subsection{Origin Classifier Training and Conformal Calibration}
\label{sec:origin-head}
Our primary origin classifier, termed the \emph{raw CLS head}, is an $L_2$-regularized logistic regression probe trained on frozen DINOv3 CLS embeddings from 1,500 genuine NIH and 1,500 Cheff-identified fit images. We select the inverse regularization parameter $C\in\{0.01,0.1,1,10\}$ that maximizes rejection of Cheff-identified development images at a 5\% false-alarm rate on genuine NIH development images (breaking ties toward smaller $C$). Because SynthCheX is completely excluded from training and model selection, evaluating the raw CLS head on SynthCheX measures zero-shot transfer to an unseen generator architecture. Here, ``raw'' denotes that the probe is trained on canonically preprocessed images without synthetic post-processing augmentations (contrast with the augmented head in Section~\ref{sec:processing-design}).

Once the probe weights are frozen, we calibrate its decision threshold on the disjoint 3,000-image NIH calibration split to target a nominal false-alarm rate $\alpha=0.05$ (with $\alpha\in\{0.01, 0.10\}$ reported in the Supplement). Let $s(x)$ denote the probe's predicted syntheticity score for image $x$, and let $s_1,\ldots,s_n$ denote the scores across the $n=3{,}000$ genuine calibration images. Following~\citet{bates2023conformal}, we compute the conformal outlier $p$-value:
\begin{equation}
p_{\rm real}(x)=\frac{1+\sum_{i=1}^{n}\mathbf{1}\{s_i\geq s(x)\}}{n+1}.
\label{eq:conformal}
\end{equation}
Intuitively, $p_{\rm real}(x)$ represents the rank fraction of genuine calibration radiographs that appear at least as suspicious as $x$ (rather than the posterior probability that $x$ is authentic). The receiver flags $x$ as generated whenever $p_{\rm real}(x)\leq\alpha$, with exact boundary ties left unflagged.

Under standard conformal inference~\cite{bates2023conformal}, if genuine calibration and test radiographs are exchangeable and $s(\cdot)$ is fixed independently of the calibration split, the expected false-alarm rate on genuine test images is bounded by $\alpha$. Crucially, this finite-sample guarantee applies solely to the marginal origin check on in-domain genuine data; it does not bound false rejections under the joint quality-and-origin rule, under post-processing distribution shifts, or across external clinical repositories such as Open-I, nor does it validate an acquisition claim.

For secondary comparisons, we calibrate two auxiliary reference sets: a 2,914-image calibration subset for our exploratory pilot head (excluding 86 images used during pilot development), and a 2,930-image subset for an 8-feature \emph{rubric-field probe} trained on MedGemma's numerical quality subscores (excluding 10 rubric-development images and 60 unavailable responses). We also benchmark a 6-feature pixel-statistics baseline and secondary DINOv3 heads trained on SynthCheX alone or on both generators jointly (Supplement S6).

\subsection{Post-Processing Robustness Protocol}
\label{sec:processing-design}
To disentangle how routine image transformations affect origin screening, we evaluate 13 controlled processing conditions across 1,000 test radiographs per cohort. Our \emph{canonical} preprocessing pipeline converts each input to grayscale luminance, applies aspect-preserving Lanczos resizing to a 256-pixel longest side, zero-pads to a $256\times 256$ square, and replicates across RGB channels without histogram equalization before feeding DINOv3's 224-pixel input transform. We then subject each canonical test image to 12 perturbations: JPEG compression at quality levels $q\in\{95, 85, 75, 50\}$; bilinear downsampling to $\{192, 128, 64\}$ pixels followed by restoration to 256 pixels; direct 512-pixel extraction from native files (\emph{native 512}; note that Cheff-identified archives are stored at $256\times 256$ and thus upscaled); 4-pixel border masking; 4-pixel boundary padding; 64-pixel central occlusion masking; and 26-pixel peripheral border masking.

Across all 13 conditions, we compare three detector configurations:
\begin{enumerate}
\item \textbf{Raw head / canonical threshold:} The primary raw CLS head evaluated using the fixed canonical NIH calibration threshold.
\item \textbf{Raw head / matched threshold:} The same frozen raw CLS head, but with its conformal threshold recalibrated on genuine NIH calibration images subjected to the exact same processing perturbation.
\item \textbf{Augmented head / matched threshold:} A DINOv3 CLS probe trained on five processed views per fit image (Supplement S6) and evaluated under condition-matched calibration thresholds.
\end{enumerate}
Comparing Configurations~(1) and~(2) cleanly separates \emph{threshold drift} (score translation that invalidates a static threshold) from \emph{ranking loss} (degraded separability between genuine and synthetic distributions that persists even after optimal threshold recalibration). Comparing Configurations~(2) and~(3) tests whether multi-condition training recovers ranking separability. Prior to test evaluation, we prespecified an efficacy criterion for the augmented head: achieving at least a $+5.0$ percentage-point (pp) mean increase in Cheff-identified rejection across the 13 conditions while incurring at most $+1.0$\,pp additional genuine NIH false alarms.

\subsection{Technical Quality Check and Rubrics}
\label{sec:quality-rule}
The receiver's quality gate queries MedGemma 1.5 using a structured evaluation rubric selected on development data (the \emph{alternative rubric}; Supplement S5). A radiograph passes the quality gate if and only if MedGemma classifies it as an assessable chest radiograph, assigns an anatomical visibility score $\geq 3$ (on a 0--4 scale), assigns a non-null artifact severity score $\leq 1$ (on a 0--4 scale), and the image passes a deterministic flat-image guard (requiring a pixel dynamic range $>1$ or standard deviation $>0.5$ on $[0,255]$).

To reflect realistic asymmetry between submitter and receiver, the submitter's Qwen selector does not have access to the receiver's rubric. Instead, Qwen evaluates candidates under an independent \emph{original rubric} that scores technical quality $T\in[0,4]$, anatomical plausibility $A\in[0,4]$, and artifact severity $V\in[0,4]$, combining them into a 100-point composite score:
\begin{equation}
Q=10T+10A+5(4-V)\in[0,100].
\label{eq:quality-score}
\end{equation}
Consequently, the submitter and receiver differ in both vision-language backbone and scoring rubric.

\subsection{Candidate Selection Strategies}
\label{sec:selection}
Our selection experiment tests whether a submitter who filters a large candidate pool of size $B$ down to $K=20$ submissions can increase the rate at which synthetic radiographs pass both intake gates. We evaluate nested candidate pools of size $B\in\{100, 250, 500, 1{,}000\}$ for both synthetic cohorts, plus $B=5{,}000$ for SynthCheX, comparing three Qwen-driven selection policies:
\begin{itemize}
\item \textbf{Uniform selection:} Randomly samples $K$ candidates without replacement.
\item \textbf{Quality selection:} Selects the top $K$ candidates ranked by Qwen's composite quality score $Q$ (placing unavailable responses last).
\item \textbf{Quality-constrained surrogate selection:} Queries Qwen for an additional 0--10 \emph{surrogate authenticity rating} estimating how closely a candidate resembles a genuine clinical radiograph. This policy filters for candidates satisfying $Q\geq 75$ with a valid surrogate rating, ranking survivors primarily by surrogate authenticity and secondarily by $Q$ (submitting fewer than $K$ images if fewer qualify).
\end{itemize}
A secondary tournament-style pairwise tie-breaking policy is evaluated at $B=100$ in Supplement S7.

\paragraph{Statistical reporting.}
All pass, rejection, and flag proportions retain the full attempted cohort as the denominator so that unavailable model responses are never silently omitted, accompanied by Wilson 95\% confidence intervals. Paired cross-condition and cross-strategy contrasts use 2,000 cluster-bootstrap resamples at the base-image level (clustering all processed views or pool draws of a given image) conditioned on the frozen models and thresholds (Supplement S9).

\section{Results}
\subsection{Quality-Guided Selection Boosts Quality Passage}
\label{sec:quality-results}
When selecting $K=20$ submissions from a pool of $B=1{,}000$ candidates, Qwen-guided quality selection increases receiver quality-gate passage from 12/20 (uniform baseline) to 19/20 for Cheff-identified radiographs, and from 8/20 to 17/20 for SynthCheX. Despite differences in both backbone architecture and prompt rubric, Qwen rankings transfer effectively to MedGemma's technical quality gate.

Evaluated across the complete uncurated test splits, MedGemma's quality gate passes 1,717 of 2,297 genuine NIH radiographs (74.7\%; 95\% CI: 72.9--76.5\%), 668 of 1,000 Cheff-identified images (66.8\%; 63.8--69.6\%), and 2,656 of 5,000 SynthCheX images (53.1\%; 51.7--54.5\%), with NIH non-passes including 78 unparseable responses. Notably, genuine NIH quality pass rates vary sharply with acquisition projection: 83.1\% of PA radiographs pass (1,460/1,756) compared to only 47.5\% of portable AP radiographs (257/541), reflecting the higher prevalence of positioning and apparatus artifacts in bedside AP imaging.

\subsection{Held-Out Audit of the Quality Rubric}
Because receiver intake depends directly on the quality rubric, we audited the development-selected alternative rubric against the original rubric on 100 held-out radiographs per cohort subjected to six unseen synthetic degradations (two severities each of Gaussian blur, downsampling, and additive noise), alongside 60 non-radiograph invalid inputs (e.g., uniform fields and synthetic noise).

On the 60 non-radiograph inputs, the alternative rubric rejects all 60 even without the deterministic flat-image guard, whereas the original rubric erroneously passes 5; enabling the flat-image guard reduces false passes to 0/60 for both rubrics (95\% CI: 0.0--6.02\%). On clinical and synthetic radiographs, however, the alternative rubric exhibits clear trade-offs: it fails to return a parseable response on 3/100 clean NIH images (vs.\ 0/100 for the original rubric) and produces more unavailable outputs across degraded pairs (41 vs.\ 2). While the alternative rubric ranks the clean image strictly above its degraded counterpart in 650 of 1,800 pairs across cohorts compared to 582 of 1,800 for the original rubric, a $+3.8$\,pp difference whose 95\% CI $[-1.2, +8.8]$ spans zero, it also exhibits more preference reversals where the degraded copy scores higher (426 vs.\ 228), and underperforms on SynthCheX pairs specifically (142/600 vs.\ 209/600). In adherence to our pre-registered protocol, we retain the frozen alternative rubric for all primary intake experiments while noting these mixed psychometric trade-offs.

\subsection{Origin Screening on Canonical Images}
\label{sec:origin-results}
On canonically preprocessed test images, the primary DINOv3 raw CLS head achieves strong separation while respecting its nominal 5\% false-alarm calibration (Table~\ref{tab:origin}): it flags 110 of 2,297 genuine NIH radiographs (4.8\%), rejects 1,000 of 1,000 in-domain Cheff-identified images (100.0\%), and rejects 4,677 of 5,000 unseen SynthCheX images (93.5\%). Probing MedGemma's domain-specific SigLIP encoder (\emph{medical head}) yields a slightly higher full-curve SynthCheX AUROC (0.9891 vs.\ 0.9849) and nearly identical rejection at the 5\% threshold (4,673/5,000). By contrast, the 8-feature rubric-field probe performs near chance (rejecting only 1.3\% of Cheff and 2.0\% of SynthCheX), whereas a simple 6-feature pixel-statistics baseline rejects 60.3\% of Cheff and 55.6\% of SynthCheX, confirming that low-level intensity statistics already carry substantial collection-identifying signal.

\begin{table*}[t]
\centering
\resizebox{\textwidth}{!}{
\begin{tabular}{lcccc}
\toprule
Origin classifier & NIH flagged (False Alarm) & Cheff rejected & SynthCheX rejected & Unavailable \\
\midrule
DINOv3 CLS head, raw (primary) & 110 (4.0--5.7) & 1,000 (99.6--100.0) & 4,677 (92.8--94.2) & 0/0/0 \\
Medical head, raw & 107 (3.9--5.6) & 1,000 (99.6--100.0) & 4,673 (92.7--94.1) & 0/0/0 \\
DINOv3 CLS head, augmented & 97 (3.5--5.1) & 1,000 (99.6--100.0) & 4,644 (92.1--93.6) & 0/0/0 \\
Rubric-field probe & 22 (0.6--1.4) & 13 (0.8--2.2) & 99 (1.6--2.4) & 78/12/80 \\
Pixel-statistics classifier & 50 (1.7--2.9) & 603 (57.2--63.3) & 2,782 (54.3--57.0) & 0/0/0 \\
\bottomrule
\end{tabular}
}
\caption{Origin-check performance on canonical test radiographs at the nominal $\alpha=5\%$ false-alarm threshold. Entries report raw counts over all attempted test images (2,297 NIH, 1,000 Cheff-identified, 5,000 SynthCheX) with Wilson 95\% confidence intervals in parentheses (in \%). Thresholds are calibrated on 3,000 genuine NIH images (2,930 valid responses for the rubric-field probe). The final column tracks unavailable outputs (NIH/Cheff/SynthCheX).}
\label{tab:origin}
\end{table*}

Training distribution strongly governs cross-generator transfer. A secondary DINOv3 CLS head trained exclusively on non-pilot Cheff-identified images allows 311 of 5,000 SynthCheX images to pass unflagged (flagging 101/2,297 NIH), whereas a head trained exclusively on SynthCheX allows 114 of 1,000 Cheff-identified images to pass (flagging 165/2,297 NIH). Training jointly on both generators rejects all 6,000 synthetic test images (flagging 134/2,297 NIH), but represents an in-distribution evaluation rather than unseen-generator transfer (Supplement S10).

Enforcing the quality and origin gates jointly highlights the compound attrition on authentic clinical data alongside residual synthetic leakage: across the full test splits, the combined rule admits 1,635 of 2,297 genuine NIH radiographs (excluding 662 genuine images, or 28.8\%, across quality failures, unavailable replies, and origin false alarms), 0 of 1,000 Cheff-identified images, and 152 of 5,000 SynthCheX images (3.0\%). Those 152 SynthCheX radiographs pass both pixel checks without any clinical acquisition record (Supplement S10).

\subsection{Candidate Selection Does Not Improve Joint Evasion}
\label{sec:selection-results}
\begin{figure*}[t]
\centering
\includegraphics[width=\textwidth]{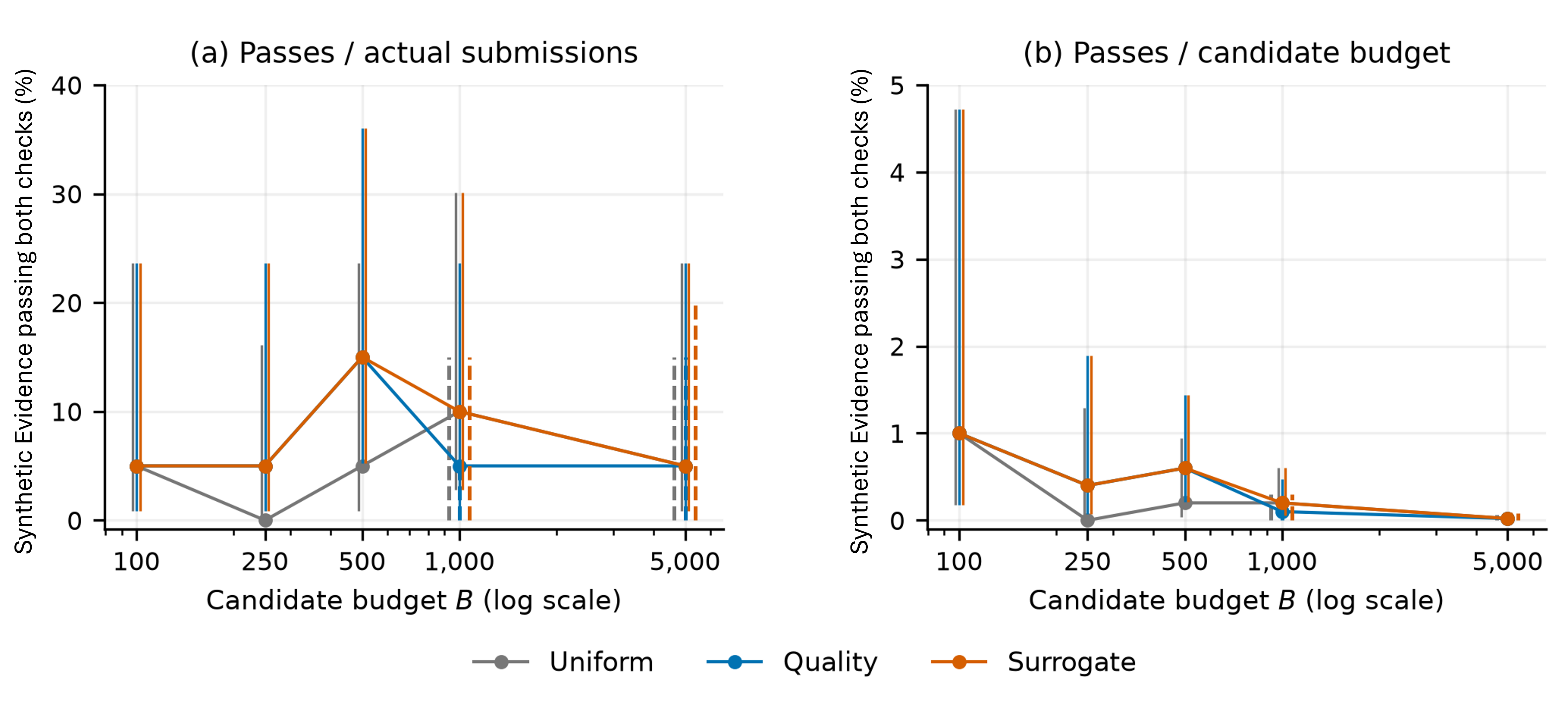}
\caption{Joint passage of SynthCheX candidates through both quality and origin gates at submission limit $K=20$: (a) pass rate per submitted radiograph and (b) yield per candidate in the pool of size $B$. Solid error bars denote exact Wilson 95\% intervals on the fixed evaluation lists; dashed intervals reflect 2,000 bootstrap resamples of the selection process ($B\in\{1{,}000, 5{,}000\}$).}
\label{fig:selection}
\end{figure*}

Although Qwen selection reliably increases MedGemma quality passage (Section~\ref{sec:quality-results}), we find no evidence that it increases joint passage through both intake gates (Figure~\ref{fig:selection}). For Cheff-identified candidates at $B=1{,}000$ and $K=20$, all three strategies yield 0/20 joint passes (95\% upper bound: 16.1\%), a direct consequence of the primary origin classifier rejecting 100\% of the Cheff-identified test pool (Table~\ref{tab:origin}). For SynthCheX at $B=1{,}000$, uniform, quality-guided, and surrogate-guided selection yield 2/20, 1/20, and 2/20 joint passes, respectively. Relative to uniform random submission, quality-constrained surrogate selection yields zero net gain; relative to pure quality selection, its $+5.0$\,pp difference per submitted image ($95\%$ bootstrap CI: $[-5.0, +10.0]$) and $+0.001$ yield per pool candidate ($[-0.001, +0.002]$) are statistically indistinguishable from zero. Expanding the pool to $B=5{,}000$ yields 1/20 joint passes under both quality and surrogate selection. Because Qwen's visual realism judgments do not anti-correlate with DINOv3's high-frequency origin features, black-box visual curation fails to amplify synthetic leakage.

\subsection{Impact of Post-Processing: Threshold Drift vs.\ Ranking Loss}
\begin{figure*}[t]
\centering
\includegraphics[width=0.85\textwidth]{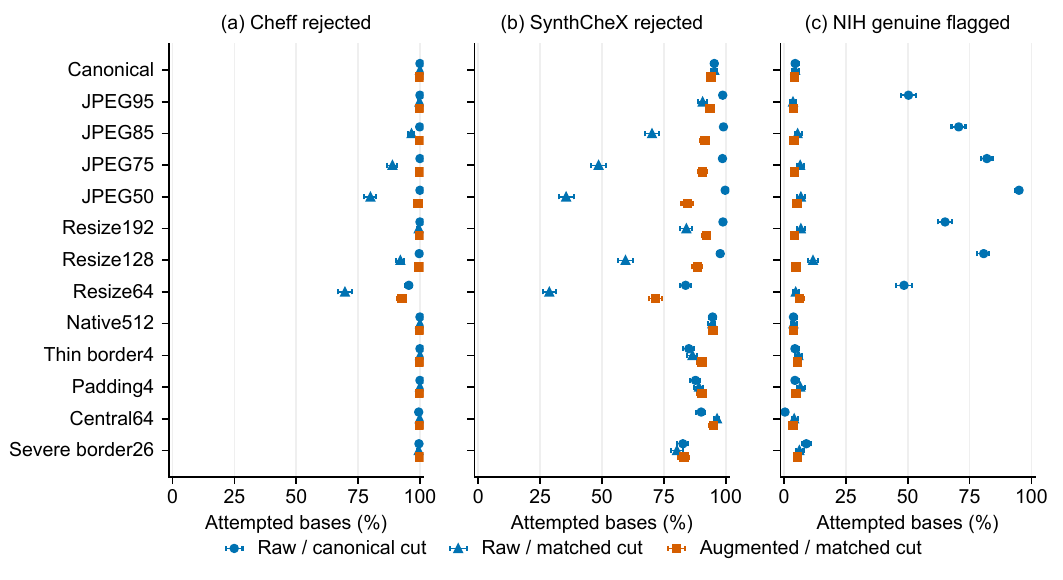}
\caption{Percentage of the 1,000 test radiographs per cohort rejected (a, b) or falsely flagged (c) across 13 processing conditions at nominal $\alpha=5\%$. ``Cut'' denotes the conformal decision threshold: \emph{Raw/canonical} uses the static canonical threshold; \emph{Raw/matched} recalibrates the threshold per condition; \emph{Augmented/matched} pairs multi-condition training with matched recalibration.}
\vspace{-3mm}
\label{fig:processing}
\end{figure*}

Evaluating the 13 post-processing conditions confirms that routine image transformations induce both threshold drift and ranking loss (Figure~\ref{fig:processing}). Under JPEG compression at quality 75, applying the static canonical threshold (\emph{Raw/canonical}) causes the raw CLS head to reject 987 of 1,000 SynthCheX images, but only because global score translation simultaneously triggers false alarms on 820 of 1,000 genuine NIH radiographs (82.0\%). Even when threshold drift is removed altogether by evaluating threshold-free separability, substantial ranking loss remains: SynthCheX-versus-NIH AUROC drops from 0.987 under canonical preprocessing to 0.841 under JPEG 75 and 0.828 under 128-pixel downsampling.

Condition-matched recalibration (\emph{Raw/matched}) corrects threshold drift on genuine images at the expense of synthetic detection recall. Under JPEG 75, recalibrating the threshold restores the NIH false-alarm rate to 65/1,000 (6.5\%), but reduces SynthCheX rejection from 987/1,000 to 486/1,000 (48.6\%). Similarly, under 128-pixel downsampling, matched recalibration lowers NIH false alarms from 807/1,000 to 117/1,000 while reducing SynthCheX rejections from 978/1,000 to 595/1,000. Averaged equally across all 13 conditions, condition-matched recalibration reduces genuine NIH false alarms by $34.01$\,pp while reducing SynthCheX rejections by $19.45$\,pp (Supplement S10).

Multi-condition training (\emph{Augmented/matched}) substantially mitigates ranking loss on unseen generators while nearly achieving our prespecified Cheff target. Across the 13 conditions under matched thresholds, the augmented head reduces genuine NIH false alarms by $1.32$\,pp relative to the raw head (95\% paired CI: $[-1.96, -0.67]$) and increases Cheff-identified rejection by $+4.97$\,pp ($[+4.45, +5.51]$)—falling just $0.03$\,pp shy of the prespecified $+5.0$\,pp point-estimate target. Crucially, on the unseen SynthCheX generator, multi-condition augmentation improves mean rejection by $+15.45$\,pp ($[+14.47, +16.51]$).

\subsection{Cross-Dataset Generalization and External Baselines}
To test whether conformal false-alarm calibration survives a change of clinical institution, we evaluate the 957 genuine Open-I radiographs using the frozen NIH-calibrated threshold ($\alpha=5\%$). Despite containing zero synthetic images, Open-I triggers false alarms on 568 of 957 genuine radiographs under the primary DINOv3 raw CLS head (59.4\%; 95\% CI: 56.2--62.4\%) and 152 of 957 under the MedGemma medical head (15.9\%; 13.7--18.3\%). Thus, domain shift across genuine hospital collections alone is sufficient to invalidate in-domain false-alarm guarantees.

Similarly, evaluating the external localized-forgery detector MedForge-Reasoner across 300-image subsets per cohort reveals severe cross-task miscalibration: its categorical verdicts falsely label 178 of 300 genuine NIH radiographs as fake (59.3\%; 53.7--64.7\%) while misclassifying 127 of 300 Cheff-identified (42.3\%) and 50 of 300 SynthCheX (16.7\%) full-image syntheses as authentic.

\section{Source Provenance, Intended Use, and Reuse}
\label{sec:source}
Our empirical findings demonstrate a fundamental limitation of pixel-based intake screening: passing quality and origin checks neither certifies clinical acquisition nor prevents legitimate synthetic augmentations from being conflated with clinical records. We therefore pair pixel screening with a lightweight provenance registry that records three decoupled fields for every submission: (1) technical quality status, (2) predicted origin status, and (3) external claim-support status (\emph{verified}, \emph{unsupported}, or \emph{unknown}) paired with declared purpose (\emph{clinical acquisition} vs.\ \emph{declared augmentation}).

We validate this architecture through a controlled policy-replay experiment over 8 base radiographs instantiated across 24 claim-and-purpose variants and 72 policy evaluations (3 intake policies $\times$ 24 variants, holding pixel bytes identical across paired variants). A provenance-aware registry policy immediately rejects all unsupported clinical acquisition claims without relying on brittle pixel classifiers, while admitting declared synthetic augmentations for training use. Conversely, layering pixel-level quality and origin gates atop verified clinical acquisitions reduces accepted genuine examinations from 4/4 to 2/4 due to false rejections. Finally, auditing cross-repository export and import confirms that whereas a permissive importer silently promotes stripped verification fields into trusted records, a hash-verifying intake importer preserves missing or mismatched provenance strictly in the \emph{unknown} state. While our laboratory registry does not implement hospital PACS integration or C2PA hardware signing, it demonstrates the structural necessity of separating provenance verification from pixel classification.

\section{Limitations}
\paragraph{Dataset lineage and pretraining overlap.}
Because the legacy Cheff-identified archive lacks generation seeds, prompts, and checkpoint metadata, our findings characterize this specific artifact collection rather than the theoretical limits of the Cheff architecture. Similarly, SynthCheX was upstream-filtered using HealthGPT, a framework that includes Qwen-derived variants~\cite{healthgpt2026officialreadme}, so we do not claim architectural independence between our Qwen submitter and SynthCheX curation. Furthermore, we cannot rule out pretraining exposure of public NIH or Open-I radiographs within web-scale DINOv3 or medical-scale MedGemma corpora, and Open-I report groupings do not guarantee patient-level disjointness.

\paragraph{Automated proxies vs.\ clinical ground truth.}
Both Qwen and MedGemma serve as automated algorithmic gates rather than board-certified radiologist evaluations; passing MedGemma's rubric does not imply diagnostic clinical utility. Likewise, our DINOv3 and SigLIP linear probes are statistical collection discriminators that remain susceptible to acquisition and compression shortcuts, rather than certified forensic detectors.

\paragraph{Confounded spatial masks and historical metadata.}
Our spatial masking perturbations simultaneously alter border geometry, intensity histograms, and spectral energy, precluding causal attribution to specific artifacts such as burned-in text annotations. Additionally, while all model checkpoints, split assignments, and evaluation predictions were verified, an upstream intermediate metadata file referenced by the primary raw DINOv3 selection logs could not be reconstructed byte-for-byte.

\section{Conclusion}
When generated chest radiographs are submitted at dataset intake, pixel-level screening cannot substitute for provenance verification. Although black-box candidate selection via Qwen substantially increases passage through MedGemma's technical quality gate, it provides no measurable advantage in evading both quality and DINOv3 origin checks simultaneously. Nevertheless, even without adversarial selection or post-processing, 152 of 5,000 SynthCheX radiographs pass both pixel checks undetected. Moreover, statistical origin screening is highly sensitive to routine distribution shifts: JPEG compression and resizing induce severe threshold drift and ranking loss, while applying an NIH-calibrated threshold to genuine Open-I radiographs triggers a 59.4\% false-alarm rate. Because pixel checks can filter obvious artifacts but cannot substantiate where an image came from, medical dataset intake pipelines must record and verify source provenance and intended use independently of the image pixels.

\FloatBarrier

%% file: sections/supplement.tex
\section{Reader Guide}
\label{sec:supp-guide}
This supplement provides complete methodological specifications, extended empirical evaluations, and reproducibility audit logs supporting the main paper. Section~\ref{sec:supp-defs} formalizes shared terminology and notation. Sections~\ref{sec:supp-data}--\ref{sec:supp-stats} detail our experimental design, covering dataset curation and split construction (\ref{sec:supp-data}), model checkpoints and feature extraction (\ref{sec:supp-models}), technical quality prompts and schema parsing (\ref{sec:supp-quality}), origin classifier training and conformal calibration (\ref{sec:supp-origin}), black-box candidate selection strategies (\ref{sec:supp-selection}), post-processing perturbations (\ref{sec:supp-processing}), and statistical inference protocols (\ref{sec:supp-stats}). Section~\ref{sec:supp-results} presents extended empirical diagnostics, Section~\ref{sec:supp-registry} details the provenance registry replay experiment, and Section~\ref{sec:supp-records} documents execution logs, protocol amendments, and compute costs.

\paragraph{Conventions for reading the tables.} Unless a caption explicitly states otherwise, the following reporting conventions apply throughout all supplementary tables:
\begin{itemize}
\item \emph{Denominators.} All proportions (pass, rejection, and flag rates) retain the full attempted test cohort in the denominator: 2,297 NIH, 1,000 Cheff-identified, and 5,000 SynthCheX radiographs.
\item \emph{Unavailable outputs.} An output is recorded as unavailable (U) whenever a model response fails schema parsing or an image fails decoding/processing. Unavailable outputs are never counted as passes, flags, or detections, and always remain in the denominator.
\item \emph{Confidence intervals.} Binomial proportions are accompanied by exact Wilson 95\% confidence intervals. Area under the ROC curve (AUROC) values and paired cross-method contrasts report 95\% bootstrap intervals computed from 2,000 base-image resamples, conditioned on the frozen models and calibrated thresholds (Section~\ref{sec:supp-stats}).
\item \emph{Calibration thresholds.} Decision thresholds correspond to nominal false-alarm rates $\alpha\in\{1\%, 5\%, 10\%\}$ calibrated on disjoint genuine NIH radiographs ($n=3{,}000$ for primary heads, $n=2{,}914$ for the legacy pilot head, and $n=2{,}930$ valid responses for the rubric-field probe).
\item \emph{Clustered observations.} Rows within a table frequently evaluate the same base radiographs across multiple processing conditions, selection strategies, or classifier heads; consequently, rows do not represent independent image samples.
\item \emph{Classifier nomenclature.} Tables follow the classifier naming conventions defined in Section~\ref{sec:supp-defs}. For secondary transfer heads, the suffix following the slash denotes the generator cohort(s) used during linear probe fitting (e.g., \texttt{CLS / Cheff}).
\item \emph{Abbreviations.} \texttt{N}, \texttt{C}, and \texttt{S} denote NIH, Cheff-identified, and SynthCheX, respectively; \texttt{U} denotes unavailable outputs; \texttt{CI} denotes confidence interval; and \texttt{k/n} denotes $k$ events out of $n$ attempts. In table stubs, \texttt{alt.} and \texttt{orig.} refer to the alternative and original quality rubrics, while \texttt{N} in column headers denotes sample count.
\item \emph{Counting units.} Tabulated counts refer to individual base radiographs unless the caption specifies another unit (e.g., processed view, control exposure, distinct pixel pattern, degradation pair, severity chain, or inference task).
\end{itemize}
All prompts, JSON schemas, and checkpoint commit hashes (Sections~\ref{sec:supp-models}, \ref{sec:supp-quality}, and~\ref{sec:supp-selection}) are reproduced verbatim without post hoc editing.

\section{Shared Definitions and Notation}
\label{sec:supp-defs}
\paragraph{Collections.} \emph{NIH} refers to genuine clinical chest radiographs drawn from the ChestX-ray14 subset of MaCheX. \emph{Cheff-identified} refers to a legacy archive of 3,000 radiographs identified as outputs of the Cheff cascaded latent diffusion model, for which exact checkpoint hashes and conditioning inputs are unrecovered. \emph{SynthCheX} refers to synthetic radiographs sampled from the fixed SynthCheX-75K-v2 release generated via Sana 0.6B. \emph{Open-I} refers to genuine frontal chest radiographs from the Indiana University hospital collection, held out exclusively for out-of-distribution false-alarm testing. We refer to each of these four cohorts as a \emph{collection}, reserving the term \emph{source} strictly for provenance claims recorded in the intake registry. The \emph{conditioning} of a synthetic image denotes the upstream input (e.g., text prompt or pathology label) supplied to the generator.

\paragraph{Models.} \emph{Qwen} (\texttt{Qwen3.8-27B}) serves as the submitter's black-box selector for scoring and ranking candidate images. \emph{MedGemma} (\texttt{MedGemma 1.5 4B IT}) enforces the receiver's technical quality gate, and its internal SigLIP vision tower supplies domain-specific features for the medical origin head. \emph{DINOv3} (\texttt{ViT-7B/16}) supplies general-purpose visual representations for the primary origin classifiers. \emph{MedForge-Reasoner} serves as an external zero-shot forensic baseline in our transfer evaluation. Exact repository revisions are listed in Section~\ref{sec:supp-models}.

\paragraph{Pipeline roles.} The \emph{submitter} curates synthetic candidates for dataset submission; acting as the adversary in our threat model, it observes only Qwen's outputs. The \emph{receiver} screens incoming submissions using frozen quality and origin models with pre-calibrated thresholds. Finally, the \emph{registry} audits the claimed acquisition source and declared purpose of each submission independently of pixel content.

\paragraph{Screening gates and outcomes.} The \emph{quality check} denotes the receiver's primary MedGemma evaluation gate (Section~\ref{sec:supp-quality}). An \emph{origin classifier} is an $L_2$-regularized logistic regression head trained over frozen visual features, and the \emph{origin check} pairs that classifier with a conformal decision threshold. For an input image $x$, the \emph{origin score} $s(x)$ (labeled ``suspicion score'' in several diagnostic plots) is the classifier's predicted syntheticity score, where larger values indicate greater likelihood of synthesis. The origin check \emph{flags} $x$ when $s(x)$ exceeds the calibrated threshold: flagging a generated image constitutes a correct \emph{rejection}, whereas flagging a genuine radiograph constitutes a \emph{false alarm}. The \emph{false-alarm rate} is the proportion of attempted genuine images flagged by the origin check. An image \emph{passes both checks} if and only if it passes the quality check and remains unflagged by the origin check.

\paragraph{Image preprocessing.} Our \emph{canonical} preprocessing pipeline converts each radiograph to grayscale luminance, resizes the longest dimension to 256 pixels via aspect-preserving Lanczos interpolation, zero-pads to a $256\times 256$ square, and replicates across three RGB channels without histogram equalization. Section~\ref{sec:supp-processing} defines the 12 additional post-processing perturbations.

\paragraph{Classifier heads and calibration policies.} A \emph{raw head} is trained solely on canonically preprocessed fit images, whereas an \emph{augmented head} is trained across five processed views per fit image (Section~\ref{sec:supp-origin}). The \emph{pilot head} is a legacy DINOv3 CLS probe preserved unchanged from our exploratory pilot study. The \emph{rubric-field probe} is a logistic regression classifier trained on eight numerical and indicator outputs from MedGemma's alternative quality rubric, while the \emph{pixel-statistic classifier} is trained on six low-level luminance and spectral statistics. Our \emph{primary head} is the raw DINOv3 CLS probe fitted on NIH and Cheff-identified images. Together, the seven \emph{core heads} comprise: (1) the pilot head, (2) raw DINOv3 CLS (primary), (3) raw DINOv3 mean-patch, (4) raw medical mean-patch (MedGemma SigLIP), (5) augmented DINOv3 CLS, (6) the rubric-field probe, and (7) the pixel-statistic classifier.

\paragraph{Secondary transfer heads and threshold regimes.} Secondary probes append a slash indicating the generator cohort(s) used during fitting: \texttt{CLS / Cheff} denotes a DINOv3 CLS probe trained on NIH vs.\ Cheff-identified images, \texttt{CLS / SynthCheX} denotes training on NIH vs.\ SynthCheX, and \texttt{Mixed} denotes joint training on NIH vs.\ an equal mixture of both generators. A \emph{canonical threshold} is calibrated exclusively on canonically preprocessed genuine NIH images. A \emph{matched threshold} is recalibrated on genuine NIH calibration images subjected to the exact same post-processing condition as the evaluated test set. Finally, the \emph{global threshold} is calibrated on a deterministic five-condition mixture (Section~\ref{sec:supp-processing}). We evaluate nominal false-alarm targets $\alpha\in\{0.01, 0.05, 0.10\}$, with $\alpha=0.05$ serving as the primary operating point in the main paper.

\paragraph{Quality rubrics.} The \emph{original rubric} scores technical quality $T\in[0,4]$, anatomical plausibility $A\in[0,4]$, and artifact severity $V\in[0,4]$, combining them into the composite score $Q=10T+10A+5(4-V)\in[0,100]$. The \emph{alternative rubric} records binary chest-radiograph recognition, binary assessability, ordinal anatomical visibility ($0$--$4$), ordinal structural artifact severity ($0$--$4$, or \texttt{null} when unassessable), and categorical uncertainty (\texttt{low}, \texttt{moderate}, or \texttt{high}); its fields are evaluated directly without mapping to a 100-point scale. A model response is \emph{valid} if it parses as JSON conforming strictly to the target schema (Section~\ref{sec:supp-quality}).

\paragraph{Candidate selection.} The \emph{candidate budget} $B$ denotes the size of the synthetic pool available to the submitter, and the \emph{submission limit} $K$ denotes the maximum number of images the submitter may forward to the receiver. \emph{Uniform}, \emph{quality}, and \emph{quality-constrained surrogate} selection policies are formalized in Section~\ref{sec:supp-selection}. The \emph{surrogate rating} is Qwen's integer judgment ($0$--$10$) estimating how closely a candidate resembles an authentic clinical radiograph.

\paragraph{Splits and ranking metrics.} The disjoint \emph{fit}, \emph{development}, \emph{calibration}, and \emph{test} splits govern probe training, hyperparameter/rubric selection, conformal threshold calibration, and held-out evaluation, respectively (Section~\ref{sec:supp-data}). AUROC measures the probability that a randomly drawn synthetic image receives a higher origin score than a randomly drawn genuine image; \emph{standardized partial AUC} integrates the ROC curve over false-positive rates in $[0, 0.10]$ normalized to unit scale (Section~\ref{sec:supp-stats}).

\section{Dataset and Split Construction}
\label{sec:supp-data}
Table~1 of the main paper summarizes our sample allocation across splits; here we detail the upstream archive audits, deduplication checks, and patient-disjointness guarantees.

\paragraph{NIH genuine cohort.} The upstream NIH archive index contains 112,120 distinct PNG radiographs and eight metadata tables; to minimize I/O overhead, we audited the full index and decoded only the sampled subset. The official upstream train/validation and test manifests list 86,524 and 25,596 images, respectively, with zero patient overlap (each list contains one more image than the widely cited counts of 86,523 and 25,595, an upstream discrepancy with no documented explanation). We enforce a strict one-image-per-patient constraint across all splits. After excluding the 500 upstream test patients used in our exploratory pilot study, all remaining 2,297 test patients are assigned to our held-out test split rather than sampling multiple radiographs per patient to pad the sample size.

\paragraph{Cheff-identified cohort.} The legacy Cheff-identified archive contains 8,672 PNG files and 27 CSV logs distributed across multiple subdirectories, from which our 3,000-image cohort is drawn. We did not filter candidates using legacy checker logs, as cross-auditing revealed duplicated score rows and filename mismatches that preclude verifying independent generation runs. Because exact diffusion checkpoints, conditioning prompts, and rejection-sampling histories are unrecovered, we treat this archive as an observational collection of legacy synthetic artifacts. To prevent data leakage from our exploratory pilot, all 1,250 Cheff-identified images used in the pilot (including pilot test images) are reassigned strictly to the fit split, supplemented by 250 previously unused images ($1{,}500$ fit total), while 500 and 1,000 unused images form the development and test splits.

\paragraph{SynthCheX cohort.} SynthCheX samples are drawn from a pinned snapshot of the 75,649-image SynthCheX-75K-v2 release. We sample 6,500 records uniformly at random, preserving upstream release identifiers and excluding 500 pilot images from the 5,000-image test split. Notably, SynthCheX was curated prior to public release using HealthGPT, a framework that includes Qwen-based vision-language variants~\cite{healthgpt2026officialreadme}; because the exact filtering checkpoint and conditioning mappings are undocumented in the release card, we do not assume architectural independence between SynthCheX curation and our Qwen submitter. Moreover, while the release provides filenames, pathology labels, and pixel arrays, it omits patient or report conditioning identifiers; we therefore audit for exact pixel duplicates and recoverable conditioning groups, explicitly treating unrecovered lineage as unknown rather than independent.

\paragraph{Open-I external genuine cohort.} Our out-of-distribution clinical panel comprises 957 frontal chest radiographs curated from the Indiana University Open-I repository~\cite{demnerfushman2016indiana}, grouped by radiology report. Open-I images are evaluated strictly at test time under the frozen NIH-calibrated thresholds without any threshold retuning. Because report groupings in the public archive do not guarantee unique patients and pretraining exposure cannot be ruled out, Open-I serves as an observational domain-shift audit rather than a clinical trial. No Open-I pixels are redistributed in our figures or code bundle.

\paragraph{Integrity hashes and deduplication audits.} Across all 17,297 radiographs in the four primary splits, we log cryptographic hashes of both the raw file bytes and canonically preprocessed pixel arrays. Prior to model inference, we verified zero exact pixel duplicates across splits and audited all recoverable conditioning groups, supplemented by a low-resolution perceptual similarity screen (noting that perceptual screening cannot rule out latent generator memorization or shared text conditioning). Our machine-readable manifests log whether each record was extracted from local Parquet shards or fetched via API, alongside complete SHA256 checksums.

\paragraph{Pre-committed identifiers and label blinding.} All split assignments and image identifiers were locked prior to model execution. Corrupt or unreadable assets remain in the denominator as unavailable attempts and are never silently replaced with backup images. During inference, models receive only pixel arrays and opaque hash identifiers; test-split collection labels were strictly withheld during development and hyperparameter selection.

\paragraph{Observational stratification.} Recorded NIH view projections (AP vs.\ PA) and report-mined pathology labels (\emph{No Finding} vs.\ \emph{other finding}) serve strictly as post hoc observational strata rather than ground-truth clinical diagnoses or model inputs. When reporting projection-standardized rates, we apply fixed development-split weights ($0.178$ AP, $0.822$ PA) and never condition intake decisions on submitter-declared metadata.

\section{Model and Checkpoint Details}
\label{sec:supp-models}
\begin{table*}[t]
\centering
\begin{tabular}{lll}
\toprule
Checkpoint & Revision & Processor input\\
\midrule
Qwen3.8-27B & \texttt{1\allowbreak d\allowbreak 4\allowbreak b\allowbreak f\allowbreak 0\allowbreak f\allowbreak 2\allowbreak f\allowbreak f\allowbreak 6\allowbreak 0\allowbreak 1\allowbreak 2\allowbreak f\allowbreak d\allowbreak 8\allowbreak 2\allowbreak 0\allowbreak 3\allowbreak 9f2fa52739d0dd7c60c0} & maximum 262,144 pixels\\
MedGemma 1.5 4B IT & \texttt{9\allowbreak 1\allowbreak 8\allowbreak 5\allowbreak 0\allowbreak 5\allowbreak 4\allowbreak 7\allowbreak d\allowbreak 9\allowbreak f\allowbreak 0\allowbreak b\allowbreak 2\allowbreak f\allowbreak d\allowbreak d\allowbreak 2\allowbreak 1\allowbreak a\allowbreak a7c5f4f3d1a8a52c243b} & $896\times896$\\
DINOv3 ViT-7B/16 & \texttt{b\allowbreak 8\allowbreak 0\allowbreak 3\allowbreak 6\allowbreak 7\allowbreak 7\allowbreak 5\allowbreak 3\allowbreak 7\allowbreak 7\allowbreak 3\allowbreak 6\allowbreak 4\allowbreak 8\allowbreak a\allowbreak 6\allowbreak 7\allowbreak 9\allowbreak 3\allowbreak 235ab9c65cdbb029506f} & $224\times224$\\
\bottomrule
\end{tabular}
\caption{Pinned Hugging Face model revisions (IT denotes instruction-tuned). Processor input indicates the internal vision-processor resolution cap or target grid, which is distinct from the native storage dimensions of the source radiographs.}
\label{tab:suppmodels}
\end{table*}

Table~\ref{tab:suppmodels} lists the exact repository commit hashes for our three core backbones: \texttt{Qwen/\allowbreak Qwen3.8-\allowbreak 27B}, \texttt{google/\allowbreak medgemma-\allowbreak 1.5-\allowbreak 4b-\allowbreak it}, and \texttt{facebook/\allowbreak dinov3-\allowbreak vit7b16-\allowbreak pretrain-\allowbreak lvd1689m}~\cite{qwen38card,medgemma15card,simeoni2025dinov3}. Note that Qwen's 262,144-pixel ceiling bounds the visual token budget rather than forcing upsampling of $256\times 256$ canonical inputs. All experiments were executed under Python 3.11.2, PyTorch 2.14.1+cu130, Transformers 5.19.0, NumPy 2.4.6, scikit-learn 1.9.1, and Pillow 12.3.0 (full environment lockfiles are included in the reproducibility archive). All neural backbones remain frozen and fully GPU-resident in native BF16 precision without weight quantization, CPU parameter offloading, or mixed-precision autocasting.

\subsection{Feature Extraction Backbones}
For DINOv3 (\texttt{ViT-7B/16}), we apply the official 224-pixel vision processor to extract either the final-layer CLS token embedding or the spatial average across patch tokens (strictly excluding CLS and register tokens). For our domain-specific \emph{medical representation}, we run the standalone MedGemma 1.5 SigLIP vision tower at its native $896\times 896$ input resolution and average its 4,096 patch embeddings in FP32 prior to the multimodal projection layer. We explicitly extract standalone vision-tower embeddings rather than caching intermediate hook activations during text generation, as development audits revealed minor floating-point discrepancies between hooked generation states and standalone forward passes. Every persisted feature vector is cryptographically linked to its exact input pixel hash.

\subsection{Vision-Language Decoding Configurations}
Qwen inference uses deterministic greedy decoding with thinking mode disabled and a 384-token generation ceiling; MedGemma inference uses deterministic greedy decoding with a 1,024-token ceiling. We locked Qwen's non-thinking configuration after verifying on a 96-image development panel that both rubrics consistently yielded schema-valid JSON responses well within the token budget and confirming exact processor-to-decoder tensor routing.

\subsection{MedForge-Reasoner Transfer Protocol}
\label{sec:supp-medforge}
To evaluate whether a specialized medical forgery detector transfers to full-image synthetic chest radiographs, we benchmark \texttt{RichardChenZH/\allowbreak MedForge-\allowbreak Reasoner} pinned to revision \texttt{a864eed9b76cde46f0e154c3497d646550995c87}~\cite{medforge2026checkpoint,chen2026medforgersi}. Because MedForge-Reasoner is a Qwen-family vision-language model trained to localize regional tampering in medical scans, evaluating it on whole-image diffusion syntheses constitutes a cross-task transfer test rather than a reproduction of its native benchmark. Following the official reference implementation, we resize canonical RGB inputs to $1024\times 1024$ (which upsamples $256\times 256$ canonical arrays without restoring native high-frequency detail) and run batch-size-one greedy decoding in native BF16 with a 2,048-token ceiling using the exact official prompts reproduced below.

Based on a 4-image GPU timing calibration (56.998\,s total wall time, including 35.414\,s generation), we pre-committed the evaluation scope to our 300-image canonical processing panel per collection (900 images total) plus all unique radiographs selected by Qwen across the fixed test submission lists that were not already present in the 900-image panel (195 additional images, for 1,095 total evaluations). Submissions overlapping the 900-image panel reuse the cached verdict directly.

We parse MedForge-Reasoner outputs by mapping an explicit, unambiguous concluding classification (\texttt{real}, \texttt{authentic} vs.\ \texttt{deepfake}, \texttt{fake}, \texttt{forged}, \texttt{manipulated}) following a closed \texttt{<think>} reasoning block to a binary verdict. Unclosed reasoning blocks, self-contradictory labels, truncated generations hitting the 2,048-token cap, or missing classifications are recorded as unavailable. Because MedForge-Reasoner emits discrete categorical verdicts without a continuous calibrated score, it defines a single native operating point; we do not heuristically convert text rationales into probabilities or interpolate 1\%, 5\%, or 10\% false-alarm curves. Section~\ref{sec:supp-medforge-results} presents the resulting transfer performance.

\paragraph{Exact official system and user prompts.}
\begin{lstlisting}
You are an expert in medical image forensics. Analyze the provided image to determine if it is a deepfake or authentic. First, perform a step-by-step examination of the image content, looking for artifacts, inconsistencies, or biological implausibilities. Use <think> tags to articulate your reasoning process. If you identify manipulated regions, localize them using bounding boxes within your reasoning. Conclude your analysis with a final classification.
\end{lstlisting}
\begin{lstlisting}
Is this image deepfake or real?
\end{lstlisting}

\section{Quality Prompts and Response Parsing}
\label{sec:supp-quality}
\subsection{Prompt Delivery and Blinding}
Both Qwen (the submitter's selector) and MedGemma (the receiver's quality gate) receive only the pixel array (or ordered image pair for pairwise tie-breaking) and the prompt text formatted via each model's official chat template. All filenames, pathology labels, radiology reports, and provenance claims are stripped prior to inference. If a model emits native reasoning tokens, they are logged for auditing but never substituted for the required JSON payload. Unparseable outputs and token-limit truncations remain in the denominator as unavailable attempts without post hoc manual repair.

\subsection{Original Rubric (Qwen Selector and MedGemma Baseline)}
Qwen applies the original rubric to score every candidate pool, and MedGemma evaluates the same rubric as a baseline comparison:
\begin{lstlisting}
Assess only the supplied chest radiograph pixels for technical quality, visible anatomical plausibility, and conspicuous artifacts. The image's source is not given. Do not infer an acquisition record or certify its origin. Findings, abnormal anatomy, medical devices, portable/AP positioning, rotation, and nonstandard anatomy are not automatic failures; assess whether the visible appearance is coherent in its possible clinical and acquisition context. Do not reward absence of disease or devices. Do not diagnose the patient. Do not use any filename, label, report, or source metadata.

Return exactly one JSON object, without Markdown or other text:
{"technical_quality":0,"anatomical_plausibility":0,"artifact_severity":0,"assessable":true,"uncertainty":"moderate","rationale":"brief visible evidence"}

Use integer scores from 0 through 4 with these anchors:
technical_quality: 0 = unreadable/essential chest content unavailable; 1 = severe technical degradation; 2 = noticeable degradation with substantial chest content still visible; 3 = minor limitations; 4 = clearly displayed chest content without conspicuous technical degradation.
anatomical_plausibility: 0 = grossly incoherent visible anatomy; 1 = several severe unexplained structural inconsistencies; 2 = noticeable unexplained inconsistencies; 3 = minor uncertain inconsistencies; 4 = no conspicuous unexplained structural inconsistency. Use uncertainty when pathology, projection, or limited resolution could explain an appearance.
artifact_severity: 0 = no conspicuous unexplained synthesis-like or rendering artifact; 1 = minor localized artifact; 2 = moderate artifact; 3 = severe artifact; 4 = pervasive artifact. This is a visible-artifact assessment, not a real/fake verdict.
assessable: true if enough chest content is visible to apply the rubric; otherwise false. uncertainty: exactly "low", "moderate", or "high". rationale: describe visible evidence in at most 60 words, without asserting verified acquisition, patient identity, or clinical usefulness.
\end{lstlisting}
\begin{lstlisting}
{
  "type": "object",
  "additionalProperties": false,
  "required": ["technical_quality", "anatomical_plausibility", "artifact_severity", "assessable", "uncertainty", "rationale"],
  "properties": {
    "technical_quality": {"type": "integer", "minimum": 0, "maximum": 4},
    "anatomical_plausibility": {"type": "integer", "minimum": 0, "maximum": 4},
    "artifact_severity": {"type": "integer", "minimum": 0, "maximum": 4},
    "assessable": {"type": "boolean"},
    "uncertainty": {"enum": ["low", "moderate", "high"]},
    "rationale": {"type": "string"}
  }
}
\end{lstlisting}

\subsection{Alternative Rubric (Primary MedGemma Quality Gate)}
\begin{lstlisting}
Assess the pixels without assuming they depict a chest radiograph. First decide whether recognizable chest-radiograph content is present and whether enough of it is visible to assess. Flat fields, noise, blank images and unrelated pictures are not assessable chest radiographs; do not invent anatomical structures. The source is unknown. Do not certify acquisition, origin or clinical usefulness, diagnose disease, or use filename/metadata. Pathology, devices, AP/portable positioning and unusual coherent anatomy are not failures by themselves.
Return one JSON object with is_chest_radiograph (boolean), assessable (boolean), visibility (integer0..4), structural_artifact_severity (integer0..4 or null when unassessable), uncertainty (low/moderate/high), and rationale (brief visible evidence).
Visibility anchors:0=chest content absent/unreadable;1=severe visibility loss;2=substantial visible chest content with noticeable limitations;3=minor visibility limitations;4=chest structures clearly displayed at this supplied resolution. Structural-artifact anchors:0=no conspicuous unexplained structural/rendering inconsistency;1=minor localized;2=moderate;3=severe;4=pervasive. Separate visibility from artifacts; uncertain anatomy explained by disease/projection should raise uncertainty rather than automatically raise artifact severity. If not a recognizable or assessable chest radiograph, visibility=0 and structural_artifact_severity=null. Do not reward absence of pathology or devices.
\end{lstlisting}
\begin{lstlisting}
{
  "type": "object",
  "additionalProperties": false,
  "required": [
    "is_chest_radiograph",
    "assessable",
    "visibility",
    "structural_artifact_severity",
    "uncertainty",
    "rationale"
  ],
  "properties": {
    "is_chest_radiograph": {
      "type": "boolean"
    },
    "assessable": {
      "type": "boolean"
    },
    "visibility": {
      "type": "integer",
      "minimum": 0,
      "maximum": 4
    },
    "structural_artifact_severity": {
      "type": [
        "integer",
        "null"
      ],
      "minimum": 0,
      "maximum": 4
    },
    "uncertainty": {
      "enum": [
        "low",
        "moderate",
        "high"
      ]
    },
    "rationale": {
      "type": "string"
    }
  }
}
\end{lstlisting}

\subsection{Schema Validation and Decision Rules}
A response is \emph{valid} if and only if it parses as a JSON object conforming strictly to the target schema above. Among schema-valid responses, outputs indicating \texttt{assessable: false}, \texttt{is\_chest\_radiograph: false}, or \texttt{structural\_artifact\_severity: null} are tracked as distinct semantic states, whereas JSON parse errors and runtime exceptions are recorded as \emph{unavailable}.

Under the \emph{original rubric}, a valid response passes if \texttt{assessable: true} and $Q\geq 75$. Under the \emph{alternative rubric}, a valid response passes if \texttt{is\_chest\_radiograph: true}, \texttt{assessable: true}, $\texttt{visibility}\geq 3$, and non-null $\texttt{structural\_artifact\_severity}\leq 1$. In addition, our deterministic \emph{flat-image guard} rejects any input whose 8-bit grayscale dynamic range is $\leq 1$ and standard deviation is $\leq 0.5$ on $[0, 255]$. We report quality passage both with and without this guard.

\subsection{Pairwise Ordering Rules for Degradation Audits}
To evaluate psychometric sensitivity to synthetic image degradations, we test whether each rubric ranks an undegraded base image above its degraded counterparts. Under the original rubric, pairs are ordered by composite score $Q$. Under the alternative rubric, pairs are ordered lexicographically by \texttt{visibility} (higher is better) and then by \texttt{structural\_artifact\_severity} (lower is better), placing schema-valid unassessable, non-chest, or null-artifact responses at the bottom of the ordering. Each clean--degraded comparison yields one of four outcomes: \emph{strict} (undegraded ranked strictly higher), \emph{tie} (equal rank), \emph{reversal} (degraded ranked strictly higher), or \emph{unavailable} (either view failed schema parsing).

\subsection{Development Selection of the Alternative Rubric}
We compared the original and alternative rubrics on a development stress panel comprising 100 pilot/development radiographs per collection evaluated across 10 conditions: canonical plus three severities each of Gaussian blur ($\sigma\in\{0.75, 1.5, 3.0\}$), bilinear downsampling ($\{192, 128, 64\}$ pixels), and additive Gaussian noise ($\sigma_{\text{pixel}}\in\{4, 12, 32\}$ on $[0,255]$). This yields 900 clean--degraded pairs per collection (2,700 pairs total). We also evaluated 70 non-radiograph control exposures spanning 60 distinct synthetic pixel patterns (flat fields, random noise, phase-scrambled radiographs, and checkerboards; a distinct pattern fails if any repeated exposure passes).

Prior to inspecting development outcomes, we locked three joint eligibility criteria required to adopt the alternative rubric as our primary receiver gate:
\begin{enumerate}
\item \emph{Invalid-pattern rejection:} Passing $\leq 10\%$ ($\leq 6/60$) of distinct non-radiograph control patterns without the deterministic flat-image guard.
\item \emph{Clean genuine assessability:} Increasing the rate of non-assessable or unavailable responses on the 100 clean NIH development radiographs by at most $+5.0$ percentage points (pp) relative to the original rubric.
\item \emph{Degradation sensitivity:} Increasing the pooled proportion of strict clean-over-degraded orderings across all 2,700 pairs by at least $+5.0$\,pp relative to the original rubric.
\end{enumerate}

\begin{figure*}[t]
\centering
\includegraphics[width=\textwidth]{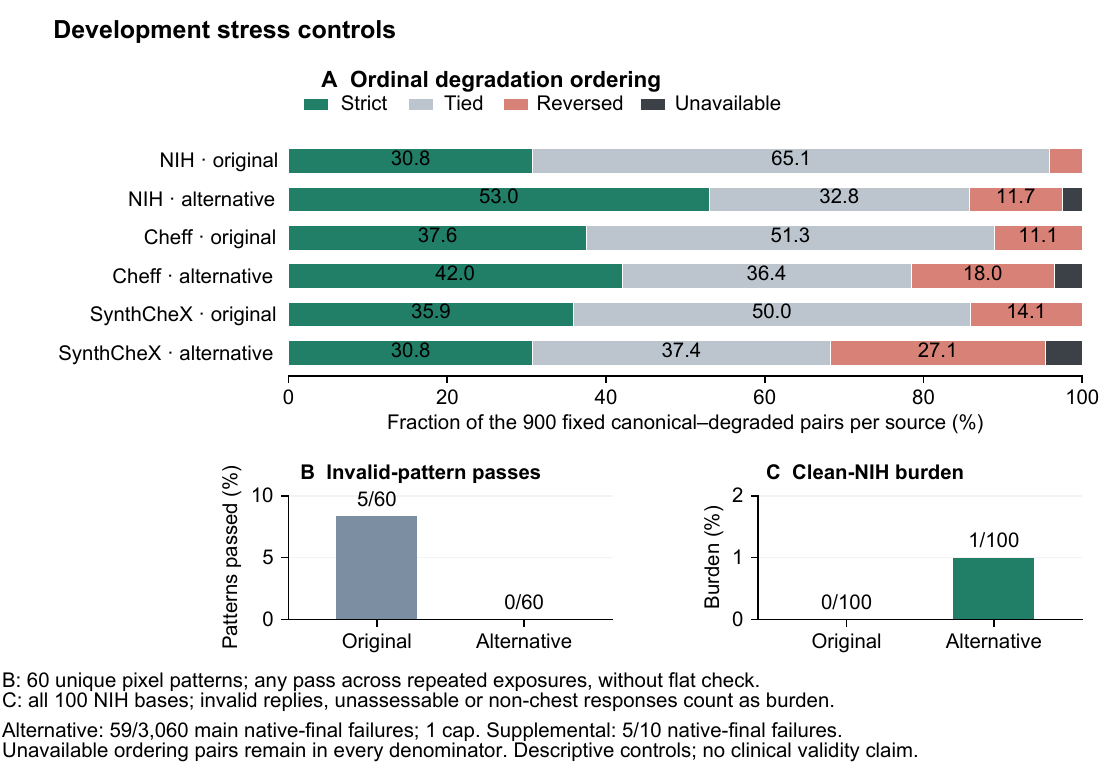}
\caption{Development stress evaluation of the quality rubrics across 100 radiographs (900 clean--degraded pairs) per collection and 60 distinct non-radiograph control patterns. Panel~A reports pairwise ordering outcomes (strict, tied, reversed, unavailable); Panel~B reports invalid-pattern passes without the deterministic flat-image guard; Panel~C (``burden'') reports clean NIH radiographs lacking a valid assessable judgment. While the alternative rubric satisfies all three pre-registered adoption criteria on pooled development data, it exhibits lower strict ordering on SynthCheX and higher parse unavailability (``native-final failures'' and ``cap'' denote JSON parse failures and token-limit truncations, respectively).}
\label{fig:development-quality-stress}
\end{figure*}

On the development stress panel (Figure~\ref{fig:development-quality-stress}), the alternative rubric met all three pre-committed criteria: (1) without the flat-image guard, it passed $0/60$ invalid patterns versus $5/60$ for the original rubric; (2) on clean NIH images, it produced $1/100$ non-assessable/unavailable responses versus $0/100$ for the original rubric ($+1.0$\,pp); and (3) across the 2,700 degradation pairs, it achieved $1{,}132/2{,}700$ strict orderings versus $938/2{,}700$ for the original rubric ($+7.19$\,pp pooled gain). We note, however, that this pooled improvement masks cohort heterogeneity: on SynthCheX development pairs specifically, strict orderings decreased from $323/900$ to $277/900$, and the alternative rubric incurred higher parse/token-cap unavailability (59 failed final replies and 1 token-cap truncation, yielding 97 unavailable ordering pairs). Because the three pre-registered criteria were satisfied on development data, we locked the alternative rubric (paired with the flat-image guard) as the primary quality gate and subjected it to an independent held-out validity audit on test data.

\subsection{Held-Out Rubric Audit and Post-Processing Quality Protocol}
To evaluate rubric validity out-of-sample, our held-out test panel samples 100 test radiographs per collection (using fresh random seeds) across six unseen degradation parameterizations: Gaussian blur at $\sigma\in\{1.0, 2.0\}$, resampling through $\{160, 96\}$ pixels, and additive noise at $\sigma_{\text{pixel}}\in\{8, 24\}$. Together with the canonical view, this yields 2,100 radiograph evaluations (1,800 clean--degraded pairs across collections) plus 70 non-radiograph control exposures (60 distinct patterns), totaling 2,170 evaluations per rubric (reusing 300 cached canonical test assessments and executing 1,870 new model calls per rubric).

In parallel, we evaluate quality-gate stability across four realistic post-processing routes—canonical, JPEG 75, resize 128, and native 512—across a fixed subset of 300 test radiographs per collection (3,600 views per rubric, reusing the 900 canonical test assessments and executing 2,700 new calls per rubric). All paired comparisons cluster views at the base-image level.

\section{Origin Classifier Construction and Calibration}
\label{sec:supp-origin}
\subsection{Linear Probe Training and Regularization Selection}
\paragraph{Sample budgets.} Our primary origin classifier (raw DINOv3 CLS) is trained on 1,500 genuine NIH and 1,500 Cheff-identified fit images, with regularization selected on 1,000 NIH and 500 Cheff-identified development images. To ensure fair cross-generator comparisons at matched sample sizes, our secondary generator-specific probes (\texttt{Cheff}, \texttt{SynthCheX}, and \texttt{Mixed}) use standardized fit budgets of 1,000 genuine NIH and 1,000 synthetic images (500 per generator for \texttt{Mixed}), and development budgets of 1,000 genuine NIH and 500 synthetic images (250 per generator for \texttt{Mixed}). Deterministic hash salts ensure that the 1,000-image secondary subsets are strictly nested within the 1,500-image primary splits. SynthCheX data never enter the training or hyperparameter selection of the primary head.

\paragraph{Optimization protocol.} Feature coordinates are standardized to zero mean and unit variance using fit-split statistics exclusively. We fit $L_2$-regularized logistic regression probes via deterministic L-BFGS (up to 5,000 iterations), sweeping inverse regularization strength $C\in\{0.01, 0.1, 1, 10\}$. For each probe, we select the $C$ value that maximizes synthetic rejection on the development split at a 5\% development false-alarm rate (averaged across training views for multi-condition heads), breaking ties toward stronger regularization (smaller $C$).

\subsection{Conformal Threshold Calibration}
Following hyperparameter selection, all newly fitted DINOv3, medical SigLIP, and pixel-statistic probes are calibrated on the disjoint 3,000-image NIH calibration split. As defined in Equation~1 of the main paper, for any test input $x$ with origin score $s(x)$, we compute the conformal outlier $p$-value $p_{\rm real}(x)=(1+\sum_{i=1}^{n}\mathbf{1}\{s_i\geq s(x)\})/(n+1)$ over the $n$ genuine calibration scores and flag $x$ when $p_{\rm real}(x)\leq\alpha$ for $\alpha\in\{0.01, 0.05, 0.10\}$ (leaving exact threshold ties unflagged). Exact numerical score cutoffs, tie counts, and empirical calibration rejection rates for every head are exported in our reproducibility archive.

\subsection{Reference Calibration Sets for Auxiliary Baselines}
\paragraph{Legacy pilot head.} Because the exploratory pilot head was trained prior to final split freezing, a lineage audit identified 58 NIH calibration patients that had been used during pilot fitting and 28 used during pilot regularization tuning. We excised all 86 overlapping patients prior to test evaluation, leaving a strictly disjoint reference set of $n=2{,}914$ genuine calibration patients ($n+1=2{,}915$) while keeping the pilot probe weights unchanged.

\paragraph{Rubric-field probe.} The rubric-field probe trains an $L_2$-regularized logistic regression head on eight numerical/indicator fields extracted from MedGemma's alternative quality rubric: \texttt{visibility}, \texttt{structural\_artifact\_severity} (coded as 0 when null), \texttt{is\_chest\_radiograph}, \texttt{assessable}, a binary null-artifact indicator, and one-hot uncertainty indicators (excluding free-text rationales). Among the 1,500 NIH and 1,500 Cheff-identified fit attempts, 1,461 and 1,469 yielded valid JSON responses (976/1,000 and 488/500 on development). Unavailable responses are omitted during fitting and recorded as unavailable at test time. Because all four $C\in\{0.01, 0.1, 1, 10\}$ values rejected exactly 5/500 Cheff-identified development images at $\alpha=5\%$, our tie-breaking rule selected $C=0.01$.

For calibration, a lineage audit identified 10 NIH calibration patients whose images had been inspected during quality-rubric prompt engineering. Excluding those 10 patients leaves 2,990 eligible calibration patients, among which MedGemma returned 2,930 valid JSON responses and 60 unavailable outputs. Thus, the rubric-field probe is calibrated over $n=2{,}930$ valid reference responses ($n+1=2{,}931$; note that discrete score ties cause its 1\% and 5\% thresholds to coincide).

\paragraph{Pixel-statistic classifier.} To quantify how much collection separability arises from elementary low-level image properties alone, we fit a logistic regression probe on six standardized luminance statistics computed on $[0,1]$-normalized grayscale arrays: (1) mean intensity, (2) intensity standard deviation, (3) 256-bin luminance histogram entropy, (4) high-frequency spectral energy fraction (Fourier power in $[0.25, 0.50]$ cycles/pixel normalized by total non-DC power), (5) border-to-interior gradient ratio (comparing an 8-pixel peripheral rim against the interior), and (6) $8\times 8$ JPEG blockiness ratio (comparing gradient magnitude along 8-pixel grid boundaries against interior pixel steps). This baseline is trained on the primary NIH/Cheff fit split and calibrated on all $n=3{,}000$ genuine calibration images.

\subsection{Multi-Condition Augmented Head and Response-Feature Ablation}
Our multi-condition \emph{augmented head} trains a DINOv3 CLS logistic regression probe across five processed views of each fit image: canonical, JPEG 75, resize 128, native 512, and 4-pixel padding (weighting each view by $1/5$ so every base patient retains unit total weight during standardization and fitting). In Section~\ref{sec:supp-results}, we decompose multi-condition performance across three pre-registered contrasts:
\begin{itemize}
\item \textbf{Contrast A (Model effect under matched calibration):} Augmented CLS head with condition-matched thresholds minus raw CLS head with condition-matched thresholds.
\item \textbf{Contrast B (End-to-end policy shift):} Augmented CLS head with condition-matched thresholds minus raw CLS head with the static canonical threshold.
\item \textbf{Contrast C (Recalibration effect alone):} Raw CLS head with condition-matched thresholds minus raw CLS head with the static canonical threshold.
\end{itemize}
On the 13-condition development benchmark, Contrast~A increased mean Cheff-identified rejection by $+4.62$\,pp (95\% paired CI: $[+3.95, +5.26]$) while changing genuine NIH false alarms by $+0.12$\,pp ($[-0.58, +0.84]$). Although $+4.62$\,pp fell slightly short of our $+5.0$\,pp development target, we evaluated the frozen augmented head on the test split without further tuning as a pre-registered core comparison.

We also conducted a development ablation testing a \emph{perturbation-response probe} that concatenates canonical features with feature displacements under compression and resizing: $[h(x),\, h(J(x))-h(x),\, h(R(x))-h(x)]$, where $h(\cdot)$ denotes the frozen DINOv3 CLS embedding and $J(x), R(x)$ denote JPEG 75 and resize 128 views. Benchmarked against a 3-view ensemble average of the raw head on development data, the response-feature probe rejected 500/500 versus 495/500 Cheff-identified development images ($+1.0$\,pp gain, both flagging 49/1,000 NIH images). Because this $+1.0$\,pp gain did not meet our $+5.0$\,pp adoption threshold, the response-feature probe was not promoted to final test evaluation.

\section{Candidate Selection Procedures}
\label{sec:supp-selection}
\subsection{Selection Policies and Denominator Accounting}
In our black-box selection protocol, the submitter ranks pre-existing synthetic test pools of size $B\in\{100, 250, 500, 1{,}000\}$ (plus $B=5{,}000$ for SynthCheX) using cached Qwen judgments and forwards at most $K=20$ radiographs. Nested candidate pools are constructed via deterministic hash ordering so that smaller budgets are strict subsets of larger budgets. We compare three selection policies:
\begin{itemize}
\item \emph{Uniform:} Draws $K$ candidates uniformly at random without replacement.
\item \emph{Quality:} Ranks candidates in descending order of Qwen's composite score $Q$ (placing unavailable responses last and breaking ties via a deterministic hash) and selects the top $K$.
\item \emph{Quality-constrained surrogate:} Filters for candidates satisfying \texttt{assessable: true}, $Q\geq 75$, and a valid Qwen surrogate authenticity rating, sorting qualifying candidates by surrogate rating, then by $Q$, and finally by deterministic tie hash. If fewer than $K$ candidates qualify, fewer than $K$ are submitted; we report pass rates both as a fraction of actual submissions ($\le K$) and as a yield per pool candidate ($B$).
\end{itemize}
To quantify sensitivity to pool composition, we also evaluate 100 Monte Carlo pool draws on cached scores alongside 2,000 end-to-end bootstrap resamples that repeat candidate selection inside each resample. All submission lists and pool orderings were locked prior to joining with receiver outcomes.

\subsection{Qwen Surrogate Authenticity Prompt}
\begin{lstlisting}
Inspect only these chest-radiograph pixels. Rate how acquisition-like the image appears, using visible structural coherence and conspicuous rendering/synthesis cues. Do not claim a verified acquisition record or use pathology absence, medical devices, AP/portable position, filenames or labels to certify origin. This is a surrogate ranking judgment, not a calibrated probability. Return JSON only: assessable:boolean, acquisition_like_rank:integer0..10 (0=strong conspicuous rendering incoherence;5=ambiguous appearance;10=strongly coherent acquisition-like appearance), uncertainty:low/moderate/high, rationale:string. Unusual but coherent clinical anatomy is not evidence of synthesis. If unassessable, acquisition_like_rank=0 and uncertainty=high.
\end{lstlisting}
\begin{lstlisting}
{
  "type": "object",
  "additionalProperties": false,
  "required": [
    "assessable",
    "acquisition_like_rank",
    "uncertainty",
    "rationale"
  ],
  "properties": {
    "assessable": {
      "type": "boolean"
    },
    "acquisition_like_rank": {
      "type": "integer",
      "minimum": 0,
      "maximum": 10
    },
    "uncertainty": {
      "enum": [
        "low",
        "moderate",
        "high"
      ]
    },
    "rationale": {
      "type": "string"
    }
  }
}
\end{lstlisting}

\subsection{Pairwise Tie-Breaking Refinement}
Because discrete integer rubrics produce frequent score ties at $Q=100$, we evaluated a secondary pairwise tie-breaking refinement at $B=100, K=20$ using Qwen to compare tied top-scoring candidates against fixed anchor images in both presentation orders:
\begin{lstlisting}
Two images are supplied in order A then B. Compare their visible technical quality and conspicuous structural/rendering artifacts. Do not judge origin, absence of pathology, AP/PA or devices as quality by themselves. First assess whether each depicts enough chest-radiograph content. Return only JSON: assessable_a:boolean, assessable_b:boolean, winner:"A"/"B"/"tie"/"neither", strength:integer0..3 (0=tie/unassessable,1=slight,2=clear,3=large difference), rationale:string with visible evidence. Judge visibility and structural coherence, not verified clinical usefulness. The images sources are unknown.
\end{lstlisting}
\begin{lstlisting}
{
  "type": "object",
  "additionalProperties": false,
  "required": [
    "assessable_a",
    "assessable_b",
    "winner",
    "strength",
    "rationale"
  ],
  "properties": {
    "assessable_a": {
      "type": "boolean"
    },
    "assessable_b": {
      "type": "boolean"
    },
    "winner": {
      "enum": [
        "A",
        "B",
        "tie",
        "neither"
      ]
    },
    "strength": {
      "type": "integer",
      "minimum": 0,
      "maximum": 3
    },
    "rationale": {
      "type": "string"
    }
  }
}
\end{lstlisting}
During development, we audited 24 candidates per collection paired against 4 fixed anchors in both AB and BA orders (576 pairwise tasks, 288 swapped pairs), requiring $\geq 90\%$ valid comparability, $\geq 80\%$ order-swap consistency, and $\leq 7{,}200$\,s projected runtime for a 1,600-query budget. All 288 development pairs were comparable, achieving $90.6\%$ overall swap consistency ($261/288$: $84/96$ NIH, $87/96$ Cheff-identified, and $90/96$ SynthCheX) and a projected 1,600-query runtime of 6,116.1\,s. Pairwise outcomes are converted into continuous rankings via an $L_2$-regularized ($\lambda=1$) Bradley--Terry model (assigning $0.5$ wins to ties and ignoring self-reported strength). In our test evaluation, pairwise tie-breaking is restricted exclusively to resolving top-score ties at $B=100, K=20$.

\section{Post-Processing Perturbation Conditions}
\label{sec:supp-processing}
\paragraph{The 13 processing conditions.} We evaluate 1,000 test radiographs per collection across 13 conditions: (1) \emph{canonical} ($256\times 256$ Lanczos-resized luminance replicated to RGB); (2--5) grayscale \emph{JPEG compression} at quality factors $q\in\{95, 85, 75, 50\}$; (6--8) Lanczos \emph{downsampling and restoration} through intermediate resolutions of $\{192, 128, 64\}$ pixels back to 256 pixels; (9) \emph{native 512}, which extracts a $512\times 512$ representation directly from the source archive prior to backbone ingestion (noting that Cheff-identified archives are stored at $256\times 256$ and thus upscaled, and DINOv3/MedGemma processors still resize inputs to $224\times 224$ and $896\times 896$, respectively); (10) \emph{4-pixel border mask} (zeroing a 4-pixel peripheral rim, or $6.15\%$ of image area); (11) \emph{4-pixel padding} (resizing content to $248\times 248$ and adding a 4-pixel black border); (12) \emph{64-pixel central mask} (zeroing the central $64\times 64$ square, or $6.25\%$ of area); and (13) \emph{26-pixel border mask} (zeroing a 26-pixel peripheral rim, or $36.5\%$ of area).

\paragraph{Calibration threshold policies.} Across all 13 conditions, Table~\ref{tab:all-routes-primary-policies} compares four threshold regimes over the exact same 3,000 calibration patients and 1,000 test patients per collection: (1) \emph{Raw/canonical} (raw CLS head with the static canonical threshold); (2) \emph{Raw/matched} (raw CLS head recalibrated on genuine calibration images processed under the matching condition); (3) \emph{Augmented/matched} (5-condition augmented CLS head with condition-matched calibration thresholds); and (4) \emph{Augmented/global} (augmented CLS head calibrated once on a deterministic 5-condition mixture assigning each of the 3,000 calibration images via $\text{SHA256}(\texttt{20261014|global\_route\_calibration|base\_id})\bmod 5$ to canonical [579], JPEG 75 [621], resize 128 [579], native 512 [627], or 4-pixel padding [594]). Note that the global threshold guarantees nominal false-alarm control only marginally across that 5-condition mixture rather than conditionally within individual processing routes.

\paragraph{Handling of unavailable predictions.} Every evaluated view receives either a finite scalar origin score or an explicit unavailable status. Calibrating any continuous feature head requires all 3,000 reference scores to be finite (whereas the rubric-field probe calibrates over its 2,930 valid JSON responses).

\section{Statistical Procedures}
\label{sec:supp-stats}
\paragraph{Proportions and Wilson intervals.} All pass, flag, and rejection rates retain all attempted radiographs in the denominator (including zero-count numerators) and are reported with exact Wilson 95\% confidence intervals. For a fixed 20-image submission list, the Wilson interval characterizes binomial uncertainty on that specific list conditional on the pool.

\paragraph{Cluster bootstrap inference.} To account for repeated measurements across processed views, candidate pools, and classifier heads, all AUROC intervals, standardized partial AUC intervals (integrated over $\text{FPR}\in[0, 0.10]$), and paired differences are computed via 2,000 non-parametric cluster-bootstrap resamples at the base-image level within each collection. For candidate selection at $B\in\{1{,}000, 5{,}000\}$, our end-to-end bootstrap resamples candidate pools and re-executes the selection rule inside each replicate to capture selection variance. All intervals condition on the frozen fitted models and calibrated thresholds.

\paragraph{Representation proximity via MMD$^2$.} To quantify distributional divergence between synthetic cohorts and genuine NIH radiographs in each frozen embedding space, we compute the unbiased squared Maximum Mean Discrepancy ($\text{MMD}^2$) using the degree-3 polynomial kernel $k(x,y)=(x^\top y/d + 1)^3$, where $d$ is the embedding dimension. We evaluate fixed panels of 500 radiographs per collection alongside a disjoint 500-vs.-500 genuine NIH--NIH reference split, reporting both raw embeddings and $\ell_2$-normalized angular embeddings scaled by $\sqrt{d}$ (retaining negative unbiased estimates without truncation). Because embedding scales differ across architectures, $\text{MMD}^2$ magnitudes are compared only within a given backbone and normalization.

\paragraph{Nearest-neighbor audits.} To visualize embedding proximity without cherry-picking, we select one query radiograph per collection via a deterministic hash rule from each 500-image panel and retrieve its cosine nearest neighbor from the 1,500-image genuine NIH fit split across DINOv3 CLS, DINOv3 mean-patch, and MedGemma SigLIP mean-patch spaces. Note that nearest-neighbor retrieval against a 1,500-image reference illustrates representation geometry rather than proving absence of training-set memorization.

\paragraph{Within-collection correlations and score shifts.} To probe low-level feature dependencies without confounding collection labels, we compute within-collection Spearman rank correlations ($\rho$) between origin scores and the six pixel statistics (with 2,000 bootstrap resamples). For post-processing perturbations, we report median paired score shifts on the 1,000 genuine NIH test images alongside each head's canonical interquartile range (IQR) as a natural scale reference.

\section{Extended Results and Diagnostics}
\label{sec:supp-results}
This section presents the complete empirical tables and diagnostic figures referenced in the main paper, organized by evaluation question. All tables follow the reporting conventions established in Section~\ref{sec:supp-guide}.

\subsection{Technical Quality Evaluation on the Test Split}
\label{sec:supp-quality-results}
\begin{figure*}[tbp]
\centering
\includegraphics[width=\textwidth]{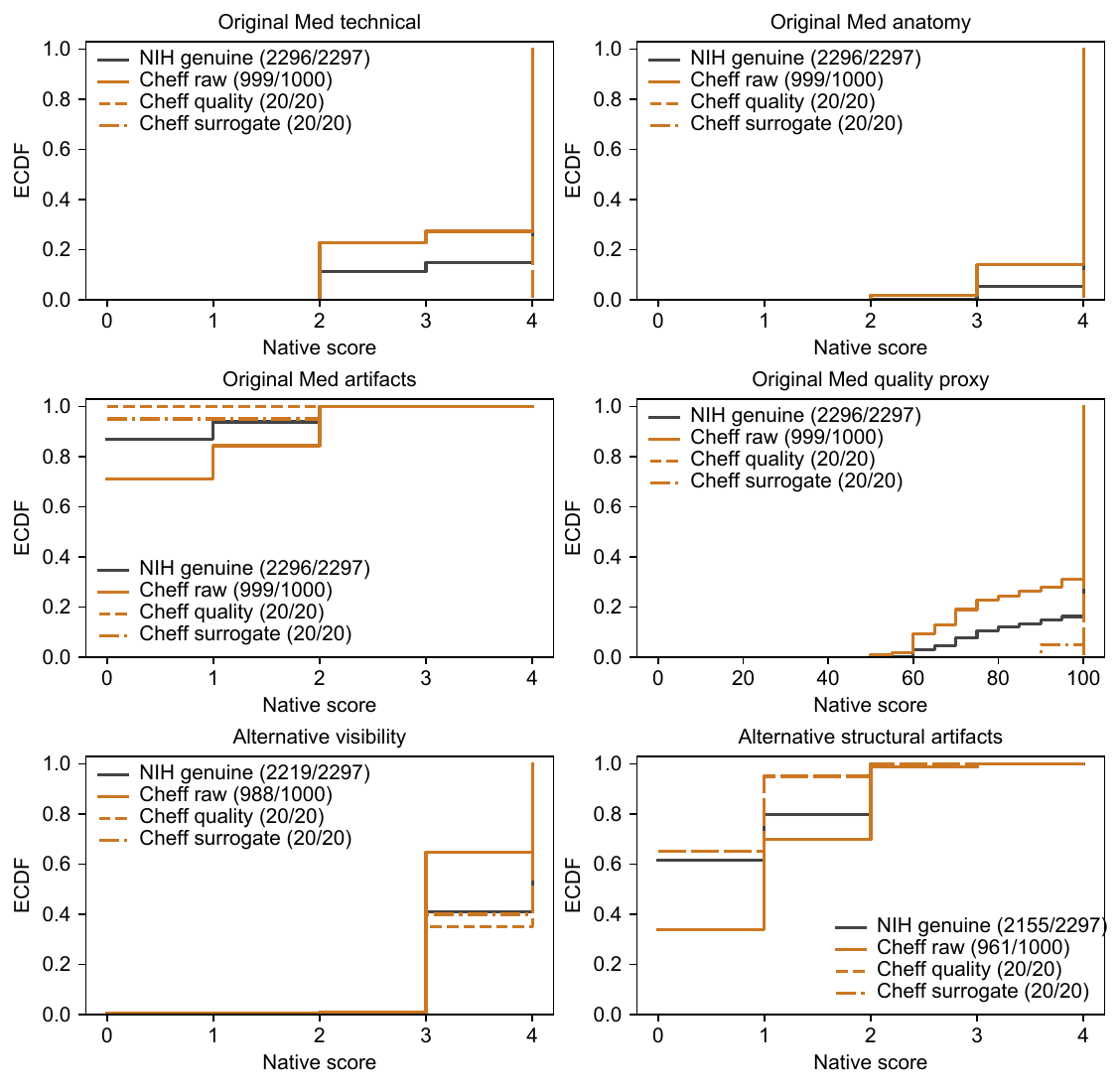}
\caption{Empirical cumulative distributions of canonical quality subscores for Cheff-identified versus genuine NIH radiographs. In the legend, ``raw'' denotes the full uncurated Cheff-identified test split ($N=1{,}000$), while ``quality K20'' and ``surrogate K$\leq$20'' denote the subsets selected at $B=1{,}000$. ``Med'' denotes MedGemma.}
\label{fig:quality-components-cheff}
\end{figure*}

\begin{figure*}[tbp]
\centering
\includegraphics[width=\textwidth]{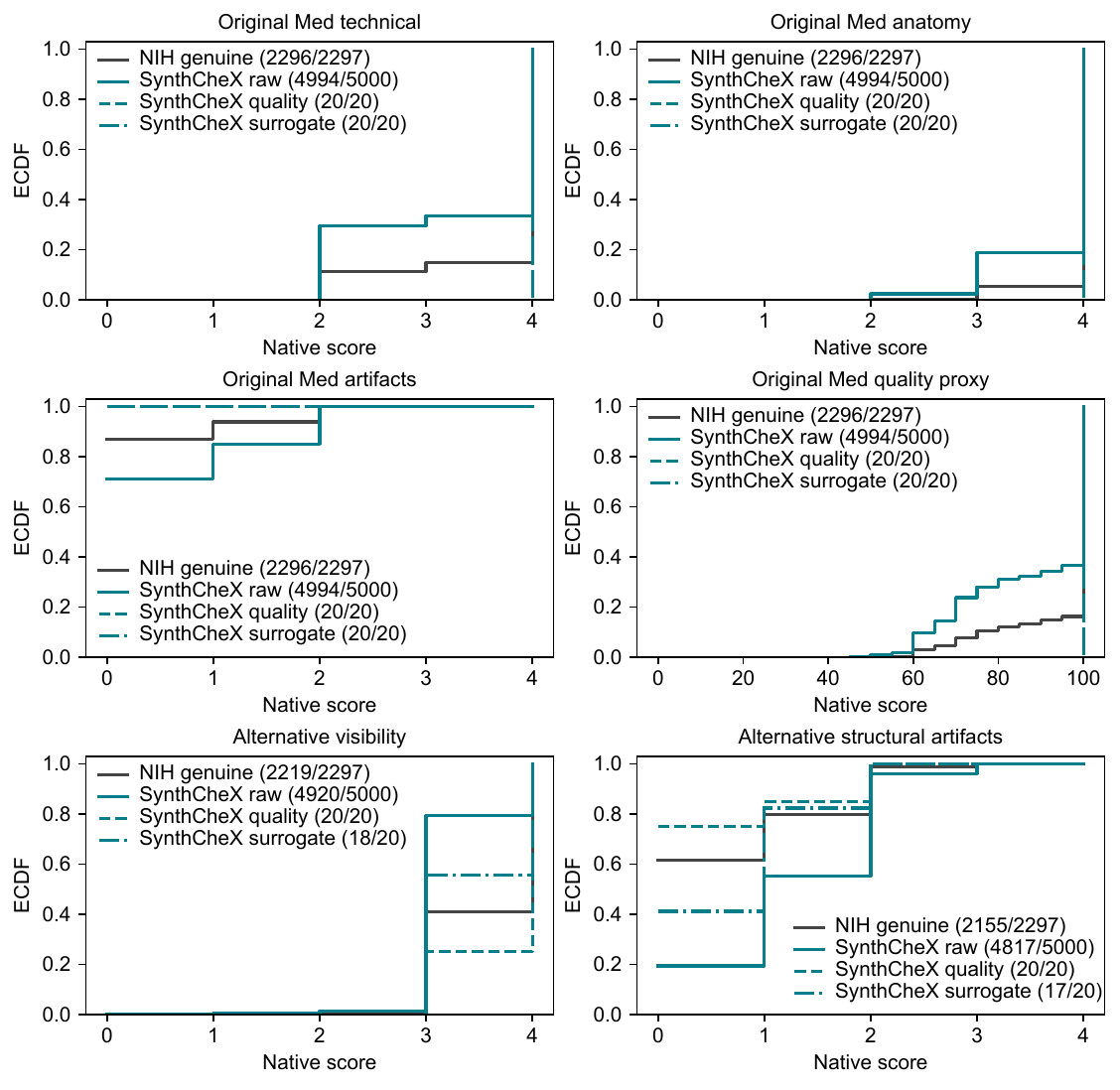}
\caption{Empirical cumulative distributions of canonical quality subscores for SynthCheX versus genuine NIH radiographs, following the conventions of Figure~\ref{fig:quality-components-cheff}.}
\label{fig:quality-components-synth}
\end{figure*}

\begin{figure*}[tbp]
\centering
\includegraphics[width=\textwidth]{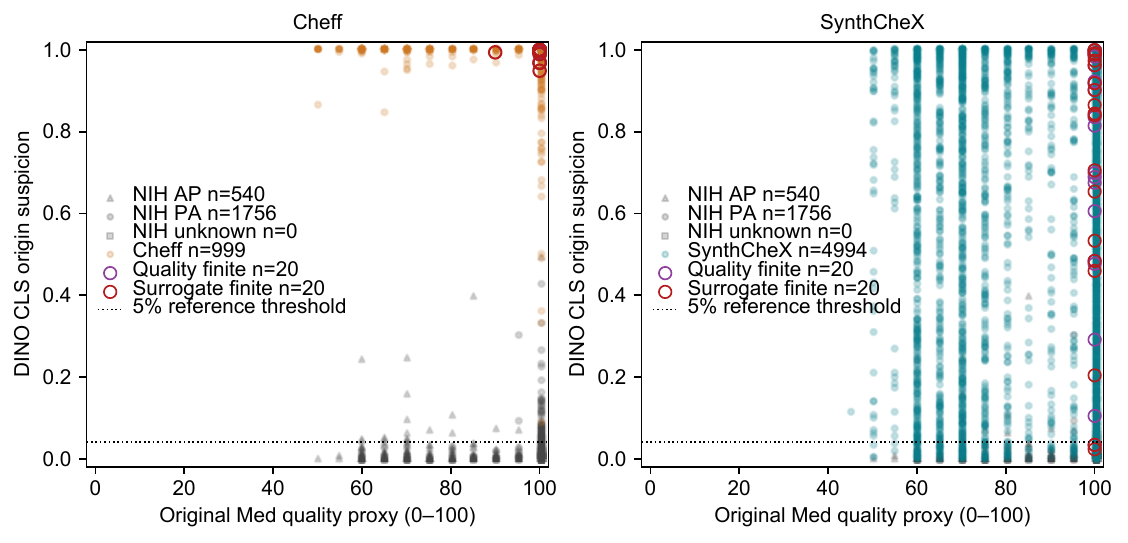}
\caption{Orthogonality of technical quality and predicted origin across the test cohorts. Each point plots a radiograph's MedGemma original-rubric score ($Q\in[0,100]$) against its primary DINOv3 raw CLS origin score (``suspicion''). Circled points highlight candidates selected under quality and surrogate policies at $B=1{,}000, K=20$, and the horizontal dotted line marks the NIH-calibrated $\alpha=5\%$ origin threshold.}
\label{fig:quality-origin}
\end{figure*}

\emph{Question.} What fraction of each test cohort passes each automated quality evaluator, and how do subscore distributions compare between genuine and generated radiographs? Table~\ref{tab:descriptive-quality-evaluators} summarizes test-split performance across all three evaluator configurations: MedGemma under the primary alternative rubric, Qwen under the original rubric, and MedGemma under the original rubric. Table~\ref{tab:independent-selected-quality-proximity} evaluates the 12 fixed $K=20$ submission lists ($B=1{,}000$), reporting both MedGemma pass rates and $\text{P}(\text{genuine below})$—the proportion of genuine NIH radiographs ranked strictly below the selected synthetic submissions on each rubric's native scale. Tables~\ref{tab:native-quality-field-contrasts-1} and~\ref{tab:native-quality-field-contrasts-2} report generated-minus-genuine mean and median subscore differences, while Figures~\ref{fig:quality-components-cheff}--\ref{fig:quality-origin} visualize the empirical cumulative distribution functions (ECDFs) and the joint distribution of quality versus origin scores.

\begin{table*}[tbp]
\centering
\begingroup
\setlength{\tabcolsep}{4pt}
\renewcommand{\arraystretch}{1.15}
\begin{tabular}{@{}lccccccc@{}}
\toprule
Evaluator & Collection & \shortstack{{Attempted}\\{N}} & \shortstack{{Valid}\\{of attempted}} & \shortstack{{Unavailable}\\{of attempted}} & \shortstack{{Unassessable}\\{of valid}} & \shortstack{{Guarded pass}\\{of attempted}} & \shortstack{{Top score}\\{of valid}} \\
\midrule
MedGemma alt. & NIH & 2297 & 2219/2297 & 78/2297 & 2/2219 & 1717/2297 & 1229/2219 \\
MedGemma alt. & Cheff & 1000 & 988/1000 & 12/1000 & 6/988 & 668/1000 & 305/988 \\
MedGemma alt. & SynthCheX & 5000 & 4920/5000 & 80/5000 & 12/4920 & 2656/5000 & 856/4920 \\
Qwen original & NIH & 2297 & 2297/2297 & 0/2297 & 0/2297 & 2059/2297 & 1558/2297 \\
Qwen original & Cheff & 1000 & 1000/1000 & 0/1000 & 1/1000 & 695/1000 & 416/1000 \\
Qwen original & SynthCheX & 5000 & 5000/5000 & 0/5000 & 0/5000 & 3056/5000 & 1238/5000 \\
MedGemma original & NIH & 2297 & 2296/2297 & 1/2297 & 0/2296 & 2116/2297 & 1923/2296 \\
MedGemma original & Cheff & 1000 & 999/1000 & 1/1000 & 0/999 & 809/1000 & 688/999 \\
MedGemma original & SynthCheX & 5000 & 4994/5000 & 6/5000 & 1/4994 & 3806/5000 & 3168/4994 \\
\bottomrule
\end{tabular}
\endgroup
\caption{Test-split performance of all three quality evaluator configurations across collections. Valid, unavailable, and guarded-pass proportions use all attempted images as the denominator; unassessable and top-score counts use schema-valid replies as the denominator. Under the original rubric, ``top score'' denotes $Q=100$; under the alternative rubric, it denotes the optimal ordinal pair ($\texttt{visibility}=4$, $\texttt{structural\_artifact\_severity}=0$).}
\label{tab:descriptive-quality-evaluators}
\end{table*}

\begin{table*}[tbp]
\centering
\begin{tabular}{lllrrrr}
\toprule
Evaluator & Collection & Strategy & \shortstack{{Pass k/n}\\{[95\% CI]}} & \shortstack{{P(genuine below)}\\{[95\% CI]}} & \shortstack{{Ordinal n}\\{genuine /}\\{selected}} & \shortstack{{Unavailable}\\{genuine /}\\{selected}} \\
\midrule
MedGemma alt. & Cheff & Uniform & \shortstack{{12/20}\\{[0.387, 0.781]}} & \shortstack{{0.242}\\{[0.164, 0.317]}} & 2155 / 18 & 78 / 0 \\
MedGemma alt. & Cheff & Quality & \shortstack{{19/20}\\{[0.764, 0.991]}} & \shortstack{{0.337}\\{[0.276, 0.395]}} & 2155 / 20 & 78 / 0 \\
MedGemma alt. & Cheff & Surrogate & \shortstack{{19/20}\\{[0.764, 0.991]}} & \shortstack{{0.333}\\{[0.269, 0.391]}} & 2155 / 20 & 78 / 0 \\
MedGemma alt. & SynthCheX & Uniform & \shortstack{{8/20}\\{[0.219, 0.613]}} & \shortstack{{0.179}\\{[0.098, 0.267]}} & 2155 / 18 & 78 / 1 \\
MedGemma alt. & SynthCheX & Quality & \shortstack{{17/20}\\{[0.640, 0.948]}} & \shortstack{{0.344}\\{[0.271, 0.410]}} & 2155 / 20 & 78 / 0 \\
MedGemma alt. & SynthCheX & Surrogate & \shortstack{{14/20}\\{[0.481, 0.855]}} & \shortstack{{0.258}\\{[0.183, 0.333]}} & 2155 / 17 & 78 / 2 \\
MedGemma orig. & Cheff & Uniform & \shortstack{{14/20}\\{[0.481, 0.855]}} & \shortstack{{0.112}\\{[0.079, 0.144]}} & 2296 / 20 & 1 / 0 \\
MedGemma orig. & Cheff & Quality & \shortstack{{20/20}\\{[0.839, 1.000]}} & \shortstack{{0.162}\\{[0.147, 0.178]}} & 2296 / 20 & 1 / 0 \\
MedGemma orig. & Cheff & Surrogate & \shortstack{{20/20}\\{[0.839, 1.000]}} & \shortstack{{0.161}\\{[0.145, 0.176]}} & 2296 / 20 & 1 / 0 \\
MedGemma orig. & SynthCheX & Uniform & \shortstack{{15/20}\\{[0.531, 0.888]}} & \shortstack{{0.121}\\{[0.094, 0.148]}} & 2296 / 20 & 1 / 0 \\
MedGemma orig. & SynthCheX & Quality & \shortstack{{20/20}\\{[0.839, 1.000]}} & \shortstack{{0.162}\\{[0.147, 0.178]}} & 2296 / 20 & 1 / 0 \\
MedGemma orig. & SynthCheX & Surrogate & \shortstack{{20/20}\\{[0.839, 1.000]}} & \shortstack{{0.162}\\{[0.147, 0.178]}} & 2296 / 20 & 1 / 0 \\
\bottomrule
\end{tabular}
\caption{MedGemma quality evaluations of the 12 fixed submission lists ($B=1{,}000$, $K=20$) across both synthetic collections, three Qwen selection strategies, and both MedGemma rubrics. $\text{P}(\text{genuine below})$ reports the fraction of ordinal-ranked genuine NIH images rated strictly below a selected synthetic submission.}
\label{tab:independent-selected-quality-proximity}
\end{table*}

\begin{table*}[tbp]
\centering
\begin{tabular}{lllrrr}
\toprule
Evaluator & Collection & Rating field & \shortstack{{Mean difference}\\{[95\% CI]}} & \shortstack{{Median difference}\\{[95\% CI]}} & \shortstack{{Finite n}\\{genuine / generated}} \\
\midrule
MedGemma alt. & Cheff & Visibility & \shortstack{{-0.252}\\{[-0.293, -0.213]}} & \shortstack{{-1.000}\\{[-1.000, -1.000]}} & 2219 / 988 \\
MedGemma alt. & Cheff & Artifacts & \shortstack{{0.378}\\{[0.313, 0.443]}} & \shortstack{{1.000}\\{[1.000, 1.000]}} & 2155 / 961 \\
MedGemma alt. & SynthCheX & Visibility & \shortstack{{-0.397}\\{[-0.422, -0.372]}} & \shortstack{{-1.000}\\{[-1.000, -1.000]}} & 2219 / 4920 \\
MedGemma alt. & SynthCheX & Artifacts & \shortstack{{0.693}\\{[0.649, 0.734]}} & \shortstack{{1.000}\\{[1.000, 1.000]}} & 2155 / 4817 \\
Qwen orig. & Cheff & Technical & \shortstack{{-0.484}\\{[-0.546, -0.423]}} & \shortstack{{-1.000}\\{[-1.000, -1.000]}} & 2297 / 1000 \\
Qwen orig. & Cheff & Anatomy & \shortstack{{-0.217}\\{[-0.254, -0.182]}} & \shortstack{{0.000}\\{[0.000, 0.000]}} & 2297 / 1000 \\
Qwen orig. & Cheff & Artifacts & \shortstack{{0.253}\\{[0.212, 0.290]}} & \shortstack{{0.000}\\{[0.000, 0.000]}} & 2297 / 1000 \\
Qwen orig. & Cheff & Score & \shortstack{{-8.279}\\{[-9.392, -7.165]}} & \shortstack{{-10.000}\\{[-10.000, -10.000]}} & 2297 / 1000 \\
Qwen orig. & SynthCheX & Technical & \shortstack{{-0.736}\\{[-0.772, -0.701]}} & \shortstack{{-1.000}\\{[-1.000, -1.000]}} & 2297 / 5000 \\
Qwen orig. & SynthCheX & Anatomy & \shortstack{{-0.306}\\{[-0.325, -0.287]}} & \shortstack{{0.000}\\{[0.000, 0.000]}} & 2297 / 5000 \\
\bottomrule
\end{tabular}
\caption{Generated-minus-genuine differences across individual quality rating fields (Part 1 of 2). Positive differences indicate higher scores on generated images (favorable for visibility, technical quality, anatomical plausibility, and composite score $Q$; unfavorable for artifact severity).}
\label{tab:native-quality-field-contrasts-1}
\end{table*}

\begin{table*}[tbp]
\centering
\begin{tabular}{lllrrr}
\toprule
Evaluator & Collection & Rating field & \shortstack{{Mean difference}\\{[95\% CI]}} & \shortstack{{Median difference}\\{[95\% CI]}} & \shortstack{{Finite n}\\{genuine / generated}} \\
\midrule
Qwen orig. & SynthCheX & Artifacts & \shortstack{{0.328}\\{[0.306, 0.349]}} & \shortstack{{0.000}\\{[0.000, 1.000]}} & 2297 / 5000 \\
Qwen orig. & SynthCheX & Score & \shortstack{{-12.060}\\{[-12.677, -11.443]}} & \shortstack{{-15.000}\\{[-15.000, -10.000]}} & 2297 / 5000 \\
MedGemma orig. & Cheff & Technical & \shortstack{{-0.241}\\{[-0.300, -0.183]}} & \shortstack{{0.000}\\{[0.000, 0.000]}} & 2296 / 999 \\
MedGemma orig. & Cheff & Anatomy & \shortstack{{-0.101}\\{[-0.129, -0.074]}} & \shortstack{{0.000}\\{[0.000, 0.000]}} & 2296 / 999 \\
MedGemma orig. & Cheff & Artifacts & \shortstack{{0.255}\\{[0.203, 0.308]}} & \shortstack{{0.000}\\{[0.000, 0.000]}} & 2296 / 999 \\
MedGemma orig. & Cheff & Score & \shortstack{{-4.697}\\{[-5.722, -3.688]}} & \shortstack{{0.000}\\{[0.000, 0.000]}} & 2296 / 999 \\
MedGemma orig. & SynthCheX & Technical & \shortstack{{-0.370}\\{[-0.406, -0.332]}} & \shortstack{{0.000}\\{[0.000, 0.000]}} & 2296 / 4994 \\
MedGemma orig. & SynthCheX & Anatomy & \shortstack{{-0.156}\\{[-0.172, -0.140]}} & \shortstack{{0.000}\\{[0.000, 0.000]}} & 2296 / 4994 \\
MedGemma orig. & SynthCheX & Artifacts & \shortstack{{0.249}\\{[0.218, 0.279]}} & \shortstack{{0.000}\\{[0.000, 0.000]}} & 2296 / 4994 \\
MedGemma orig. & SynthCheX & Score & \shortstack{{-6.503}\\{[-7.093, -5.903]}} & \shortstack{{0.000}\\{[0.000, 0.000]}} & 2296 / 4994 \\
\bottomrule
\end{tabular}
\caption{Continuation of Table~\ref{tab:native-quality-field-contrasts-1} (Part 2 of 2).}
\label{tab:native-quality-field-contrasts-2}
\end{table*}

\emph{Conclusion.} Across all three evaluator configurations, synthetic cohorts receive lower mean quality ratings and higher artifact severity ratings than genuine NIH radiographs, yet between 53.1\% and 92.1\% of attempted cohorts still pass the binary quality gates. Qwen-guided candidate selection consistently elevates MedGemma pass rates under both rubrics. Crucially, however, even among SynthCheX radiographs achieving a perfect quality score ($Q=100$), DINOv3 origin scores span nearly the entire $[0,1]$ range (Figure~\ref{fig:quality-origin}), confirming that high visual quality does not confer evasion against origin screening.

\subsection{Held-Out Quality Validity and Post-Processing Stability}
\emph{Question.} On held-out test data, do the quality rubrics reject non-radiograph inputs, preserve assessability on clean clinical scans, and rank clean radiographs above synthetically degraded copies? Table~\ref{tab:heldout-quality-ordering} reports pairwise ordering outcomes across the 600 held-out degradation pairs per collection, Table~\ref{tab:heldout-quality-invalid} reports invalid-pattern rejections and clean NIH assessability, Table~\ref{tab:matched-quality-routes} tracks quality passage across four post-processing routes, Table~\ref{tab:heldout-quality-severity} decomposes monotonic ordering chains across degradation families, and Table~\ref{tab:heldout-quality-ceilings} tracks ceiling effects at the top of each rubric's scale.

\begin{table*}[tbp]
\centering
\begin{tabular}{llrrrr}
\toprule
Rubric & Collection & Strict & Tied & Reversed & Unavailable \\
\midrule
Original & NIH & 179 & 370 & 49 & 2 \\
Original & Cheff-identified & 194 & 323 & 83 & 0 \\
Original & SynthCheX & 209 & 295 & 96 & 0 \\
Alternative & NIH & 274 & 207 & 95 & 24 \\
Alternative & Cheff-identified & 234 & 224 & 134 & 8 \\
Alternative & SynthCheX & 142 & 252 & 197 & 9 \\
\bottomrule
\end{tabular}
\caption{Held-out pairwise ordering outcomes across 600 clean--degraded test pairs per collection and rubric. ``Strict'' denotes pairs where the clean radiograph is rated strictly above its degraded copy.}
\label{tab:heldout-quality-ordering}
\end{table*}

\begin{table*}[tbp]
\centering
\begin{tabular}{lrrrr}
\toprule
Rubric & \shortstack{{Invalid pass}\\{No guard (CI)}} & \shortstack{{Invalid pass}\\{Guard (CI)}} & \shortstack{{NIH images without an}\\{assessable judgment}} & \shortstack{{NIH}\\{unavailable}} \\
\midrule
Original & \shortstack{{5/60}\\{[3.61, 18.07]}} & \shortstack{{0/60}\\{[0.00, 6.02]}} & 0/100 & 0/100 \\
Alternative & \shortstack{{0/60}\\{[0.00, 6.02]}} & \shortstack{{0/60}\\{[0.00, 6.02]}} & 3/100 & 3/100 \\
\bottomrule
\end{tabular}
\caption{Held-out false passes on 60 distinct non-radiograph invalid patterns (with Wilson 95\% CIs in percent) and non-assessable/unavailable rates across 100 clean genuine NIH radiographs.}
\label{tab:heldout-quality-invalid}
\end{table*}

\begin{table*}[tbp]
\centering
\begin{tabular}{llrrrr}
\toprule
Rubric & Collection & Canonical & JPEG75 & Resize128 & Native512 \\
\midrule
Original & NIH & 278 (0) & 282 (0) & 278 (0) & 286 (0)\\
Original & Cheff-identified & 241 (1) & 243 (0) & 226 (0) & 248 (0)\\
Original & SynthCheX & 235 (0) & 237 (0) & 232 (0) & 241 (1)\\
Alternative & NIH & 223 (11) & 224 (5) & 235 (2) & 233 (10)\\
Alternative & Cheff-identified & 189 (3) & 196 (4) & 207 (3) & 188 (3)\\
Alternative & SynthCheX & 182 (5) & 169 (8) & 188 (5) & 170 (7)\\
\bottomrule
\end{tabular}
\caption{Guarded quality passes across 300 test radiographs per collection evaluated under four processing conditions (with unavailable response counts in parentheses).}
\label{tab:matched-quality-routes}
\end{table*}

\begin{table*}[tbp]
\centering
\begin{tabular}{lllrrrrrrr}
\toprule
Rubric & Collection & Family & \shortstack{{Expected}\\{chains}} & Valid & \shortstack{{Non-}\\{increasing}} & Strict & Reversed & Tie & Unavailable \\
\midrule
Original & NIH & Blur & 100 & 100 & 85 & 10 & 15 & 77 & 0\\
Original & NIH & Resample & 100 & 100 & 83 & 2 & 17 & 86 & 0\\
Original & NIH & Noise & 100 & 98 & 84 & 17 & 14 & 70 & 2\\
Original & Cheff-identified & Blur & 100 & 100 & 72 & 15 & 28 & 63 & 0\\
Original & Cheff-identified & Resample & 100 & 100 & 73 & 5 & 27 & 78 & 0\\
Original & Cheff-identified & Noise & 100 & 100 & 71 & 10 & 29 & 65 & 0\\
Original & SynthCheX & Blur & 100 & 100 & 51 & 7 & 49 & 49 & 0\\
Original & SynthCheX & Resample & 100 & 100 & 68 & 3 & 32 & 79 & 0\\
Original & SynthCheX & Noise & 100 & 100 & 55 & 14 & 45 & 47 & 0\\
Alternative & NIH & Blur & 100 & 97 & 64 & 5 & 33 & 75 & 3\\
Alternative & NIH & Resample & 100 & 96 & 64 & 6 & 32 & 73 & 4\\
Alternative & NIH & Noise & 100 & 92 & 66 & 27 & 26 & 49 & 8\\
Alternative & Cheff-identified & Blur & 100 & 99 & 58 & 8 & 41 & 74 & 1\\
Alternative & Cheff-identified & Resample & 100 & 96 & 60 & 3 & 36 & 81 & 4\\
Alternative & Cheff-identified & Noise & 100 & 97 & 65 & 11 & 32 & 69 & 3\\
Alternative & SynthCheX & Blur & 100 & 98 & 38 & 4 & 60 & 73 & 2\\
Alternative & SynthCheX & Resample & 100 & 98 & 47 & 0 & 51 & 79 & 2\\
Alternative & SynthCheX & Noise & 100 & 98 & 43 & 1 & 55 & 70 & 2\\
\bottomrule
\end{tabular}
\caption{Held-out monotonic severity-chain analysis across 100 three-step degradation ladders (canonical $\to$ mild $\to$ severe) per collection, rubric, and perturbation family.}
\label{tab:heldout-quality-severity}
\end{table*}

\begin{table*}[tbp]
\centering
\begin{tabular}{llrrrr}
\toprule
Rubric & Collection & \shortstack{{Valid /}\\{attempted}} & \shortstack{{Replies at}\\{top score}} & \shortstack{{Of these,}\\{eligible}} & \shortstack{{Top score /}\\{all valid}} \\
\midrule
Original & NIH & 698/700 & 445 & 445 & 445/698\\
Original & Cheff-identified & 700/700 & 376 & 376 & 376/700\\
Original & SynthCheX & 700/700 & 347 & 347 & 347/700\\
Original & Non-image & 67/70 & 0 & 0 & 0/67\\
Alternative & NIH & 691/700 & 155 & 155 & 155/691\\
Alternative & Cheff-identified & 692/700 & 105 & 105 & 105/692\\
Alternative & SynthCheX & 696/700 & 55 & 55 & 55/696\\
Alternative & Non-image & 62/70 & 0 & 0 & 0/62\\
\bottomrule
\end{tabular}
\caption{Frequency of ceiling scores across the 700 held-out radiograph exposures per collection and 70 non-radiograph control exposures.}
\label{tab:heldout-quality-ceilings}
\end{table*}

\emph{Conclusion.} On held-out data, the alternative rubric rejects all $60/60$ non-radiograph control patterns even without the flat-image guard (vs.\ $5/60$ false passes for the original rubric, reduced to $0/60$ with the guard). However, the alternative rubric exhibits clear trade-offs on radiograph pairs: it fails to return a valid response on $3/100$ clean NIH images (vs.\ $0/100$), yields more preference reversals ($426$ vs.\ $228$ of 1,800 pairs) and unavailable pairs ($41$ vs.\ $2$), and improves strict ordering on NIH ($274$ vs.\ $179$) and Cheff-identified pairs ($234$ vs.\ $194$) while declining on SynthCheX ($142$ vs.\ $209$).

\subsection{Origin Screening: Operating Points, Ranking, Cross-Generator Transfer, and Open-I}
\emph{Question.} How do all seven core origin heads and nine secondary transfer heads perform across false-alarm thresholds ($\alpha\in\{1\%, 5\%, 10\%\}$), threshold-free ranking metrics, and out-of-distribution genuine Open-I radiographs? Table~\ref{tab:supp-origin-canonical} reports $\alpha=5\%$ canonical performance across all seven core heads, while Tables~\ref{tab:supp-core-alpha-1-10}--\ref{tab:paired-core-ranking-all} report $\alpha\in\{1\%, 10\%\}$ operating points, AUROC / partial AUC values, and paired ranking contrasts. Tables~\ref{tab:supp-matched-alpha-5}--\ref{tab:paired-matched-ranking-all} report the corresponding metrics for the nine secondary generator-transfer heads (note that in Tables~\ref{tab:supp-matched-alpha-5}--\ref{tab:matched-operating-extended-2}, synthetic columns report unflagged origin passes, the complement of rejections). Finally, Table~\ref{tab:supp-external-genuine} reports false alarms on the 957 genuine Open-I radiographs.

\begin{table*}[tbp]
\centering
\begin{tabular}{lcccc}
\toprule
Origin classifier & NIH flagged & Cheff rejected & SynthCheX rejected & \shortstack{{Unavailable}\\{(NIH/C/S)}} \\
\midrule
Pilot CLS head & 131 (4.8--6.7) & 1,000 (99.6--100.0) & 4,436 (87.8--89.6) & 0/0/0 \\
DINOv3 CLS, raw (primary) & 110 (4.0--5.7) & 1,000 (99.6--100.0) & 4,677 (92.8--94.2) & 0/0/0 \\
DINOv3 mean, raw & 123 (4.5--6.4) & 1,000 (99.6--100.0) & 4,407 (87.2--89.0) & 0/0/0 \\
Medical mean, raw & 107 (3.9--5.6) & 1,000 (99.6--100.0) & 4,673 (92.7--94.1) & 0/0/0 \\
DINOv3 CLS, augmented & 97 (3.5--5.1) & 1,000 (99.6--100.0) & 4,644 (92.1--93.6) & 0/0/0 \\
Rubric fields & 22 (0.6--1.4) & 13 (0.8--2.2) & 99 (1.6--2.4) & 78/12/80 \\
Pixel statistics & 50 (1.7--2.9) & 603 (57.2--63.3) & 2,782 (54.3--57.0) & 0/0/0 \\
\bottomrule
\end{tabular}
\caption{Origin-check performance across all seven core heads on canonical test images at the nominal $\alpha=5\%$ false-alarm rate (with Wilson 95\% CIs in percent).}
\label{tab:supp-origin-canonical}
\end{table*}

\begin{table*}[tbp]
\centering
\begingroup
\setlength{\tabcolsep}{3pt}
\renewcommand{\arraystretch}{1.15}
\begin{tabular}{@{}lcccccc@{}}
\toprule
Method & $\alpha$ (\%) & \shortstack{{NIH reference}\\{N; U}} & \shortstack{{NIH flags}\\{count; \% [95\%]}} & \shortstack{{Cheff rejections}\\{count; \% [95\%]}} & \shortstack{{SynthCheX rejections}\\{count; \% [95\%]}} & \shortstack{{Unavailable}\\{N/C/S}} \\
\midrule
Pilot CLS & 1 & \shortstack{{2914}\\{U=0}} & \shortstack{{39/2297; 1.70}\\{[1.24, 2.31]}} & \shortstack{{997/1000; 99.70}\\{[99.12, 99.90]}} & \shortstack{{3944/5000; 78.88}\\{[77.73, 79.99]}} & 0/0/0 \\
Pilot CLS & 10 & \shortstack{{2914}\\{U=0}} & \shortstack{{237/2297; 10.32}\\{[9.14, 11.63]}} & \shortstack{{1000/1000; 100.00}\\{[99.62, 100.00]}} & \shortstack{{4636/5000; 92.72}\\{[91.97, 93.41]}} & 0/0/0 \\
DINOv3 CLS raw & 1 & \shortstack{{3000}\\{U=0}} & \shortstack{{26/2297; 1.13}\\{[0.77, 1.65]}} & \shortstack{{999/1000; 99.90}\\{[99.44, 99.98]}} & \shortstack{{4339/5000; 86.78}\\{[85.81, 87.69]}} & 0/0/0 \\
DINOv3 CLS raw & 10 & \shortstack{{3000}\\{U=0}} & \shortstack{{200/2297; 8.71}\\{[7.62, 9.93]}} & \shortstack{{1000/1000; 100.00}\\{[99.62, 100.00]}} & \shortstack{{4780/5000; 95.60}\\{[95.00, 96.13]}} & 0/0/0 \\
DINOv3 mean raw & 1 & \shortstack{{3000}\\{U=0}} & \shortstack{{29/2297; 1.26}\\{[0.88, 1.81]}} & \shortstack{{999/1000; 99.90}\\{[99.44, 99.98]}} & \shortstack{{3808/5000; 76.16}\\{[74.96, 77.32]}} & 0/0/0 \\
DINOv3 mean raw & 10 & \shortstack{{3000}\\{U=0}} & \shortstack{{255/2297; 11.10}\\{[9.88, 12.45]}} & \shortstack{{1000/1000; 100.00}\\{[99.62, 100.00]}} & \shortstack{{4631/5000; 92.62}\\{[91.86, 93.31]}} & 0/0/0 \\
Medical mean raw & 1 & \shortstack{{3000}\\{U=0}} & \shortstack{{27/2297; 1.18}\\{[0.81, 1.70]}} & \shortstack{{999/1000; 99.90}\\{[99.44, 99.98]}} & \shortstack{{4210/5000; 84.20}\\{[83.16, 85.18]}} & 0/0/0 \\
Medical mean raw & 10 & \shortstack{{3000}\\{U=0}} & \shortstack{{189/2297; 8.23}\\{[7.17, 9.42]}} & \shortstack{{1000/1000; 100.00}\\{[99.62, 100.00]}} & \shortstack{{4825/5000; 96.50}\\{[95.95, 96.97]}} & 0/0/0 \\
DINOv3 CLS augmented & 1 & \shortstack{{3000}\\{U=0}} & \shortstack{{24/2297; 1.04}\\{[0.70, 1.55]}} & \shortstack{{999/1000; 99.90}\\{[99.44, 99.98]}} & \shortstack{{4337/5000; 86.74}\\{[85.77, 87.65]}} & 0/0/0 \\
DINOv3 CLS augmented & 10 & \shortstack{{3000}\\{U=0}} & \shortstack{{168/2297; 7.31}\\{[6.32, 8.45]}} & \shortstack{{1000/1000; 100.00}\\{[99.62, 100.00]}} & \shortstack{{4773/5000; 95.46}\\{[94.85, 96.00]}} & 0/0/0 \\
Rubric fields & 1 & \shortstack{{2930 (valid)}\\{U=60}} & \shortstack{{22/2297; 0.96}\\{[0.63, 1.45]}} & \shortstack{{13/1000; 1.30}\\{[0.76, 2.21]}} & \shortstack{{99/5000; 1.98}\\{[1.63, 2.40]}} & 78/12/80 \\
Rubric fields & 10 & \shortstack{{2930 (valid)}\\{U=60}} & \shortstack{{258/2297; 11.23}\\{[10.00, 12.59]}} & \shortstack{{197/1000; 19.70}\\{[17.35, 22.28]}} & \shortstack{{1385/5000; 27.70}\\{[26.48, 28.96]}} & 78/12/80 \\
Pixel statistics & 1 & \shortstack{{3000}\\{U=0}} & \shortstack{{17/2297; 0.74}\\{[0.46, 1.18]}} & \shortstack{{376/1000; 37.60}\\{[34.65, 40.64]}} & \shortstack{{1554/5000; 31.08}\\{[29.81, 32.38]}} & 0/0/0 \\
Pixel statistics & 10 & \shortstack{{3000}\\{U=0}} & \shortstack{{83/2297; 3.61}\\{[2.92, 4.46]}} & \shortstack{{670/1000; 67.00}\\{[64.03, 69.84]}} & \shortstack{{3126/5000; 62.52}\\{[61.17, 63.85]}} & 0/0/0 \\
\bottomrule
\end{tabular}
\endgroup
\caption{Canonical origin-check performance across the seven core heads at nominal $\alpha=1\%$ and $\alpha=10\%$ false-alarm rates.}
\label{tab:supp-core-alpha-1-10}
\end{table*}

\begin{table*}[tbp]
\centering
\begingroup
\setlength{\tabcolsep}{4pt}
\renewcommand{\arraystretch}{1.15}
\begin{tabular}{@{}lccccc@{}}
\toprule
Method & Generated collection & \shortstack{{AUROC}\\{[95\%]}} & \shortstack{{Std. partial AUC}\\{[95\%]}} & \shortstack{{Finite N}\\{NIH/generated}} & \shortstack{{Unavailable U}\\{NIH/generated}} \\
\midrule
Pilot CLS & Cheff & \shortstack{{0.9996}\\{[0.9992, 0.9999]}} & \shortstack{{0.9981}\\{[0.9960, 0.9995]}} & 2297/1000 & 0/0 \\
Pilot CLS & SynthCheX & \shortstack{{0.9704}\\{[0.9671, 0.9736]}} & \shortstack{{0.9182}\\{[0.9085, 0.9270]}} & 2297/5000 & 0/0 \\
DINOv3 CLS raw & Cheff & \shortstack{{0.9999}\\{[0.9997, 1.0000]}} & \shortstack{{0.9994}\\{[0.9982, 1.0000]}} & 2297/1000 & 0/0 \\
DINOv3 CLS raw & SynthCheX & \shortstack{{0.9849}\\{[0.9825, 0.9871]}} & \shortstack{{0.9569}\\{[0.9496, 0.9630]}} & 2297/5000 & 0/0 \\
DINOv3 mean raw & Cheff & \shortstack{{0.9997}\\{[0.9993, 1.0000]}} & \shortstack{{0.9986}\\{[0.9962, 0.9999]}} & 2297/1000 & 0/0 \\
DINOv3 mean raw & SynthCheX & \shortstack{{0.9699}\\{[0.9665, 0.9732]}} & \shortstack{{0.9182}\\{[0.9083, 0.9272]}} & 2297/5000 & 0/0 \\
Medical mean raw & Cheff & \shortstack{{0.9999}\\{[0.9999, 1.0000]}} & \shortstack{{0.9997}\\{[0.9994, 0.9999]}} & 2297/1000 & 0/0 \\
Medical mean raw & SynthCheX & \shortstack{{0.9891}\\{[0.9874, 0.9906]}} & \shortstack{{0.9573}\\{[0.9507, 0.9632]}} & 2297/5000 & 0/0 \\
DINOv3 CLS augmented & Cheff & \shortstack{{0.9999}\\{[0.9997, 1.0000]}} & \shortstack{{0.9994}\\{[0.9984, 1.0000]}} & 2297/1000 & 0/0 \\
DINOv3 CLS augmented & SynthCheX & \shortstack{{0.9864}\\{[0.9843, 0.9883]}} & \shortstack{{0.9581}\\{[0.9515, 0.9643]}} & 2297/5000 & 0/0 \\
Rubric fields & Cheff & \shortstack{{0.6301}\\{[0.6098, 0.6493]}} & \shortstack{{0.5196}\\{[0.5112, 0.5286]}} & 2219/988 & 78/12 \\
Rubric fields & SynthCheX & \shortstack{{0.7181}\\{[0.7049, 0.7314]}} & \shortstack{{0.5341}\\{[0.5262, 0.5430]}} & 2219/4920 & 78/80 \\
Pixel statistics & Cheff & \shortstack{{0.9321}\\{[0.9224, 0.9408]}} & \shortstack{{0.8118}\\{[0.7914, 0.8325]}} & 2297/1000 & 0/0 \\
Pixel statistics & SynthCheX & \shortstack{{0.8962}\\{[0.8884, 0.9031]}} & \shortstack{{0.7908}\\{[0.7739, 0.8064]}} & 2297/5000 & 0/0 \\
\bottomrule
\end{tabular}
\endgroup
\caption{Full-range AUROC and standardized partial AUC ($\text{FPR}\in[0, 0.10]$) on canonical test radiographs across the seven core heads.}
\label{tab:supp-canonical-auc}
\end{table*}

\begin{table*}[tbp]
\centering
\begin{tabular}{llrrrr}
\toprule
\shortstack{{First minus}\\{second}} & Collection & \shortstack{{AUROC difference}\\{[95\% CI]}} & \shortstack{{Std. partial AUC difference}\\{[95\% CI]}} & \shortstack{{Finite in both}\\{NIH / gen.}} & \shortstack{{Not finite}\\{NIH / gen.}} \\
\midrule
\shortstack{{DINOv3 mean}\\{$-$ DINOv3 CLS}} & Cheff & \shortstack{{\ensuremath{-0.000156726}}\\{[\ensuremath{-0.000404495},}\\{\ensuremath{-6.51937\times10^{-6}}]}} & \shortstack{{\ensuremath{-0.000824875}}\\{[\ensuremath{-0.00212892},}\\{\ensuremath{-3.43125\times10^{-5}}]}} & 2297 / 1000 & 0 / 0 \\
\shortstack{{DINOv3 mean}\\{$-$ DINOv3 CLS}} & SynthCheX & \shortstack{{\ensuremath{-0.0149568}}\\{[\ensuremath{-0.0173688},}\\{\ensuremath{-0.012378}]}} & \shortstack{{\ensuremath{-0.0386599}}\\{[\ensuremath{-0.0456914},}\\{\ensuremath{-0.0320829}]}} & 2297 / 5000 & 0 / 0 \\
\shortstack{{Medical mean}\\{$-$ DINOv3 CLS}} & Cheff & \shortstack{{\ensuremath{6.53026\times10^{-5}}}\\{[\ensuremath{-6.66195\times10^{-5}},}\\{\ensuremath{0.000289987}]}} & \shortstack{{\ensuremath{0.000343698}}\\{[\ensuremath{-0.000350629},}\\{\ensuremath{0.00152625}]}} & 2297 / 1000 & 0 / 0 \\
\shortstack{{Medical mean}\\{$-$ DINOv3 CLS}} & SynthCheX & \shortstack{{\ensuremath{0.00422125}}\\{[\ensuremath{0.0017536},}\\{\ensuremath{0.00678711}]}} & \shortstack{{\ensuremath{0.000384208}}\\{[\ensuremath{-0.0074951},}\\{\ensuremath{0.00813689}]}} & 2297 / 5000 & 0 / 0 \\
\shortstack{{DINOv3 CLS aug.}\\{$-$ DINOv3 CLS}} & Cheff & \shortstack{{\ensuremath{4.78886\times10^{-6}}}\\{[\ensuremath{-1.8731\times10^{-5}},}\\{\ensuremath{3.96169\times10^{-5}}]}} & \shortstack{{\ensuremath{2.52045\times10^{-5}}}\\{[\ensuremath{-9.8584\times10^{-5}},}\\{\ensuremath{0.00020851}]}} & 2297 / 1000 & 0 / 0 \\
\shortstack{{DINOv3 CLS aug.}\\{$-$ DINOv3 CLS}} & SynthCheX & \shortstack{{\ensuremath{0.00152634}}\\{[\ensuremath{0.000202819},}\\{\ensuremath{0.00294557}]}} & \shortstack{{\ensuremath{0.00124176}}\\{[\ensuremath{-0.0022921},}\\{\ensuremath{0.00470716}]}} & 2297 / 5000 & 0 / 0 \\
\shortstack{{Rubric fields}\\{$-$ DINOv3 CLS}} & Cheff & \shortstack{{\ensuremath{-0.369806}}\\{[\ensuremath{-0.389266},}\\{\ensuremath{-0.350225}]}} & \shortstack{{\ensuremath{-0.479743}}\\{[\ensuremath{-0.488167},}\\{\ensuremath{-0.470702}]}} & 2219 / 988 & 78 / 12 \\
\shortstack{{Rubric fields}\\{$-$ DINOv3 CLS}} & SynthCheX & \shortstack{{\ensuremath{-0.266956}}\\{[\ensuremath{-0.280416},}\\{\ensuremath{-0.253566}]}} & \shortstack{{\ensuremath{-0.422904}}\\{[\ensuremath{-0.433036},}\\{\ensuremath{-0.411494}]}} & 2219 / 4920 & 78 / 80 \\
\shortstack{{Rubric fields}\\{$-$ Medical mean}} & Cheff & \shortstack{{\ensuremath{-0.369872}}\\{[\ensuremath{-0.389395},}\\{\ensuremath{-0.35026}]}} & \shortstack{{\ensuremath{-0.480093}}\\{[\ensuremath{-0.488404},}\\{\ensuremath{-0.471087}]}} & 2219 / 988 & 78 / 12 \\
\shortstack{{Rubric fields}\\{$-$ Medical mean}} & SynthCheX & \shortstack{{\ensuremath{-0.270818}}\\{[\ensuremath{-0.283639},}\\{\ensuremath{-0.257932}]}} & \shortstack{{\ensuremath{-0.422774}}\\{[\ensuremath{-0.432709},}\\{\ensuremath{-0.411797}]}} & 2219 / 4920 & 78 / 80 \\
\shortstack{{Pixel statistics}\\{$-$ DINOv3 CLS}} & Cheff & \shortstack{{\ensuremath{-0.0677423}}\\{[\ensuremath{-0.0774572},}\\{\ensuremath{-0.0590053}]}} & \shortstack{{\ensuremath{-0.18758}}\\{[\ensuremath{-0.207919},}\\{\ensuremath{-0.166761}]}} & 2297 / 1000 & 0 / 0 \\
\shortstack{{Pixel statistics}\\{$-$ DINOv3 CLS}} & SynthCheX & \shortstack{{\ensuremath{-0.0887234}}\\{[\ensuremath{-0.0966377},}\\{\ensuremath{-0.081393}]}} & \shortstack{{\ensuremath{-0.166057}}\\{[\ensuremath{-0.183759},}\\{\ensuremath{-0.149282}]}} & 2297 / 5000 & 0 / 0 \\
\bottomrule
\end{tabular}
\caption{Paired bootstrap differences in AUROC and standardized partial AUC between core heads (first head minus second head) over images with finite scores under both heads.}
\label{tab:paired-core-ranking-all}
\end{table*}

\begin{table*}[tbp]
\centering
\begingroup
\setlength{\tabcolsep}{4pt}
\renewcommand{\arraystretch}{1.15}
\begin{tabular}{@{}lccccc@{}}
\toprule
Method & \shortstack{{NIH reference}\\{N; U}} & \shortstack{{NIH flags}\\{count; \% [95\%]}} & \shortstack{{Cheff passes}\\{origin check}\\{count; \% [95\%]}} & \shortstack{{SynthCheX passes}\\{origin check}\\{count; \% [95\%]}} & \shortstack{{Unavailable}\\{N/C/S}} \\
\midrule
CLS / Cheff & \shortstack{{3000}\\{U=0}} & \shortstack{{101/2297; 4.40}\\{[3.63, 5.31]}} & \shortstack{{0/1000; 0.00}\\{[0.00, 0.38]}} & \shortstack{{311/5000; 6.22}\\{[5.58, 6.92]}} & 0/0/0 \\
CLS / SynthCheX & \shortstack{{3000}\\{U=0}} & \shortstack{{165/2297; 7.18}\\{[6.20, 8.31]}} & \shortstack{{114/1000; 11.40}\\{[9.58, 13.52]}} & \shortstack{{0/5000; 0.00}\\{[0.00, 0.08]}} & 0/0/0 \\
CLS / Mixed & \shortstack{{3000}\\{U=0}} & \shortstack{{134/2297; 5.83}\\{[4.95, 6.87]}} & \shortstack{{0/1000; 0.00}\\{[0.00, 0.38]}} & \shortstack{{0/5000; 0.00}\\{[0.00, 0.08]}} & 0/0/0 \\
Mean / Cheff & \shortstack{{3000}\\{U=0}} & \shortstack{{114/2297; 4.96}\\{[4.15, 5.93]}} & \shortstack{{0/1000; 0.00}\\{[0.00, 0.38]}} & \shortstack{{673/5000; 13.46}\\{[12.54, 14.43]}} & 0/0/0 \\
Mean / SynthCheX & \shortstack{{3000}\\{U=0}} & \shortstack{{135/2297; 5.88}\\{[4.99, 6.91]}} & \shortstack{{177/1000; 17.70}\\{[15.46, 20.19]}} & \shortstack{{0/5000; 0.00}\\{[0.00, 0.08]}} & 0/0/0 \\
Mean / Mixed & \shortstack{{3000}\\{U=0}} & \shortstack{{162/2297; 7.05}\\{[6.08, 8.17]}} & \shortstack{{4/1000; 0.40}\\{[0.16, 1.02]}} & \shortstack{{1/5000; 0.02}\\{[0.00, 0.11]}} & 0/0/0 \\
Medical / Cheff & \shortstack{{3000}\\{U=0}} & \shortstack{{91/2297; 3.96}\\{[3.24, 4.84]}} & \shortstack{{0/1000; 0.00}\\{[0.00, 0.38]}} & \shortstack{{552/5000; 11.04}\\{[10.20, 11.94]}} & 0/0/0 \\
Medical / SynthCheX & \shortstack{{3000}\\{U=0}} & \shortstack{{188/2297; 8.18}\\{[7.13, 9.38]}} & \shortstack{{82/1000; 8.20}\\{[6.66, 10.06]}} & \shortstack{{0/5000; 0.00}\\{[0.00, 0.08]}} & 0/0/0 \\
Medical / Mixed & \shortstack{{3000}\\{U=0}} & \shortstack{{126/2297; 5.49}\\{[4.63, 6.49]}} & \shortstack{{0/1000; 0.00}\\{[0.00, 0.38]}} & \shortstack{{0/5000; 0.00}\\{[0.00, 0.08]}} & 0/0/0 \\
\bottomrule
\end{tabular}
\endgroup
\caption{Secondary generator-transfer heads evaluated at the nominal $\alpha=5\%$ false-alarm threshold. Synthetic columns report unflagged origin passes (the complement of rejections).}
\label{tab:supp-matched-alpha-5}
\end{table*}

\begin{table*}[tbp]
\centering
\begin{tabular}{lrrrrr}
\toprule
Head & \shortstack{{Nominal}\\{$\alpha$ (\%)}} & \shortstack{{NIH flags}\\{[95\% CI]}} & \shortstack{{Cheff passes}\\{origin check}\\{[95\% CI]}} & \shortstack{{SynthCheX passes}\\{origin check}\\{[95\% CI]}} & \shortstack{{Reference n;}\\{unavailable}\\{NIH/Cheff/SynthCheX}} \\
\midrule
CLS / Cheff & 1 & \shortstack{{20/2297}\\{[0.56, 1.34]}} & \shortstack{{1/1000}\\{[0.02, 0.56]}} & \shortstack{{695/5000}\\{[12.97, 14.89]}} & \shortstack{{Ref. 3000}\\{U: 0 / 0 / 0}} \\
CLS / Cheff & 10 & \shortstack{{215/2297}\\{[8.24, 10.62]}} & \shortstack{{0/1000}\\{[0.00, 0.38]}} & \shortstack{{196/5000}\\{[3.42, 4.49]}} & \shortstack{{Ref. 3000}\\{U: 0 / 0 / 0}} \\
CLS / SynthCheX & 1 & \shortstack{{33/2297}\\{[1.02, 2.01]}} & \shortstack{{213/1000}\\{[18.87, 23.94]}} & \shortstack{{0/5000}\\{[0.00, 0.08]}} & \shortstack{{Ref. 3000}\\{U: 0 / 0 / 0}} \\
CLS / SynthCheX & 10 & \shortstack{{333/2297}\\{[13.12, 16.00]}} & \shortstack{{69/1000}\\{[5.49, 8.64]}} & \shortstack{{0/5000}\\{[0.00, 0.08]}} & \shortstack{{Ref. 3000}\\{U: 0 / 0 / 0}} \\
CLS / Mixed & 1 & \shortstack{{19/2297}\\{[0.53, 1.29]}} & \shortstack{{2/1000}\\{[0.05, 0.73]}} & \shortstack{{2/5000}\\{[0.01, 0.15]}} & \shortstack{{Ref. 3000}\\{U: 0 / 0 / 0}} \\
CLS / Mixed & 10 & \shortstack{{232/2297}\\{[8.93, 11.40]}} & \shortstack{{0/1000}\\{[0.00, 0.38]}} & \shortstack{{0/5000}\\{[0.00, 0.08]}} & \shortstack{{Ref. 3000}\\{U: 0 / 0 / 0}} \\
Mean / Cheff & 1 & \shortstack{{32/2297}\\{[0.99, 1.96]}} & \shortstack{{3/1000}\\{[0.10, 0.88]}} & \shortstack{{1314/5000}\\{[25.08, 27.52]}} & \shortstack{{Ref. 3000}\\{U: 0 / 0 / 0}} \\
Mean / Cheff & 10 & \shortstack{{255/2297}\\{[9.88, 12.45]}} & \shortstack{{0/1000}\\{[0.00, 0.38]}} & \shortstack{{412/5000}\\{[7.51, 9.03]}} & \shortstack{{Ref. 3000}\\{U: 0 / 0 / 0}} \\
Mean / SynthCheX & 1 & \shortstack{{32/2297}\\{[0.99, 1.96]}} & \shortstack{{269/1000}\\{[24.24, 29.73]}} & \shortstack{{0/5000}\\{[0.00, 0.08]}} & \shortstack{{Ref. 3000}\\{U: 0 / 0 / 0}} \\
\bottomrule
\end{tabular}
\caption{Secondary generator-transfer heads evaluated at nominal $\alpha=1\%$ and $\alpha=10\%$ thresholds (Part 1 of 2). Synthetic columns report unflagged origin passes.}
\label{tab:matched-operating-extended-1}
\end{table*}

\begin{table*}[tbp]
\centering
\begin{tabular}{lrrrrr}
\toprule
Head & \shortstack{{Nominal}\\{$\alpha$ (\%)}} & \shortstack{{NIH flags}\\{[95\% CI]}} & \shortstack{{Cheff passes}\\{origin check}\\{[95\% CI]}} & \shortstack{{SynthCheX passes}\\{origin check}\\{[95\% CI]}} & \shortstack{{Reference n;}\\{unavailable}\\{NIH/Cheff/SynthCheX}} \\
\midrule
Mean / SynthCheX & 10 & \shortstack{{300/2297}\\{[11.74, 14.50]}} & \shortstack{{119/1000}\\{[10.04, 14.05]}} & \shortstack{{0/5000}\\{[0.00, 0.08]}} & \shortstack{{Ref. 3000}\\{U: 0 / 0 / 0}} \\
Mean / Mixed & 1 & \shortstack{{34/2297}\\{[1.06, 2.06]}} & \shortstack{{15/1000}\\{[0.91, 2.46]}} & \shortstack{{9/5000}\\{[0.09, 0.34]}} & \shortstack{{Ref. 3000}\\{U: 0 / 0 / 0}} \\
Mean / Mixed & 10 & \shortstack{{299/2297}\\{[11.70, 14.46]}} & \shortstack{{1/1000}\\{[0.02, 0.56]}} & \shortstack{{0/5000}\\{[0.00, 0.08]}} & \shortstack{{Ref. 3000}\\{U: 0 / 0 / 0}} \\
Medical / Cheff & 1 & \shortstack{{26/2297}\\{[0.77, 1.65]}} & \shortstack{{1/1000}\\{[0.02, 0.56]}} & \shortstack{{1123/5000}\\{[21.32, 23.64]}} & \shortstack{{Ref. 3000}\\{U: 0 / 0 / 0}} \\
Medical / Cheff & 10 & \shortstack{{188/2297}\\{[7.13, 9.38]}} & \shortstack{{0/1000}\\{[0.00, 0.38]}} & \shortstack{{301/5000}\\{[5.39, 6.71]}} & \shortstack{{Ref. 3000}\\{U: 0 / 0 / 0}} \\
Medical / SynthCheX & 1 & \shortstack{{57/2297}\\{[1.92, 3.20]}} & \shortstack{{149/1000}\\{[12.83, 17.24]}} & \shortstack{{1/5000}\\{[0.00, 0.11]}} & \shortstack{{Ref. 3000}\\{U: 0 / 0 / 0}} \\
Medical / SynthCheX & 10 & \shortstack{{311/2297}\\{[12.20, 15.00]}} & \shortstack{{52/1000}\\{[3.99, 6.76]}} & \shortstack{{0/5000}\\{[0.00, 0.08]}} & \shortstack{{Ref. 3000}\\{U: 0 / 0 / 0}} \\
Medical / Mixed & 1 & \shortstack{{33/2297}\\{[1.02, 2.01]}} & \shortstack{{3/1000}\\{[0.10, 0.88]}} & \shortstack{{0/5000}\\{[0.00, 0.08]}} & \shortstack{{Ref. 3000}\\{U: 0 / 0 / 0}} \\
Medical / Mixed & 10 & \shortstack{{227/2297}\\{[8.73, 11.17]}} & \shortstack{{0/1000}\\{[0.00, 0.38]}} & \shortstack{{0/5000}\\{[0.00, 0.08]}} & \shortstack{{Ref. 3000}\\{U: 0 / 0 / 0}} \\
\bottomrule
\end{tabular}
\caption{Continuation of Table~\ref{tab:matched-operating-extended-1} (Part 2 of 2).}
\label{tab:matched-operating-extended-2}
\end{table*}

\begin{table*}[tbp]
\centering
\begin{tabular}{llrrrr}
\toprule
Head & Collection & \shortstack{{AUROC}\\{[95\% CI]}} & \shortstack{{Std. partial AUC}\\{[95\% CI]}} & \shortstack{{Finite n}\\{NIH / gen.}} & \shortstack{{Unavailable n}\\{NIH / gen.}} \\
\midrule
CLS / Cheff & Cheff & \shortstack{{\ensuremath{0.999872}}\\{[\ensuremath{0.99964}, \ensuremath{0.999998}]}} & \shortstack{{\ensuremath{0.999329}}\\{[\ensuremath{0.998107}, \ensuremath{0.999989}]}} & 2297 / 1000 & 0 / 0 \\
CLS / Cheff & SynthCheX & \shortstack{{\ensuremath{0.986047}}\\{[\ensuremath{0.983776}, \ensuremath{0.988085}]}} & \shortstack{{\ensuremath{0.959035}}\\{[\ensuremath{0.951916}, \ensuremath{0.964967}]}} & 2297 / 5000 & 0 / 0 \\
CLS / SynthCheX & Cheff & \shortstack{{\ensuremath{0.966237}}\\{[\ensuremath{0.959058}, \ensuremath{0.973158}]}} & \shortstack{{\ensuremath{0.919539}}\\{[\ensuremath{0.907532}, \ensuremath{0.931643}]}} & 2297 / 1000 & 0 / 0 \\
CLS / SynthCheX & SynthCheX & \shortstack{{\ensuremath{0.999996}}\\{[\ensuremath{0.99999}, \ensuremath{1}]}} & \shortstack{{\ensuremath{0.99998}}\\{[\ensuremath{0.99995}, \ensuremath{0.999998}]}} & 2297 / 5000 & 0 / 0 \\
CLS / Mixed & Cheff & \shortstack{{\ensuremath{0.99971}}\\{[\ensuremath{0.999378}, \ensuremath{0.999929}]}} & \shortstack{{\ensuremath{0.998474}}\\{[\ensuremath{0.996728}, \ensuremath{0.999629}]}} & 2297 / 1000 & 0 / 0 \\
CLS / Mixed & SynthCheX & \shortstack{{\ensuremath{0.999792}}\\{[\ensuremath{0.999399}, \ensuremath{0.99999}]}} & \shortstack{{\ensuremath{0.998905}}\\{[\ensuremath{0.996837}, \ensuremath{0.999949}]}} & 2297 / 5000 & 0 / 0 \\
Mean / Cheff & Cheff & \shortstack{{\ensuremath{0.999685}}\\{[\ensuremath{0.99924}, \ensuremath{0.999954}]}} & \shortstack{{\ensuremath{0.998341}}\\{[\ensuremath{0.996001}, \ensuremath{0.999757}]}} & 2297 / 1000 & 0 / 0 \\
Mean / Cheff & SynthCheX & \shortstack{{\ensuremath{0.966726}}\\{[\ensuremath{0.963128}, \ensuremath{0.970131}]}} & \shortstack{{\ensuremath{0.908285}}\\{[\ensuremath{0.897911}, \ensuremath{0.917496}]}} & 2297 / 5000 & 0 / 0 \\
Mean / SynthCheX & Cheff & \shortstack{{\ensuremath{0.951915}}\\{[\ensuremath{0.943813}, \ensuremath{0.959914}]}} & \shortstack{{\ensuremath{0.88998}}\\{[\ensuremath{0.875868}, \ensuremath{0.903272}]}} & 2297 / 1000 & 0 / 0 \\
\bottomrule
\end{tabular}
\caption{Full-range AUROC and standardized partial AUC ($\text{FPR}\in[0, 0.10]$) for the nine secondary generator-transfer heads on canonical test images (Part 1 of 2).}
\label{tab:matched-ranking-all-1}
\end{table*}

\begin{table*}[tbp]
\centering
\begin{tabular}{llrrrr}
\toprule
Head & Collection & \shortstack{{AUROC}\\{[95\% CI]}} & \shortstack{{Std. partial AUC}\\{[95\% CI]}} & \shortstack{{Finite n}\\{NIH / gen.}} & \shortstack{{Unavailable n}\\{NIH / gen.}} \\
\midrule
Mean / SynthCheX & SynthCheX & \shortstack{{\ensuremath{0.999975}}\\{[\ensuremath{0.999955}, \ensuremath{0.999991}]}} & \shortstack{{\ensuremath{0.999867}}\\{[\ensuremath{0.999761}, \ensuremath{0.99995}]}} & 2297 / 5000 & 0 / 0 \\
Mean / Mixed & Cheff & \shortstack{{\ensuremath{0.998768}}\\{[\ensuremath{0.998087}, \ensuremath{0.999353}]}} & \shortstack{{\ensuremath{0.993942}}\\{[\ensuremath{0.990759}, \ensuremath{0.99671}]}} & 2297 / 1000 & 0 / 0 \\
Mean / Mixed & SynthCheX & \shortstack{{\ensuremath{0.999619}}\\{[\ensuremath{0.999124}, \ensuremath{0.999914}]}} & \shortstack{{\ensuremath{0.998005}}\\{[\ensuremath{0.995409}, \ensuremath{0.999554}]}} & 2297 / 5000 & 0 / 0 \\
Medical / Cheff & Cheff & \shortstack{{\ensuremath{0.999897}}\\{[\ensuremath{0.99981}, \ensuremath{0.999963}]}} & \shortstack{{\ensuremath{0.999459}}\\{[\ensuremath{0.998999}, \ensuremath{0.999805}]}} & 2297 / 1000 & 0 / 0 \\
Medical / Cheff & SynthCheX & \shortstack{{\ensuremath{0.982348}}\\{[\ensuremath{0.979953}, \ensuremath{0.984484}]}} & \shortstack{{\ensuremath{0.935143}}\\{[\ensuremath{0.926342}, \ensuremath{0.943188}]}} & 2297 / 5000 & 0 / 0 \\
Medical / SynthCheX & Cheff & \shortstack{{\ensuremath{0.979815}}\\{[\ensuremath{0.975402}, \ensuremath{0.983743}]}} & \shortstack{{\ensuremath{0.935848}}\\{[\ensuremath{0.924825}, \ensuremath{0.945991}]}} & 2297 / 1000 & 0 / 0 \\
Medical / SynthCheX & SynthCheX & \shortstack{{\ensuremath{0.999948}}\\{[\ensuremath{0.999911}, \ensuremath{0.999976}]}} & \shortstack{{\ensuremath{0.999725}}\\{[\ensuremath{0.999529}, \ensuremath{0.999872}]}} & 2297 / 5000 & 0 / 0 \\
Medical / Mixed & Cheff & \shortstack{{\ensuremath{0.99975}}\\{[\ensuremath{0.999576}, \ensuremath{0.99989}]}} & \shortstack{{\ensuremath{0.998685}}\\{[\ensuremath{0.997768}, \ensuremath{0.99942}]}} & 2297 / 1000 & 0 / 0 \\
Medical / Mixed & SynthCheX & \shortstack{{\ensuremath{0.999945}}\\{[\ensuremath{0.999886}, \ensuremath{0.999982}]}} & \shortstack{{\ensuremath{0.999713}}\\{[\ensuremath{0.999401}, \ensuremath{0.999905}]}} & 2297 / 5000 & 0 / 0 \\
\bottomrule
\end{tabular}
\caption{Continuation of Table~\ref{tab:matched-ranking-all-1} (Part 2 of 2).}
\label{tab:matched-ranking-all-2}
\end{table*}

\begin{table*}[tbp]
\centering
\begin{tabular}{llrrrr}
\toprule
\shortstack{{First minus}\\{second}} & Collection & \shortstack{{AUROC difference}\\{[95\% CI]}} & \shortstack{{Std. partial AUC difference}\\{[95\% CI]}} & \shortstack{{Finite in both}\\{NIH / gen.}} & \shortstack{{Not finite}\\{NIH / gen.}} \\
\midrule
\shortstack{{CLS / SynthCheX}\\{$-$ CLS / Cheff}} & Cheff & \shortstack{{\ensuremath{-0.0336352}}\\{[\ensuremath{-0.0407283},}\\{\ensuremath{-0.0267199}]}} & \shortstack{{\ensuremath{-0.0797901}}\\{[\ensuremath{-0.0918953},}\\{\ensuremath{-0.0676765}]}} & 2297 / 1000 & 0 / 0 \\
\shortstack{{CLS / SynthCheX}\\{$-$ CLS / Cheff}} & SynthCheX & \shortstack{{\ensuremath{0.013949}}\\{[\ensuremath{0.0119077},}\\{\ensuremath{0.0162161}]}} & \shortstack{{\ensuremath{0.0409448}}\\{[\ensuremath{0.0350172},}\\{\ensuremath{0.0480639}]}} & 2297 / 5000 & 0 / 0 \\
\shortstack{{CLS / Mixed}\\{$-$ CLS / Cheff}} & Cheff & \shortstack{{\ensuremath{-0.000162386}}\\{[\ensuremath{-0.000311276},}\\{\ensuremath{-4.22181\times10^{-5}}]}} & \shortstack{{\ensuremath{-0.000854662}}\\{[\ensuremath{-0.00163829},}\\{\ensuremath{-0.000222201}]}} & 2297 / 1000 & 0 / 0 \\
\shortstack{{CLS / Mixed}\\{$-$ CLS / Cheff}} & SynthCheX & \shortstack{{\ensuremath{0.0137448}}\\{[\ensuremath{0.01182},}\\{\ensuremath{0.0159114}]}} & \shortstack{{\ensuremath{0.0398701}}\\{[\ensuremath{0.0347462},}\\{\ensuremath{0.0454903}]}} & 2297 / 5000 & 0 / 0 \\
\shortstack{{Mean / SynthCheX}\\{$-$ Mean / Cheff}} & Cheff & \shortstack{{\ensuremath{-0.0477701}}\\{[\ensuremath{-0.0558783},}\\{\ensuremath{-0.0397965}]}} & \shortstack{{\ensuremath{-0.108361}}\\{[\ensuremath{-0.122278},}\\{\ensuremath{-0.0954331}]}} & 2297 / 1000 & 0 / 0 \\
\shortstack{{Mean / SynthCheX}\\{$-$ Mean / Cheff}} & SynthCheX & \shortstack{{\ensuremath{0.0332486}}\\{[\ensuremath{0.0298327},}\\{\ensuremath{0.0368343}]}} & \shortstack{{\ensuremath{0.091582}}\\{[\ensuremath{0.0823968},}\\{\ensuremath{0.101947}]}} & 2297 / 5000 & 0 / 0 \\
\shortstack{{Mean / Mixed}\\{$-$ Mean / Cheff}} & Cheff & \shortstack{{\ensuremath{-0.000917283}}\\{[\ensuremath{-0.00144801},}\\{\ensuremath{-0.000477144}]}} & \shortstack{{\ensuremath{-0.00439956}}\\{[\ensuremath{-0.00670931},}\\{\ensuremath{-0.00241178}]}} & 2297 / 1000 & 0 / 0 \\
\shortstack{{Mean / Mixed}\\{$-$ Mean / Cheff}} & SynthCheX & \shortstack{{\ensuremath{0.0328933}}\\{[\ensuremath{0.029536},}\\{\ensuremath{0.0364541}]}} & \shortstack{{\ensuremath{0.0897198}}\\{[\ensuremath{0.0809758},}\\{\ensuremath{0.0994036}]}} & 2297 / 5000 & 0 / 0 \\
\shortstack{{Medical / SynthCheX}\\{$-$ Medical / Cheff}} & Cheff & \shortstack{{\ensuremath{-0.0200827}}\\{[\ensuremath{-0.0244556},}\\{\ensuremath{-0.0161701}]}} & \shortstack{{\ensuremath{-0.0636116}}\\{[\ensuremath{-0.0745901},}\\{\ensuremath{-0.0536063}]}} & 2297 / 1000 & 0 / 0 \\
\shortstack{{Medical / SynthCheX}\\{$-$ Medical / Cheff}} & SynthCheX & \shortstack{{\ensuremath{0.0175995}}\\{[\ensuremath{0.0154637},}\\{\ensuremath{0.0200084}]}} & \shortstack{{\ensuremath{0.0645812}}\\{[\ensuremath{0.0565052},}\\{\ensuremath{0.0732764}]}} & 2297 / 5000 & 0 / 0 \\
\shortstack{{Medical / Mixed}\\{$-$ Medical / Cheff}} & Cheff & \shortstack{{\ensuremath{-0.000147148}}\\{[\ensuremath{-0.000288659},}\\{\ensuremath{-2.78298\times10^{-5}}]}} & \shortstack{{\ensuremath{-0.000774466}}\\{[\ensuremath{-0.00151926},}\\{\ensuremath{-0.000146473}]}} & 2297 / 1000 & 0 / 0 \\
\shortstack{{Medical / Mixed}\\{$-$ Medical / Cheff}} & SynthCheX & \shortstack{{\ensuremath{0.0175972}}\\{[\ensuremath{0.0154691},}\\{\ensuremath{0.0200041}]}} & \shortstack{{\ensuremath{0.0645693}}\\{[\ensuremath{0.0564324},}\\{\ensuremath{0.0731979}]}} & 2297 / 5000 & 0 / 0 \\
\bottomrule
\end{tabular}
\caption{Paired bootstrap differences in AUROC and standardized partial AUC across secondary generator-transfer heads sharing the same visual backbone.}
\label{tab:paired-matched-ranking-all}
\end{table*}

\begin{table*}[tbp]
\centering
\begingroup
\setlength{\tabcolsep}{4pt}
\renewcommand{\arraystretch}{1.15}
\begin{tabular}{@{}lccccc@{}}
\toprule
Method & \shortstack{{NIH reference}\\{N; U}} & \shortstack{{1\% policy}\\{flags; \% [95\%]}} & \shortstack{{5\% policy}\\{flags; \% [95\%]}} & \shortstack{{10\% policy}\\{flags; \% [95\%]}} & \shortstack{{External U}\\{of 957}} \\
\midrule
Pilot CLS & \shortstack{{2914}\\{U=0}} & \shortstack{{257/957; 26.85}\\{[24.14, 29.75]}} & \shortstack{{491/957; 51.31}\\{[48.14, 54.46]}} & \shortstack{{611/957; 63.85}\\{[60.75, 66.83]}} & 0 \\
DINOv3 CLS raw & \shortstack{{3000}\\{U=0}} & \shortstack{{374/957; 39.08}\\{[36.04, 42.21]}} & \shortstack{{568/957; 59.35}\\{[56.21, 62.42]}} & \shortstack{{663/957; 69.28}\\{[66.28, 72.12]}} & 0 \\
DINOv3 mean raw & \shortstack{{3000}\\{U=0}} & \shortstack{{125/957; 13.06}\\{[11.07, 15.35]}} & \shortstack{{292/957; 30.51}\\{[27.68, 33.50]}} & \shortstack{{413/957; 43.16}\\{[40.05, 46.31]}} & 0 \\
Medical mean raw & \shortstack{{3000}\\{U=0}} & \shortstack{{61/957; 6.37}\\{[4.99, 8.10]}} & \shortstack{{152/957; 15.88}\\{[13.70, 18.33]}} & \shortstack{{226/957; 23.62}\\{[21.03, 26.41]}} & 0 \\
DINOv3 CLS augmented & \shortstack{{3000}\\{U=0}} & \shortstack{{312/957; 32.60}\\{[29.71, 35.64]}} & \shortstack{{493/957; 51.52}\\{[48.35, 54.67]}} & \shortstack{{607/957; 63.43}\\{[60.33, 66.42]}} & 0 \\
Pixel statistics & \shortstack{{3000}\\{U=0}} & \shortstack{{11/957; 1.15}\\{[0.64, 2.05]}} & \shortstack{{27/957; 2.82}\\{[1.95, 4.07]}} & \shortstack{{40/957; 4.18}\\{[3.08, 5.64]}} & 0 \\
\bottomrule
\end{tabular}
\endgroup
\caption{Out-of-distribution false alarms on 957 genuine Open-I frontal radiographs evaluated under frozen NIH-calibrated thresholds ($\alpha\in\{1\%, 5\%, 10\%\}$).}
\label{tab:supp-external-genuine}
\end{table*}

\emph{Conclusion.} Under canonical preprocessing, all DINOv3 and MedGemma SigLIP probes reject $\ge 99.7\%$ of in-domain Cheff-identified test images and $76.2\%$--$96.5\%$ of unseen SynthCheX images across $\alpha\in\{1\%, 5\%, 10\%\}$, achieving SynthCheX AUROCs in $[0.9699, 0.9891]$. By contrast, the rubric-field probe performs poorly (AUROC $0.6301$ Cheff, $0.7181$ SynthCheX), while the 6-feature pixel-statistic classifier achieves substantial separability (AUROC $0.9321$ and $0.8962$). Cross-generator ablations confirm that single-generator probes suffer measurable transfer loss on the unseen generator ($311/5{,}000$ SynthCheX passes for \texttt{CLS / Cheff}; $114/1{,}000$ Cheff passes for \texttt{CLS / SynthCheX} at $\alpha=5\%$). Finally, on genuine Open-I radiographs at the nominal 5\% threshold, all learned visual probes suffer severe false-alarm inflation ($15.88\%$--$59.35\%$), whereas the simple pixel-statistic probe flags $2.82\%$.

\subsection{External Baseline Comparison: MedForge-Reasoner}
\label{sec:supp-medforge-results}
\emph{Question.} How does the external localized-forgery detector MedForge-Reasoner classify genuine and full-image synthetic radiographs at its native categorical operating point, both on the 300-image canonical panels and on Qwen-selected submission lists? Across all 1,095 unique evaluated radiographs (900 panel images plus 195 unique selected submissions), MedForge-Reasoner returned 1,095 valid categorical verdicts with zero token-limit truncations. Table~\ref{tab:medforge-native} reports native classification rates on the 300-image canonical panels, and Table~\ref{tab:medforge-selected} reports origin passage on the fixed $B=1{,}000, K=20$ submission lists.

\begin{table*}[tbp]
\centering
\begin{tabular}{llrrr}
\toprule
Collection & Endpoint & Count & Rate (\%) & Wilson 95\% (\%) \\
\midrule
NIH genuine & Genuine flagged & 178/300 & 59.3 & 53.7--64.7 \\
Cheff-identified & Generated passed & 127/300 & 42.3 & 36.9--48.0 \\
SynthCheX & Generated passed & 50/300 & 16.7 & 12.9--21.3 \\
\bottomrule
\end{tabular}
\caption{Zero-shot transfer performance of MedForge-Reasoner at its native categorical operating point across 300 canonical test radiographs per collection.}
\label{tab:medforge-native}
\end{table*}

\begin{table*}[tbp]
\centering
\begin{tabular}{llrr}
\toprule
Collection & Selection & Passes & 95\% (\%) \\
\midrule
Cheff-identified & Surrogate & 4/20 & 8.1--41.6 \\
Cheff-identified & Quality & 4/20 & 8.1--41.6 \\
Cheff-identified & Uniform & 9/20 & 25.8--65.8 \\
SynthCheX & Surrogate & 1/20 & 0.9--23.6 \\
SynthCheX & Quality & 0/20 & 0.0--16.1 \\
SynthCheX & Uniform & 5/20 & 11.2--46.9 \\
\bottomrule
\end{tabular}
\caption{Number of Qwen-selected submissions ($B=1{,}000, K=20$) misclassified as authentic (``passed'') by MedForge-Reasoner.}
\label{tab:medforge-selected}
\end{table*}

\emph{Conclusion.} Trained for localized tampering rather than full-image synthesis, MedForge-Reasoner exhibits severe miscalibration on our intake task: it falsely flags $59.3\%$ ($178/300$) of genuine NIH radiographs as fake while allowing $42.3\%$ ($127/300$) of Cheff-identified and $16.7\%$ ($50/300$) of SynthCheX radiographs to pass as real.

\subsection{Post-Processing Robustness and Joint Selection Outcomes}
\emph{Question.} How do the four threshold calibration policies behave across each of the 13 post-processing perturbations, and how do all seven core origin heads perform when combined with the quality gate on Qwen-selected submissions? Table~\ref{tab:all-routes-primary-policies} reports per-condition false alarms and rejections across all 13 perturbations, Table~\ref{tab:supp-processing-aggregate} decomposes condition-averaged differences into model effects (Contrast~A), end-to-end policy effects (Contrast~B), and threshold-recalibration effects (Contrast~C), and Tables~\ref{tab:all-method-selected-joint-1}--\ref{tab:all-method-selected-joint-2} report joint quality-and-origin passage across all seven core heads at $B=1{,}000, K=20$.

\begin{table*}[tbp]
\centering
\begin{tabular}{lrrrr}
\toprule
Condition & Raw/canonical & Raw/matched & Augmented/matched & Augmented/global \\
\midrule
Canonical & \shortstack[c]{45 / 1000 / 954\\U: 0 / 0 / 0} & \shortstack[c]{45 / 1000 / 954\\U: 0 / 0 / 0} & \shortstack[c]{43 / 1000 / 940\\U: 0 / 0 / 0} & \shortstack[c]{19 / 1000 / 913\\U: 0 / 0 / 0} \\
JPEG95 & \shortstack[c]{503 / 1000 / 988\\U: 0 / 0 / 0} & \shortstack[c]{36 / 998 / 907\\U: 0 / 0 / 0} & \shortstack[c]{37 / 1000 / 937\\U: 0 / 0 / 0} & \shortstack[c]{52 / 1000 / 947\\U: 0 / 0 / 0} \\
JPEG85 & \shortstack[c]{706 / 999 / 991\\U: 0 / 0 / 0} & \shortstack[c]{55 / 966 / 702\\U: 0 / 0 / 0} & \shortstack[c]{40 / 998 / 916\\U: 0 / 0 / 0} & \shortstack[c]{25 / 998 / 899\\U: 0 / 0 / 0} \\
JPEG75 & \shortstack[c]{820 / 1000 / 987\\U: 0 / 0 / 0} & \shortstack[c]{65 / 889 / 486\\U: 0 / 0 / 0} & \shortstack[c]{41 / 998 / 906\\U: 0 / 0 / 0} & \shortstack[c]{76 / 999 / 935\\U: 0 / 0 / 0} \\
JPEG50 & \shortstack[c]{950 / 1000 / 998\\U: 0 / 0 / 0} & \shortstack[c]{68 / 800 / 355\\U: 0 / 0 / 0} & \shortstack[c]{52 / 992 / 845\\U: 0 / 0 / 0} & \shortstack[c]{614 / 1000 / 996\\U: 0 / 0 / 0} \\
Resize192 & \shortstack[c]{651 / 1000 / 989\\U: 0 / 0 / 0} & \shortstack[c]{68 / 995 / 841\\U: 0 / 0 / 0} & \shortstack[c]{41 / 998 / 922\\U: 0 / 0 / 0} & \shortstack[c]{144 / 1000 / 971\\U: 0 / 0 / 0} \\
Resize128 & \shortstack[c]{807 / 997 / 978\\U: 0 / 0 / 0} & \shortstack[c]{117 / 921 / 595\\U: 0 / 0 / 0} & \shortstack[c]{48 / 996 / 886\\U: 0 / 0 / 0} & \shortstack[c]{64 / 997 / 913\\U: 0 / 0 / 0} \\
Resize64 & \shortstack[c]{485 / 955 / 838\\U: 0 / 0 / 0} & \shortstack[c]{47 / 697 / 287\\U: 0 / 0 / 0} & \shortstack[c]{62 / 927 / 717\\U: 0 / 0 / 0} & \shortstack[c]{388 / 994 / 980\\U: 0 / 0 / 0} \\
Native512 & \shortstack[c]{38 / 1000 / 947\\U: 0 / 0 / 0} & \shortstack[c]{38 / 1000 / 946\\U: 0 / 0 / 0} & \shortstack[c]{37 / 1000 / 948\\U: 0 / 0 / 0} & \shortstack[c]{6 / 1000 / 848\\U: 0 / 0 / 0} \\
Border4 & \shortstack[c]{44 / 1000 / 851\\U: 0 / 0 / 0} & \shortstack[c]{58 / 1000 / 866\\U: 0 / 0 / 0} & \shortstack[c]{54 / 1000 / 904\\U: 0 / 0 / 0} & \shortstack[c]{22 / 1000 / 852\\U: 0 / 0 / 0} \\
Padding4 & \shortstack[c]{44 / 1000 / 878\\U: 0 / 0 / 0} & \shortstack[c]{66 / 1000 / 892\\U: 0 / 0 / 0} & \shortstack[c]{49 / 1000 / 905\\U: 0 / 0 / 0} & \shortstack[c]{26 / 1000 / 884\\U: 0 / 0 / 0} \\
Central64 & \shortstack[c]{4 / 995 / 901\\U: 0 / 0 / 0} & \shortstack[c]{41 / 999 / 965\\U: 0 / 0 / 0} & \shortstack[c]{36 / 1000 / 950\\U: 0 / 0 / 0} & \shortstack[c]{9 / 998 / 899\\U: 0 / 0 / 0} \\
Border26 & \shortstack[c]{90 / 996 / 827\\U: 0 / 0 / 0} & \shortstack[c]{62 / 995 / 803\\U: 0 / 0 / 0} & \shortstack[c]{54 / 997 / 831\\U: 0 / 0 / 0} & \shortstack[c]{48 / 997 / 823\\U: 0 / 0 / 0} \\
\bottomrule
\end{tabular}
\caption{Origin-check performance across all 13 processing conditions at nominal $\alpha=5\%$ over 1,000 test radiographs per collection. Each cell reports \texttt{NIH flags / Cheff rejections / SynthCheX rejections} (with unavailable counts \texttt{U} below).}
\label{tab:all-routes-primary-policies}
\end{table*}

\begin{table*}[tbp]
\centering
\begingroup
\setlength{\tabcolsep}{4pt}
\renewcommand{\arraystretch}{1.15}
\begin{tabular}{@{}lcccccc@{}}
\toprule
Comparison & Collection & Endpoint & \shortstack{{Baseline}\\{achieved \%}} & \shortstack{{Other}\\{achieved \%}} & \shortstack{{$\Delta$ pp}\\{[95\%]}} & \shortstack{{Unavailable}\\{views}\\{baseline/other}} \\
\midrule
A: head change & NIH & Genuine flags & 5.89 & 4.57 & \shortstack{{-1.32}\\{[-1.96, -0.67]}} & 0/0 \\
A: head change & Cheff & Generated rejection & 94.31 & 99.28 & \shortstack{{+4.97}\\{[+4.45, +5.51]}} & 0/0 \\
A: head change & SynthCheX & Generated rejection & 73.84 & 89.28 & \shortstack{{+15.45}\\{[+14.47, +16.51]}} & 0/0 \\
B: full policy & NIH & Genuine flags & 39.90 & 4.57 & \shortstack{{-35.33}\\{[-36.32, -34.31]}} & 0/0 \\
B: full policy & Cheff & Generated rejection & 99.55 & 99.28 & \shortstack{{-0.28}\\{[-0.43, -0.14]}} & 0/0 \\
B: full policy & SynthCheX & Generated rejection & 93.28 & 89.28 & \shortstack{{-4.00}\\{[-4.78, -3.18]}} & 0/0 \\
C: recalibration & NIH & Genuine flags & 39.90 & 5.89 & \shortstack{{-34.01}\\{[-34.92, -33.08]}} & 0/0 \\
C: recalibration & Cheff & Generated rejection & 99.55 & 94.31 & \shortstack{{-5.25}\\{[-5.81, -4.71]}} & 0/0 \\
C: recalibration & SynthCheX & Generated rejection & 93.28 & 73.84 & \shortstack{{-19.45}\\{[-20.35, -18.56]}} & 0/0 \\
\bottomrule
\end{tabular}
\endgroup
\caption{Condition-averaged percentages and paired bootstrap differences (in percentage points) across the 13 processing conditions at $\alpha=5\%$: Contrast~A (\emph{Augmented/matched} minus \emph{Raw/matched}), Contrast~B (\emph{Augmented/matched} minus \emph{Raw/canonical}), and Contrast~C (\emph{Raw/matched} minus \emph{Raw/canonical}).}
\label{tab:supp-processing-aggregate}
\end{table*}

\begin{table*}[tbp]
\centering
\begin{tabular}{llrr}
\toprule
Method & Strategy & \shortstack{{Cheff: passed both}\\{/ submitted}} & \shortstack{{SynthCheX: passed both}\\{/ submitted}} \\
\midrule
Pilot CLS & Uniform & \shortstack[c]{0/20\\{[0.0, 16.1]\%}} & \shortstack[c]{1/20\\{[0.9, 23.6]\%}} \\
Pilot CLS & Quality & \shortstack[c]{0/20\\{[0.0, 16.1]\%}} & \shortstack[c]{0/20\\{[0.0, 16.1]\%}} \\
Pilot CLS & Surrogate & \shortstack[c]{0/20\\{[0.0, 16.1]\%}} & \shortstack[c]{1/20\\{[0.9, 23.6]\%}} \\
DINOv3 CLS raw & Uniform & \shortstack[c]{0/20\\{[0.0, 16.1]\%}} & \shortstack[c]{2/20\\{[2.8, 30.1]\%}} \\
DINOv3 CLS raw & Quality & \shortstack[c]{0/20\\{[0.0, 16.1]\%}} & \shortstack[c]{1/20\\{[0.9, 23.6]\%}} \\
DINOv3 CLS raw & Surrogate & \shortstack[c]{0/20\\{[0.0, 16.1]\%}} & \shortstack[c]{2/20\\{[2.8, 30.1]\%}} \\
DINOv3 mean raw & Uniform & \shortstack[c]{0/20\\{[0.0, 16.1]\%}} & \shortstack[c]{2/20\\{[2.8, 30.1]\%}} \\
DINOv3 mean raw & Quality & \shortstack[c]{0/20\\{[0.0, 16.1]\%}} & \shortstack[c]{1/20\\{[0.9, 23.6]\%}} \\
DINOv3 mean raw & Surrogate & \shortstack[c]{0/20\\{[0.0, 16.1]\%}} & \shortstack[c]{1/20\\{[0.9, 23.6]\%}} \\
Medical mean raw & Uniform & \shortstack[c]{0/20\\{[0.0, 16.1]\%}} & \shortstack[c]{1/20\\{[0.9, 23.6]\%}} \\
Medical mean raw & Quality & \shortstack[c]{0/20\\{[0.0, 16.1]\%}} & \shortstack[c]{0/20\\{[0.0, 16.1]\%}} \\
Medical mean raw & Surrogate & \shortstack[c]{0/20\\{[0.0, 16.1]\%}} & \shortstack[c]{0/20\\{[0.0, 16.1]\%}} \\
\bottomrule
\end{tabular}
\caption{Joint passage through both the primary MedGemma quality gate and each origin head ($\alpha=5\%$) on the fixed $B=1{,}000, K=20$ submission lists (Part 1 of 2).}
\label{tab:all-method-selected-joint-1}
\end{table*}

\begin{table*}[tbp]
\centering
\begin{tabular}{llrr}
\toprule
Method & Strategy & \shortstack{{Cheff: passed both}\\{/ submitted}} & \shortstack{{SynthCheX: passed both}\\{/ submitted}} \\
\midrule
DINOv3 CLS augmented & Uniform & \shortstack[c]{0/20\\{[0.0, 16.1]\%}} & \shortstack[c]{2/20\\{[2.8, 30.1]\%}} \\
DINOv3 CLS augmented & Quality & \shortstack[c]{0/20\\{[0.0, 16.1]\%}} & \shortstack[c]{1/20\\{[0.9, 23.6]\%}} \\
DINOv3 CLS augmented & Surrogate & \shortstack[c]{0/20\\{[0.0, 16.1]\%}} & \shortstack[c]{1/20\\{[0.9, 23.6]\%}} \\
Rubric fields & Uniform & \shortstack[c]{12/20\\{[38.7, 78.1]\%}} & \shortstack[c]{8/20\\{[21.9, 61.3]\%}} \\
Rubric fields & Quality & \shortstack[c]{19/20\\{[76.4, 99.1]\%}} & \shortstack[c]{17/20\\{[64.0, 94.8]\%}} \\
Rubric fields & Surrogate & \shortstack[c]{19/20\\{[76.4, 99.1]\%}} & \shortstack[c]{14/20\\{[48.1, 85.5]\%}} \\
Pixel statistics & Uniform & \shortstack[c]{5/20\\{[11.2, 46.9]\%}} & \shortstack[c]{3/20\\{[5.2, 36.0]\%}} \\
Pixel statistics & Quality & \shortstack[c]{8/20\\{[21.9, 61.3]\%}} & \shortstack[c]{9/20\\{[25.8, 65.8]\%}} \\
Pixel statistics & Surrogate & \shortstack[c]{8/20\\{[21.9, 61.3]\%}} & \shortstack[c]{5/20\\{[11.2, 46.9]\%}} \\
\bottomrule
\end{tabular}
\caption{Continuation of Table~\ref{tab:all-method-selected-joint-1} (Part 2 of 2).}
\label{tab:all-method-selected-joint-2}
\end{table*}

\emph{Conclusion.} Under static canonical calibration (\emph{Raw/canonical}), JPEG compression and downsampling trigger severe threshold drift, flagging $485$--$950$ of $1{,}000$ genuine NIH radiographs. Condition-matched recalibration (\emph{Raw/matched}) restores NIH false alarms to $36$--$117$ of $1{,}000$ but exposes substantial ranking loss on unseen SynthCheX images (e.g., dropping JPEG 50 rejection from $998/1{,}000$ to $355/1{,}000$), whereas multi-condition training (\emph{Augmented/matched}) recovers $845/1{,}000$ rejection at JPEG 50 with $52/1{,}000$ NIH false alarms. Across all 13 conditions, Contrast~A improves mean Cheff rejection by $+4.97$\,pp and unseen SynthCheX rejection by $+15.45$\,pp while reducing NIH false alarms by $1.32$\,pp. On the $B=1{,}000, K=20$ selected lists, none of the five deep visual probes admits any Cheff-identified candidate or more than $2/20$ SynthCheX candidates through both gates, whereas the weak rubric-field and pixel-statistic baselines admit up to $19/20$ and $9/20$, respectively.

\subsection{Observational Strata, Score Shifts, and Pixel-Statistic Correlations}
\emph{Question.} How do quality passage and origin false alarms vary across recorded projection (AP vs.\ PA) and pathology strata (\emph{No Finding} vs.\ \emph{other finding})? How far do post-processing perturbations shift raw versus augmented origin scores relative to their canonical IQRs? And to what extent do origin scores correlate with elementary pixel statistics within each collection? Table~\ref{tab:descriptive-genuine-observational-strata} reports stratified NIH rates, Table~\ref{tab:NIH-control-score-shifts} reports paired median score shifts relative to canonical IQRs, and Tables~\ref{tab:within-source-pixel-correlations-1}--\ref{tab:within-source-pixel-correlations-4} report within-collection Spearman correlations ($\rho$) across all 72 head--collection--statistic combinations.

\begin{table*}[tbp]
\centering
\begingroup
\setlength{\tabcolsep}{4pt}
\renewcommand{\arraystretch}{1.15}
\begin{tabular}{@{}lcccccc@{}}
\toprule
Method / endpoint & \shortstack{{AP}\\{count; \%; CI}} & \shortstack{{PA}\\{count; \%; CI}} & \shortstack{{Direct std.}\\{rate}} & \shortstack{{No Finding}\\{count; \%; CI}} & \shortstack{{Other finding}\\{count; \%; CI}} & \shortstack{{Finite / U}\\{of 2297}} \\
\midrule
\shortstack{{DINOv3 CLS raw}\\{False flags}} & \shortstack{{28/541}\\{5.18\%}\\{[3.60, 7.38]}} & \shortstack{{82/1756}\\{4.67\%}\\{[3.78, 5.76]}} & 4.76\% & \shortstack{{71/1594}\\{4.45\%}\\{[3.55, 5.58]}} & \shortstack{{39/703}\\{5.55\%}\\{[4.08, 7.49]}} & 2297/0 \\
\shortstack{{DINOv3 CLS aug.}\\{False flags}} & \shortstack{{20/541}\\{3.70\%}\\{[2.41, 5.64]}} & \shortstack{{77/1756}\\{4.38\%}\\{[3.52, 5.45]}} & 4.26\% & \shortstack{{57/1594}\\{3.58\%}\\{[2.77, 4.60]}} & \shortstack{{40/703}\\{5.69\%}\\{[4.21, 7.66]}} & 2297/0 \\
\shortstack{{DINOv3 mean}\\{False flags}} & \shortstack{{25/541}\\{4.62\%}\\{[3.15, 6.73]}} & \shortstack{{98/1756}\\{5.58\%}\\{[4.60, 6.75]}} & 5.41\% & \shortstack{{81/1594}\\{5.08\%}\\{[4.11, 6.27]}} & \shortstack{{42/703}\\{5.97\%}\\{[4.45, 7.98]}} & 2297/0 \\
\shortstack{{Medical mean}\\{False flags}} & \shortstack{{35/541}\\{6.47\%}\\{[4.69, 8.86]}} & \shortstack{{72/1756}\\{4.10\%}\\{[3.27, 5.13]}} & 4.52\% & \shortstack{{38/1594}\\{2.38\%}\\{[1.74, 3.26]}} & \shortstack{{69/703}\\{9.82\%}\\{[7.83, 12.24]}} & 2297/0 \\
\shortstack{{MedGemma alt.}\\{Quality passes}} & \shortstack{{257/541}\\{47.50\%}\\{[43.33, 51.72]}} & \shortstack{{1460/1756}\\{83.14\%}\\{[81.32, 84.82]}} & 76.80\% & \shortstack{{1330/1594}\\{83.44\%}\\{[81.53, 85.18]}} & \shortstack{{387/703}\\{55.05\%}\\{[51.36, 58.69]}} & 2219/78 \\
\bottomrule
\end{tabular}
\endgroup
\caption{Stratified genuine NIH origin false alarms ($\alpha=5\%$) and primary quality pass rates across recorded projection (541 AP, 1,756 PA) and report findings (1,594 \emph{No Finding}, 703 \emph{other finding}). Direct standardized rates apply fixed development weights ($0.178$ AP, $0.822$ PA).}
\label{tab:descriptive-genuine-observational-strata}
\end{table*}

\begin{table*}[tbp]
\centering
\begin{tabular}{lrrrr}
\toprule
Condition & Raw median shift & Raw canonical IQR & Augmented median shift & Augmented canonical IQR \\
\midrule
Border4 & \ensuremath{-4.77\times10^{-6}} & \ensuremath{0.00604} & \ensuremath{+1.19\times10^{-5}} & \ensuremath{0.00106} \\
Padding4 & \ensuremath{+0.000434} & \ensuremath{0.00604} & \ensuremath{+9.69\times10^{-5}} & \ensuremath{0.00106} \\
Central64 & \ensuremath{-0.000759} & \ensuremath{0.00604} & \ensuremath{-1.37\times10^{-5}} & \ensuremath{0.00106} \\
Border26 & \ensuremath{+0.000918} & \ensuremath{0.00604} & \ensuremath{+0.000285} & \ensuremath{0.00106} \\
JPEG75 & \ensuremath{+0.2} & \ensuremath{0.00604} & \ensuremath{+6.11\times10^{-5}} & \ensuremath{0.00106} \\
Resize128 & \ensuremath{+0.295} & \ensuremath{0.00604} & \ensuremath{+0.000114} & \ensuremath{0.00106} \\
\bottomrule
\end{tabular}
\caption{Paired median shifts in genuine NIH origin scores (perturbation minus canonical) across 1,000 radiographs for the raw and augmented DINOv3 CLS heads, benchmarked against each head's canonical score interquartile range (IQR).}
\label{tab:NIH-control-score-shifts}
\end{table*}

\begin{table*}[tbp]
\centering
\begin{tabular}{llrr}
\toprule
Collection & Pixel statistic & \shortstack{{DINOv3 CLS raw}\\{$\rho$ [95\% CI]}\\{finite/attempted; U}} & \shortstack{{DINOv3 CLS aug.}\\{$\rho$ [95\% CI]}\\{finite/attempted; U}} \\
\midrule
NIH & Mean & \shortstack{{0.225}\\{[0.185, 0.265]}\\{n=2297/2297; U=0}} & \shortstack{{0.298}\\{[0.262, 0.336]}\\{n=2297/2297; U=0}} \\
NIH & Std. & \shortstack{{0.163}\\{[0.124, 0.203]}\\{n=2297/2297; U=0}} & \shortstack{{0.264}\\{[0.226, 0.302]}\\{n=2297/2297; U=0}} \\
NIH & Entropy & \shortstack{{0.285}\\{[0.245, 0.322]}\\{n=2297/2297; U=0}} & \shortstack{{0.396}\\{[0.361, 0.430]}\\{n=2297/2297; U=0}} \\
NIH & Border & \shortstack{{0.116}\\{[0.076, 0.156]}\\{n=2297/2297; U=0}} & \shortstack{{0.129}\\{[0.087, 0.167]}\\{n=2297/2297; U=0}} \\
NIH & JPEG & \shortstack{{-0.083}\\{[-0.123, -0.042]}\\{n=2297/2297; U=0}} & \shortstack{{-0.129}\\{[-0.170, -0.087]}\\{n=2297/2297; U=0}} \\
NIH & High freq. & \shortstack{{-0.060}\\{[-0.100, -0.017]}\\{n=2297/2297; U=0}} & \shortstack{{-0.034}\\{[-0.075, 0.010]}\\{n=2297/2297; U=0}} \\
Cheff & Mean & \shortstack{{0.103}\\{[0.041, 0.166]}\\{n=1000/1000; U=0}} & \shortstack{{0.045}\\{[-0.016, 0.110]}\\{n=1000/1000; U=0}} \\
Cheff & Std. & \shortstack{{0.151}\\{[0.088, 0.213]}\\{n=1000/1000; U=0}} & \shortstack{{0.163}\\{[0.103, 0.227]}\\{n=1000/1000; U=0}} \\
Cheff & Entropy & \shortstack{{0.194}\\{[0.134, 0.255]}\\{n=1000/1000; U=0}} & \shortstack{{0.192}\\{[0.131, 0.253]}\\{n=1000/1000; U=0}} \\
\bottomrule
\end{tabular}
\caption{Within-collection Spearman rank correlations ($\rho$) between canonical origin scores and the six low-level pixel statistics (Part 1 of 4).}
\label{tab:within-source-pixel-correlations-1}
\end{table*}

\begin{table*}[tbp]
\centering
\begin{tabular}{llrr}
\toprule
Collection & Pixel statistic & \shortstack{{DINOv3 CLS raw}\\{$\rho$ [95\% CI]}\\{finite/attempted; U}} & \shortstack{{DINOv3 CLS aug.}\\{$\rho$ [95\% CI]}\\{finite/attempted; U}} \\
\midrule
Cheff & Border & \shortstack{{-0.143}\\{[-0.202, -0.083]}\\{n=1000/1000; U=0}} & \shortstack{{-0.146}\\{[-0.206, -0.087]}\\{n=1000/1000; U=0}} \\
Cheff & JPEG & \shortstack{{0.191}\\{[0.128, 0.251]}\\{n=1000/1000; U=0}} & \shortstack{{0.179}\\{[0.118, 0.236]}\\{n=1000/1000; U=0}} \\
Cheff & High freq. & \shortstack{{0.060}\\{[-0.004, 0.124]}\\{n=1000/1000; U=0}} & \shortstack{{0.038}\\{[-0.024, 0.101]}\\{n=1000/1000; U=0}} \\
SynthCheX & Mean & \shortstack{{0.035}\\{[0.005, 0.062]}\\{n=5000/5000; U=0}} & \shortstack{{0.057}\\{[0.028, 0.086]}\\{n=5000/5000; U=0}} \\
SynthCheX & Std. & \shortstack{{-0.068}\\{[-0.095, -0.039]}\\{n=5000/5000; U=0}} & \shortstack{{-0.098}\\{[-0.125, -0.069]}\\{n=5000/5000; U=0}} \\
SynthCheX & Entropy & \shortstack{{0.034}\\{[0.006, 0.062]}\\{n=5000/5000; U=0}} & \shortstack{{0.047}\\{[0.020, 0.074]}\\{n=5000/5000; U=0}} \\
SynthCheX & Border & \shortstack{{0.054}\\{[0.026, 0.083]}\\{n=5000/5000; U=0}} & \shortstack{{0.031}\\{[0.003, 0.058]}\\{n=5000/5000; U=0}} \\
SynthCheX & JPEG & \shortstack{{-0.034}\\{[-0.062, -0.007]}\\{n=5000/5000; U=0}} & \shortstack{{-0.043}\\{[-0.072, -0.017]}\\{n=5000/5000; U=0}} \\
SynthCheX & High freq. & \shortstack{{-0.104}\\{[-0.131, -0.076]}\\{n=5000/5000; U=0}} & \shortstack{{-0.139}\\{[-0.167, -0.112]}\\{n=5000/5000; U=0}} \\
\bottomrule
\end{tabular}
\caption{Continuation of Table~\ref{tab:within-source-pixel-correlations-1} (Part 2 of 4).}
\label{tab:within-source-pixel-correlations-2}
\end{table*}

\begin{table*}[tbp]
\centering
\begin{tabular}{llrr}
\toprule
Collection & Pixel statistic & \shortstack{{DINOv3 mean}\\{$\rho$ [95\% CI]}\\{finite/attempted; U}} & \shortstack{{Medical mean}\\{$\rho$ [95\% CI]}\\{finite/attempted; U}} \\
\midrule
NIH & Mean & \shortstack{{0.178}\\{[0.139, 0.219]}\\{n=2297/2297; U=0}} & \shortstack{{0.325}\\{[0.289, 0.364]}\\{n=2297/2297; U=0}} \\
NIH & Std. & \shortstack{{0.047}\\{[0.005, 0.090]}\\{n=2297/2297; U=0}} & \shortstack{{0.274}\\{[0.237, 0.312]}\\{n=2297/2297; U=0}} \\
NIH & Entropy & \shortstack{{0.194}\\{[0.154, 0.234]}\\{n=2297/2297; U=0}} & \shortstack{{0.415}\\{[0.382, 0.447]}\\{n=2297/2297; U=0}} \\
NIH & Border & \shortstack{{0.054}\\{[0.012, 0.095]}\\{n=2297/2297; U=0}} & \shortstack{{0.099}\\{[0.056, 0.141]}\\{n=2297/2297; U=0}} \\
NIH & JPEG & \shortstack{{-0.028}\\{[-0.069, 0.012]}\\{n=2297/2297; U=0}} & \shortstack{{-0.055}\\{[-0.100, -0.013]}\\{n=2297/2297; U=0}} \\
NIH & High freq. & \shortstack{{-0.006}\\{[-0.047, 0.036]}\\{n=2297/2297; U=0}} & \shortstack{{-0.111}\\{[-0.152, -0.069]}\\{n=2297/2297; U=0}} \\
Cheff & Mean & \shortstack{{0.010}\\{[-0.053, 0.074]}\\{n=1000/1000; U=0}} & \shortstack{{-0.070}\\{[-0.133, -0.003]}\\{n=1000/1000; U=0}} \\
Cheff & Std. & \shortstack{{0.119}\\{[0.055, 0.182]}\\{n=1000/1000; U=0}} & \shortstack{{0.221}\\{[0.160, 0.281]}\\{n=1000/1000; U=0}} \\
Cheff & Entropy & \shortstack{{0.153}\\{[0.091, 0.216]}\\{n=1000/1000; U=0}} & \shortstack{{0.259}\\{[0.197, 0.319]}\\{n=1000/1000; U=0}} \\
\bottomrule
\end{tabular}
\caption{Continuation of Table~\ref{tab:within-source-pixel-correlations-1} (Part 3 of 4).}
\label{tab:within-source-pixel-correlations-3}
\end{table*}

\begin{table*}[tbp]
\centering
\begin{tabular}{llrr}
\toprule
Collection & Pixel statistic & \shortstack{{DINOv3 mean}\\{$\rho$ [95\% CI]}\\{finite/attempted; U}} & \shortstack{{Medical mean}\\{$\rho$ [95\% CI]}\\{finite/attempted; U}} \\
\midrule
Cheff & Border & \shortstack{{-0.122}\\{[-0.182, -0.061]}\\{n=1000/1000; U=0}} & \shortstack{{-0.051}\\{[-0.113, 0.013]}\\{n=1000/1000; U=0}} \\
Cheff & JPEG & \shortstack{{0.150}\\{[0.085, 0.210]}\\{n=1000/1000; U=0}} & \shortstack{{0.158}\\{[0.096, 0.217]}\\{n=1000/1000; U=0}} \\
Cheff & High freq. & \shortstack{{0.019}\\{[-0.043, 0.084]}\\{n=1000/1000; U=0}} & \shortstack{{-0.012}\\{[-0.075, 0.052]}\\{n=1000/1000; U=0}} \\
SynthCheX & Mean & \shortstack{{-0.003}\\{[-0.033, 0.024]}\\{n=5000/5000; U=0}} & \shortstack{{0.219}\\{[0.193, 0.246]}\\{n=5000/5000; U=0}} \\
SynthCheX & Std. & \shortstack{{-0.126}\\{[-0.153, -0.097]}\\{n=5000/5000; U=0}} & \shortstack{{-0.165}\\{[-0.193, -0.137]}\\{n=5000/5000; U=0}} \\
SynthCheX & Entropy & \shortstack{{0.108}\\{[0.081, 0.135]}\\{n=5000/5000; U=0}} & \shortstack{{0.292}\\{[0.265, 0.317]}\\{n=5000/5000; U=0}} \\
SynthCheX & Border & \shortstack{{0.100}\\{[0.072, 0.127]}\\{n=5000/5000; U=0}} & \shortstack{{0.022}\\{[-0.007, 0.049]}\\{n=5000/5000; U=0}} \\
SynthCheX & JPEG & \shortstack{{-0.065}\\{[-0.092, -0.037]}\\{n=5000/5000; U=0}} & \shortstack{{0.005}\\{[-0.023, 0.031]}\\{n=5000/5000; U=0}} \\
SynthCheX & High freq. & \shortstack{{-0.078}\\{[-0.107, -0.049]}\\{n=5000/5000; U=0}} & \shortstack{{-0.084}\\{[-0.111, -0.055]}\\{n=5000/5000; U=0}} \\
\bottomrule
\end{tabular}
\caption{Continuation of Table~\ref{tab:within-source-pixel-correlations-1} (Part 4 of 4).}
\label{tab:within-source-pixel-correlations-4}
\end{table*}

\emph{Conclusion.} While MedGemma quality pass rates vary substantially across clinical strata ($47.50\%$ AP vs.\ $83.14\%$ PA; $83.44\%$ \emph{No Finding} vs.\ $55.05\%$ \emph{other finding}), DINOv3 origin false alarms remain stable across strata ($3.58\%$--$5.97\%$), whereas the MedGemma SigLIP medical head exhibits higher false alarms on scans with clinical findings ($9.82\%$ vs.\ $2.38\%$). On paired NIH radiographs, JPEG 75 and resize 128 shift the raw CLS head's median score by $+0.200$ and $+0.295$—over $33\times$ its canonical IQR ($0.00604$)—whereas multi-condition augmentation suppresses these shifts to $<1.2\times 10^{-4}$.

\subsection{Embedding Proximity ($\text{MMD}^2$) and Visual Screening Atlases}
\begin{figure*}[tbp]
\centering
\includegraphics[width=\textwidth]{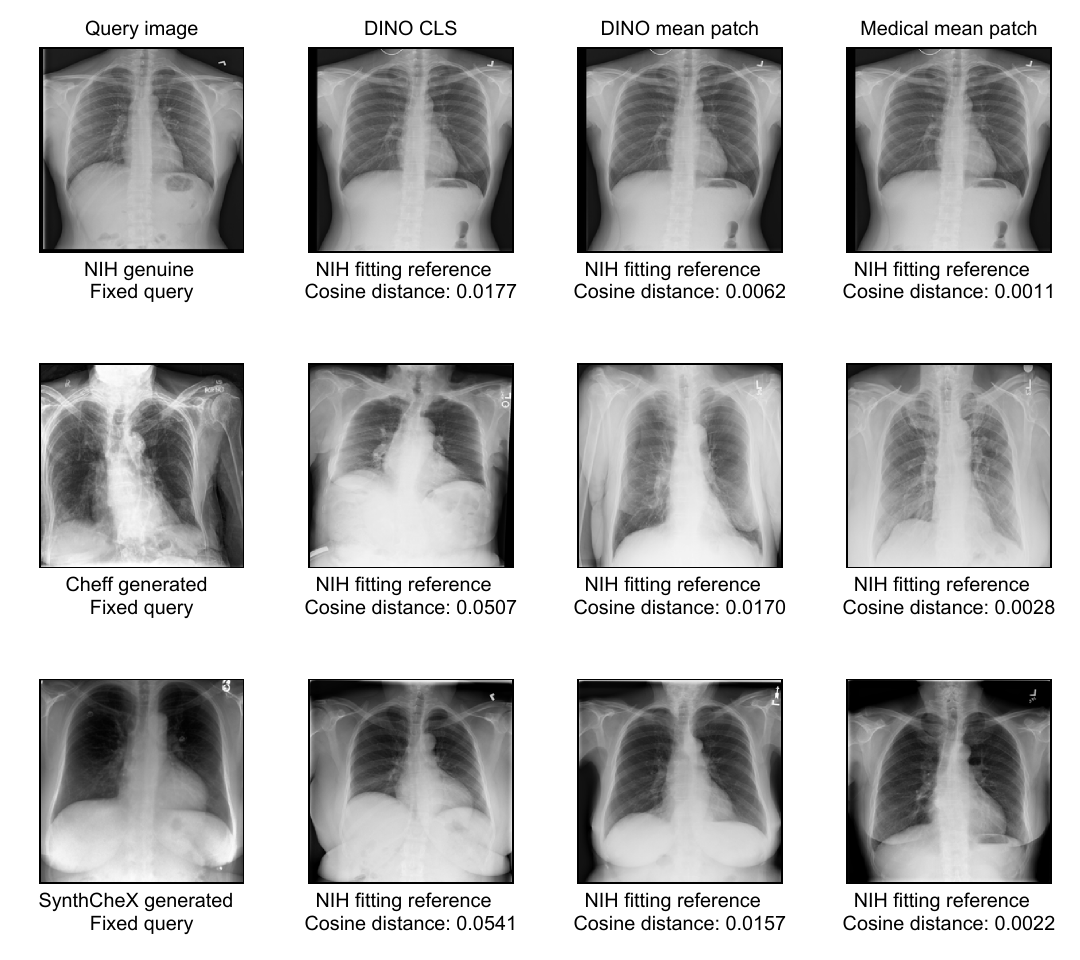}
\caption{Cosine nearest-neighbor retrievals from the 1,500-image genuine NIH fit split for hash-selected query radiographs across DINOv3 CLS, DINOv3 mean-patch, and MedGemma SigLIP mean-patch representations.}
\label{fig:nearest-neighbors}
\end{figure*}

\begin{figure*}[tbp]
\centering
\includegraphics[width=\textwidth]{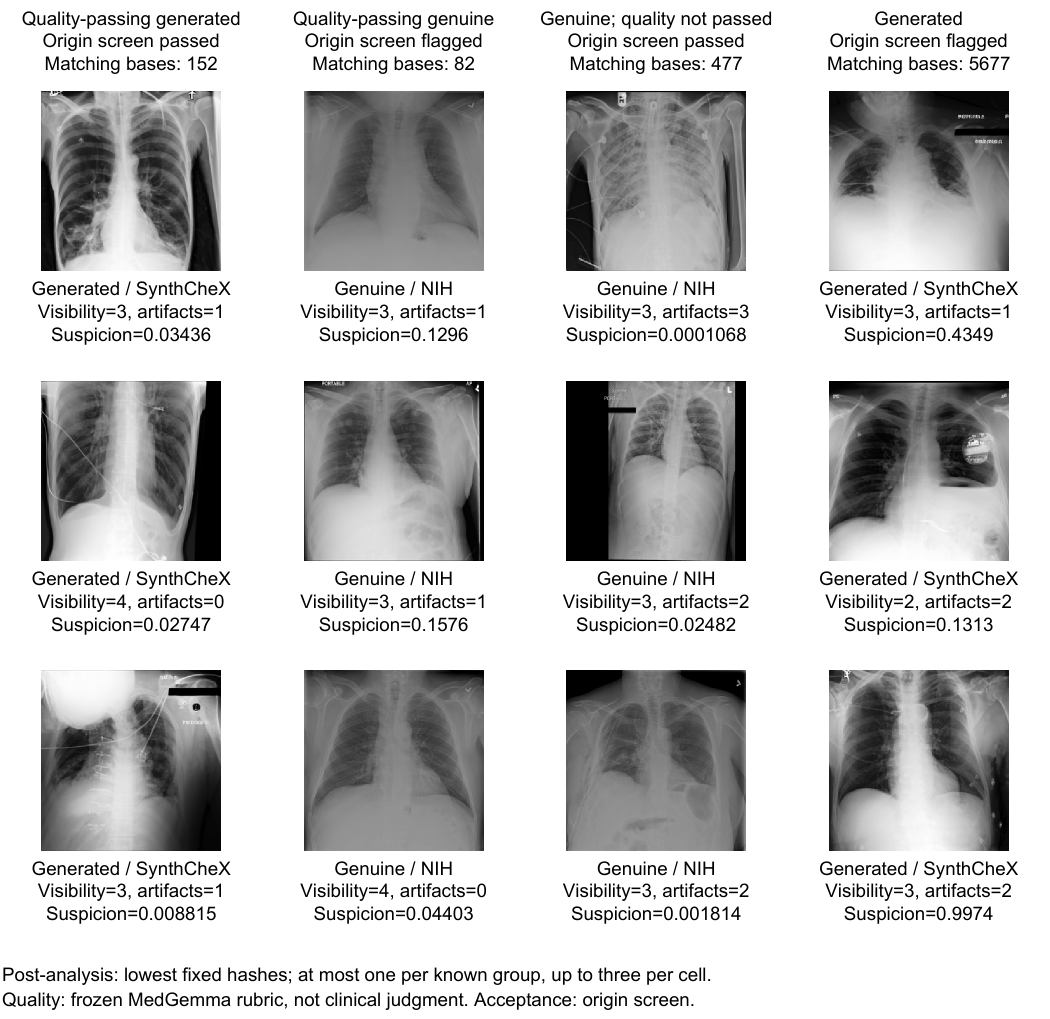}
\caption{Representative test radiographs across the four joint outcomes of the primary canonical intake policy, selected deterministically via lowest SHA256 hashes (up to three examples per cell).}
\label{fig:error-atlas}
\end{figure*}

\emph{Question.} Across visual backbones, which synthetic generator lies closer to genuine NIH radiographs in feature space, and what do representative screening outcomes look like visually? Table~\ref{tab:descriptive-embedding-mmd} reports unbiased squared MMD ($\text{MMD}^2$) between 500-image panels, Table~\ref{tab:descriptive-embedding-paired-contrasts} reports paired Cheff-minus-SynthCheX $\text{MMD}^2$ contrasts, Figure~\ref{fig:nearest-neighbors} displays hash-selected nearest neighbors in each embedding space, and Figure~\ref{fig:error-atlas} presents an error atlas across the four joint intake outcomes.

\begin{table*}[tbp]
\centering
\begingroup
\setlength{\tabcolsep}{4pt}
\renewcommand{\arraystretch}{1.15}
\begin{tabular}{@{}lccccc@{}}
\toprule
Encoder & Normalization & Contrast & \shortstack{{N}\\{per panel}} & Unbiased MMD$^2$ & Fixed-panel 95\% interval \\
\midrule
DINOv3 CLS & Raw & NIH--NIH & 500 & 3.7091e-06 & [4.6171e-06, 4.0745e-05] \\
DINOv3 CLS & Raw & Cheff--NIH & 500 & 0.0014219 & [0.0012872, 0.0016208] \\
DINOv3 CLS & Raw & SynthCheX--NIH & 500 & 0.002268 & [0.002109, 0.0024757] \\
DINOv3 CLS & Angular & NIH--NIH & 500 & 0.00074518 & [0.00088802, 0.0073874] \\
DINOv3 CLS & Angular & Cheff--NIH & 500 & 0.24395 & [0.22127, 0.2778] \\
DINOv3 CLS & Angular & SynthCheX--NIH & 500 & 0.39077 & [0.36331, 0.42667] \\
DINOv3 mean & Raw & NIH--NIH & 500 & 1.1184e-06 & [7.1223e-07, 1.4354e-05] \\
DINOv3 mean & Raw & Cheff--NIH & 500 & 0.00044665 & [0.00040789, 0.00050197] \\
DINOv3 mean & Raw & SynthCheX--NIH & 500 & 0.00048212 & [0.00044682, 0.00053339] \\
DINOv3 mean & Angular & NIH--NIH & 500 & 0.00033241 & [0.00023479, 0.0035858] \\
DINOv3 mean & Angular & Cheff--NIH & 500 & 0.10891 & [0.099416, 0.12252] \\
DINOv3 mean & Angular & SynthCheX--NIH & 500 & 0.11949 & [0.11065, 0.1321] \\
Medical mean & Raw & NIH--NIH & 500 & 0.00019184 & [0.00024992, 0.0039859] \\
Medical mean & Raw & Cheff--NIH & 500 & 0.093953 & [0.086285, 0.10576] \\
Medical mean & Raw & SynthCheX--NIH & 500 & 0.074226 & [0.064746, 0.089426] \\
Medical mean & Angular & NIH--NIH & 500 & 9.5286e-05 & [8.1253e-05, 0.00091094] \\
Medical mean & Angular & Cheff--NIH & 500 & 0.017595 & [0.016108, 0.019849] \\
Medical mean & Angular & SynthCheX--NIH & 500 & 0.017362 & [0.015347, 0.020438] \\
\bottomrule
\end{tabular}
\endgroup
\caption{Unbiased squared MMD ($\text{MMD}^2$) between fixed 500-image panels across raw and angular-normalized encoder representations.}
\label{tab:descriptive-embedding-mmd}
\end{table*}

\begin{table*}[tbp]
\centering
\begingroup
\setlength{\tabcolsep}{4pt}
\renewcommand{\arraystretch}{1.15}
\begin{tabular}{@{}lcccc@{}}
\toprule
Encoder & Normalization & \shortstack{{N}\\{per panel}} & \shortstack{{Cheff minus SynthCheX}\\{MMD$^2$ difference}} & Paired fixed-panel 95\% interval \\
\midrule
DINOv3 CLS & Raw & 500 & -0.00084617 & [-0.0010546, -0.00062756] \\
DINOv3 CLS & Angular & 500 & -0.14682 & [-0.18179, -0.10982] \\
DINOv3 mean & Raw & 500 & -3.5471e-05 & [-8.6209e-05, 1.9759e-05] \\
DINOv3 mean & Angular & 500 & -0.010572 & [-0.022885, 0.0029512] \\
Medical mean & Raw & 500 & 0.019727 & [0.005187, 0.033548] \\
Medical mean & Angular & 500 & 0.00023329 & [-0.0021904, 0.0024338] \\
\bottomrule
\end{tabular}
\endgroup
\caption{Paired differences in squared MMD to the genuine NIH panel ($\text{MMD}^2(\text{Cheff}, \text{NIH}) - \text{MMD}^2(\text{SynthCheX}, \text{NIH})$). Negative values indicate that Cheff-identified embeddings lie closer to genuine NIH radiographs than SynthCheX embeddings.}
\label{tab:descriptive-embedding-paired-contrasts}
\end{table*}

\emph{Conclusion.} Embedding proximity to genuine NIH scans depends on the visual backbone: DINOv3 CLS places Cheff-identified images significantly closer to NIH than SynthCheX under both normalizations, raw MedGemma SigLIP reverses this ordering, and DINOv3 mean-patch and angular SigLIP intervals span zero.

\section{Provenance Registry Experiment}
\label{sec:supp-registry}
\subsection{Experimental Architecture}
Our registry replay experiment evaluates our laboratory provenance registry and its cross-repository export/import verification logic. Each registry entry stores three decoupled fields: (1) technical quality outcome, (2) predicted origin outcome, and (3) external claim-support status (\emph{verified}, \emph{unsupported}, or \emph{unknown}) paired with declared purpose (\emph{clinical acquisition} vs.\ \emph{declared augmentation}). Across paired experimental variants, the image pixel bytes are held strictly identical while varying only the declared purpose and whether an authenticated acquisition record is present.

The registry validates cryptographic signatures on acquisition records and derived-file lineage receipts bound to exact file hashes outside the submitter's control, preserving any unsubstantiated claim in the \emph{unknown} state. We also contrast two cross-repository import implementations: a \emph{safe importer} that rehashes incoming pixel bytes and preserves missing or invalid verification strictly as \emph{unknown}, and a \emph{permissive importer} that trusts unverified claim strings regardless of whether verification columns are retained or stripped.

\subsection{Policy Replay Outcomes}
We evaluate 8 hash-selected base radiographs instantiated across 24 claim-and-purpose variants and replayed under 3 intake policies (72 policy replays total). Because variants reuse the same 8 base radiographs, tabulated counts represent deterministic policy verifications rather than population rate estimates. A registry-only intake policy rejects all unsupported clinical acquisition claims without invoking pixel classifiers, while admitting declared synthetic augmentations for model training. Layering pixel-level quality and origin gates atop verified clinical acquisitions reduces accepted genuine examinations from $4/4$ to $2/4$ due to pixel-gate false rejections, while leaving declared augmentation admissions unchanged.

\section{Reproduction and Execution Audit Logs}
\label{sec:supp-records}
To ensure full transparency, this section documents hardware scheduling events, lineage audits, and parser adjustments recorded during experimental execution, together with measured compute costs.

\subsection{Execution and Protocol Audit Log}
\paragraph{Shared GPU contention during development stress run.} During the development stress evaluation of the original quality rubric, an external process briefly shared one GPU; consequently, wall-clock timing for that development run reflects multi-tenant contention.

\paragraph{Automated GPU ownership locking.} All subsequent job dispatchers enforced automated device-ownership checks with 300-second polling intervals and inter-batch synchronization barriers.

\paragraph{Initial eight alternative-rubric responses.} The first eight MedGemma alternative-rubric responses were generated prior to activating the strict device-ownership lock; we preserved those cached responses unaltered and logged the shared-hardware event rather than re-running inference.

\paragraph{Pilot-head calibration deduplication.} As detailed in Section~\ref{sec:supp-origin}, auditing the legacy pilot head against the 3,000-patient calibration split identified 86 overlapping patients used during pilot fitting or tuning. While historical 3,000-patient logs are preserved in the archive for completeness, all reported pilot-head evaluations strictly use the deduplicated 2,914-patient calibration reference.

\paragraph{Development control augmentation and failure accounting.} During development stress testing, we expanded the non-radiograph control set by adding 10 distinct checkerboard patterns without altering the existing 3,060 radiograph inputs or their cached outputs. Prior to inspecting clean-NIH or control outcomes under the alternative rubric, we also locked the accounting rule that counts unparseable or token-capped responses against the clean-image assessability criterion.

\paragraph{MedForge-Reasoner post-think parser refinement.} On the initial 4-image timing benchmark, strict single-token parsing failed when MedForge-Reasoner emitted a complete sentence after its closing \texttt{</think>} tag. We preserved the raw failure record and updated the parser to extract an unambiguous concluding classification sentence following a closed \texttt{<think>} block, resolving all 4 calibration responses without re-running model inference or altering cached text.

\paragraph{MedForge-Reasoner device-map adapter.} Because the upstream MedForge-Reasoner checkpoint omits \texttt{device\_map} metadata in its configuration file, we used a read-only loader wrapper to expose device placement for validation while verifying that all parameters resided on-GPU in native BF16.

\paragraph{Resumed Qwen surrogate inference allocation.} The 6,000 Qwen surrogate rating tasks were completed across an initial run of 1,748 tasks and a resumed run of 4,252 tasks; consequently, cumulative batch-generation time is reported in Table~\ref{tab:scoped-inference-cost}, while single-process end-to-end wall time is omitted (\texttt{--}).

\paragraph{Historical feature-extraction device logs.} Early feature-extraction logs recorded tensor dtypes and parameter counts without logging per-layer GPU device indices; interrupted runs preserve their original cache keys, batch manifests, and batch timings without retrospective wall-time interpolation.

\paragraph{Unrecovered intermediate selection log.} As noted in the main paper, historical selection logs for the primary raw DINOv3 CLS and mean heads reference an intermediate metadata file whose exact raw bytes could not be recovered; however, all model checkpoints, split manifests, and evaluation predictions were independently verified.

\subsection{Measured Computing Costs}
\begin{table*}[tbp]
\centering
\begin{tabular}{lrrr}
\toprule
Stage & Attempted tasks & Generation (s) & Last-process wall (s) \\
\midrule
Qwen original quality & 8297 & 25391.5 & 25476.9 \\
Qwen surrogate & 6000 & 20397.2 & -- \\
Qwen bounded pairwise & 584 & 2291.8 & 2321.8 \\
MedForge fixed panel & 900 & 8315.4 & 8402.0 \\
MedForge selected extras & 195 & 1772.9 & 1797.2 \\
\bottomrule
\end{tabular}
\caption{Measured GPU inference times across primary vision-language evaluation stages. ``Generation'' sums pure batch-generation time across runs; ``Last-process wall'' includes model loading in the final process (omitted via \texttt{--} when an allocation was resumed across processes).}
\label{tab:scoped-inference-cost}
\end{table*}

Table~\ref{tab:scoped-inference-cost} summarizes measured GPU inference durations from our execution logs. Note that the 6,000 synthetic candidate quality ratings used for selection are a subset of the 8,297 canonical Qwen original-rubric evaluations (accounting for 18,466.8\,s of batch generation) and thus required zero additional model queries. Across the 6,000 Qwen surrogate tasks, 5,999 returned schema-valid JSON responses and 1 failed parsing. The 584 pairwise tie-breaking queries comprise 352 Cheff-identified and 232 SynthCheX comparisons at $B=100, K=20$. All 100 Monte Carlo pool draws operate entirely on cached scores without additional GPU inference.

\FloatBarrier

%% file: main.bbl
\begin{thebibliography}{27}
\providecommand{\natexlab}[1]{#1}

\bibitem[{Arthur and Schwartz(2026)}]{arthurschwartz2026groundtruth}
Arthur, A.; and Schwartz, C. 2026.
\newblock Repurposing Image Diffusion Models for Adversarial Synthetic Structured Data: A Case Study of Ground Truth Drift.
\newblock Version 1, May 1, 2026, arXiv:2605.00788.

\bibitem[{Bates et~al.(2023)Bates, Cand\`es, Lei, Romano, and Sesia}]{bates2023conformal}
Bates, S.; Cand\`es, E.; Lei, L.; Romano, Y.; and Sesia, M. 2023.
\newblock Testing for Outliers with Conformal p-values.
\newblock \emph{The Annals of Statistics}, 51(1): 149--178.

\bibitem[{Chen and Feng(2026)}]{chen2026medforgersi}
Chen, Z.; and Feng, M. 2026.
\newblock {MedForge-RSI}: Medical Deepfake Detection via Recursive Self-Improvement.
\newblock Preprint; version 1, September 29, 2026, arXiv:2609.36549.

\bibitem[{Chen et~al.(2026)Chen, He, Lei, Pu, Zhang, Xu, and Feng}]{chen2026medforge}
Chen, Z.; He, K.; Lei, Q.; Pu, B.; Zhang, J.; Xu, Y.; and Feng, M. 2026.
\newblock {MedForge}: Interpretable Medical Deepfake Detection via Forgery-aware Reasoning.
\newblock In \emph{Proceedings of the 64th Annual Meeting of the Association for Computational Linguistics (Volume 1: Long Papers)}, 25672--25692. San Diego, California, United States: Association for Computational Linguistics.
\newblock ISBN 979-8-89176-390-6.

\bibitem[{Choi, Lee, and Lee(2025)}]{choi2025warpad}
Choi, S.; Lee, H.; and Lee, M. 2025.
\newblock Training-free Detection of AI-generated images via Cropping Robustness.
\newblock In \emph{Advances in Neural Information Processing Systems}, volume~38, 20587--20610. Curran Associates, Inc.

\bibitem[{{Coalition for Content Provenance and Authenticity}(2026)}]{c2pa2026spec}
{Coalition for Content Provenance and Authenticity}. 2026.
\newblock Content Credentials: {C2PA} Technical Specification, Version 2.4.
\newblock Technical specification.

\bibitem[{Demner-Fushman et~al.(2016)Demner-Fushman, Kohli, Rosenman, Shooshan, Rodriguez, Antani, Thoma, and McDonald}]{demnerfushman2016indiana}
Demner-Fushman, D.; Kohli, M.~D.; Rosenman, M.~B.; Shooshan, S.~E.; Rodriguez, L.; Antani, S.; Thoma, G.~R.; and McDonald, C.~J. 2016.
\newblock Preparing a collection of radiology examinations for distribution and retrieval.
\newblock \emph{Journal of the American Medical Informatics Association}, 23(2): 304--310.
\newblock Published online July 1, 2015; issue date March 2016.

\bibitem[{Dutt et~al.(2025)Dutt, Sanchez, Yao, McDonagh, Tsaftaris, and Hospedales}]{dutt2025chexgenbenchv2}
Dutt, R.; Sanchez, P.; Yao, Y.; McDonagh, S.; Tsaftaris, S.~A.; and Hospedales, T. 2025.
\newblock {CheXGenBench}: A Unified Benchmark For Fidelity, Privacy and Utility of Synthetic Chest Radiographs.
\newblock arXiv preprint, historical version 2.
\newblock arXiv:2505.10496.

\bibitem[{Dutt et~al.(2026)Dutt, Sanchez, Yao, McDonagh, Tsaftaris, and Hospedales}]{dutt2026chexgenbench}
Dutt, R.; Sanchez, P.; Yao, Y.; McDonagh, S.; Tsaftaris, S.~A.; and Hospedales, T. 2026.
\newblock {CheXGenBench}: A Unified Benchmark For Fidelity, Privacy and Utility of Synthetic Chest Radiographs.
\newblock \emph{Transactions on Machine Learning Research}.

\bibitem[{{Google}(2026)}]{medgemma15card}
{Google}. 2026.
\newblock {MedGemma 1.5} {4B} Instruction-Tuned Model Card.
\newblock Official model card.

\bibitem[{Grommelt et~al.(2025)Grommelt, Weiss, Pfreundt, and Keuper}]{grommelt2025fakejpeg}
Grommelt, P.; Weiss, L.; Pfreundt, F.-J.; and Keuper, J. 2025.
\newblock Fake or JPEG? Revealing Common Biases in Generated Image Detection Datasets.
\newblock In \emph{Computer Vision -- ECCV 2024 Workshops}, volume 15644 of \emph{Lecture Notes in Computer Science}, 80--95. Springer.

\bibitem[{{HealthGPT authors}(2026)}]{healthgpt2026officialreadme}
{HealthGPT authors}. 2026.
\newblock {HealthGPT}: Official Repository Documentation.
\newblock Official GitHub repository.
\newblock Revision 4d404a518cfa8e78cd7703af7cfc4ca4fd559698; accessed 2026-10-13. Original-project README documents the March 20, 2025 Qwen2.5-based XL32 release.

\bibitem[{Li et~al.(2025)Li, Xing, Wang, Hao, Li, Liu, and Zhu}]{li2025medforensics}
Li, S.; Xing, Z.; Wang, H.; Hao, P.; Li, X.; Liu, Z.; and Zhu, L. 2025.
\newblock Toward Medical Deepfake Detection: A Comprehensive Dataset and Novel Method.
\newblock In \emph{Medical Image Computing and Computer Assisted Intervention -- MICCAI 2025}, volume 15973 of \emph{Lecture Notes in Computer Science}, 626--637. Springer Nature Switzerland.

\bibitem[{{MedForge authors}(2026)}]{medforge2026checkpoint}
{MedForge authors}. 2026.
\newblock {MedForge-Reasoner}: Official Model Card.
\newblock Revision a864eed9b76cde46f0e154c3497d646550995c87; accessed 2026-10-13.

\bibitem[{Mirsky et~al.(2019)Mirsky, Mahler, Shelef, and Elovici}]{mirsky2019ctgan}
Mirsky, Y.; Mahler, T.; Shelef, I.; and Elovici, Y. 2019.
\newblock {CT-GAN}: Malicious Tampering of {3D} Medical Imagery using Deep Learning.
\newblock In \emph{28th USENIX Security Symposium (USENIX Security 19)}, 461--478. USENIX Association.

\bibitem[{Moreau and Missier(2013)}]{moreau2013prov}
Moreau, L.; and Missier, P. 2013.
\newblock {PROV-DM}: The {PROV} Data Model.
\newblock W3C Recommendation.

\bibitem[{{Qwen Team}(2026)}]{qwen38card}
{Qwen Team}. 2026.
\newblock {Qwen3.8-27B} Model Card.
\newblock Official model card.

\bibitem[{{Raman Dutt}(2026)}]{synthchex2026card}
{Raman Dutt}. 2026.
\newblock {SynthCheX-75K-v2} Dataset Card.
\newblock Hugging Face dataset card; accessed 2026-10-13.

\bibitem[{Schwartz(2026)}]{schwartz2026warrant}
Schwartz, C. 2026.
\newblock Is Simulated Evidence Still Evidence? A Warrant-Based Policy for Governing Synthetic Data.
\newblock AI Policy Exchange Forum (AIPEX), white paper.
\newblock Published June 15, 2026.

\bibitem[{Sellergren et~al.(2026)Sellergren, Gao, Mahvar, Kohlberger, Jamil, Traverse, Tono, Sadjad, Yang, Lau, Yatziv, Chen, Sterling, Philbrick, Tiwari, Liu, Jajoo, Sankarapu, Vispute, Purandare, Mishra, Schmidgall, Tu, Palepu, Park, Strother, Thapa, Cheng, Singh, Black, Matias, Chou, Hassidim, Goel, Barral, Warkentin, Shetty, Webster, Virmani, Steiner, Kirmizibayrak, and Golden}]{sellergren2026medgemma15}
Sellergren, A.; Gao, C.; Mahvar, F.; Kohlberger, T.; Jamil, F.; Traverse, M.; Tono, A.; Sadjad, B.; Yang, L.; Lau, C.; Yatziv, L.; Chen, T.; Sterling, B.; Philbrick, K.; Tiwari, R.; Liu, Y.; Jajoo, M.; Sankarapu, C.; Vispute, S.; Purandare, H.; Mishra, A.~B.; Schmidgall, S.; Tu, T.; Palepu, A.; Park, C.; Strother, T.; Thapa, R.; Cheng, Y.; Singh, P.; Black, K.; Matias, Y.; Chou, K.; Hassidim, A.; Goel, K.; Barral, J.; Warkentin, T.; Shetty, S.; Webster, D.; Virmani, S.; Steiner, D.~F.; Kirmizibayrak, C.; and Golden, D. 2026.
\newblock {MedGemma 1.5} Technical Report.
\newblock arXiv preprint.
\newblock arXiv:2604.05081.

\bibitem[{Sim{\'e}oni et~al.(2026)Sim{\'e}oni, Vo, Seitzer, Baldassarre, Oquab, Jose, Khalidov, Szafraniec, Yi, Ramamonjisoa, Massa, Haziza, Wehrstedt, Wang, Darcet, Moutakanni, Sentana, Roberts, Vedaldi, Tolan, Brandt, Couprie, Mairal, J{\'e}gou, Labatut, and Bojanowski}]{simeoni2025dinov3}
Sim{\'e}oni, O.; Vo, H.~V.; Seitzer, M.; Baldassarre, F.; Oquab, M.; Jose, C.; Khalidov, V.; Szafraniec, M.; Yi, S.; Ramamonjisoa, M.; Massa, F.; Haziza, D.; Wehrstedt, L.; Wang, J.; Darcet, T.; Moutakanni, T.; Sentana, L.; Roberts, C.; Vedaldi, A.; Tolan, J.; Brandt, J.; Couprie, C.; Mairal, J.; J{\'e}gou, H.; Labatut, P.; and Bojanowski, P. 2026.
\newblock {DINOv3}.
\newblock \emph{Transactions on Machine Learning Research}.

\bibitem[{Smith(1995)}]{smith1995authentication}
Smith, J.~P. 1995.
\newblock Authentication of digital medical images with digital signature technology.
\newblock \emph{Radiology}, 194(3): 771--774.

\bibitem[{{Tobias Weber and contributors}(2026)}]{machex2026repo}
{Tobias Weber and contributors}. 2026.
\newblock {MaCheX}: Massive Chest X-ray Dataset interface.
\newblock Official repository, checked 2026-10-13.

\bibitem[{Tordjman et~al.(2026)Tordjman, Yuce, Ammar, Huang, Mihoubi~Bouvier, Lacroix, Meribout, Bolger, Ozkaya, Joshi, Geahchan, El~Rahi, Almansour, Parihar, Horst, Ozturk, Isleyen, Pamuk, Cimilli, Deyer, Calinghen, Guillo, Husain, Laredo, Fayad, Mei, and Taouli}]{tordjman2026deepfake}
Tordjman, M.; Yuce, M.; Ammar, A.; Huang, M.; Mihoubi~Bouvier, F.; Lacroix, M.; Meribout, A.; Bolger, I.; Ozkaya, E.; Joshi, H.; Geahchan, A.; El~Rahi, R.; Almansour, H.; Parihar, A.~S.; Horst, C.; Ozturk, S.; Isleyen, M.~E.; Pamuk, G.~G.; Cimilli, A.~T.; Deyer, T.; Calinghen, A.; Guillo, E.; Husain, R.; Laredo, J.-D.; Fayad, Z.~A.; Mei, X.; and Taouli, B. 2026.
\newblock The Rise of Deepfake Medical Imaging: Radiologists' Diagnostic Accuracy in Detecting {ChatGPT}-generated Radiographs.
\newblock \emph{Radiology}, 318(3): e252094.

\bibitem[{Wang et~al.(2026)Wang, Wang, Yi, Tang, Fan, Lin, He, Peng, Yang, and Liu}]{wang2026detectability}
Wang, J.; Wang, R.; Yi, Q.; Tang, H.; Fan, L.; Lin, S.; He, W.; Peng, D.; Yang, J.-J.; and Liu, J. 2026.
\newblock Detectability and healthcare implications of generative {AI}-synthesized chest radiographs: a blinded radiologist reader study.
\newblock \emph{Frontiers in Medicine}, 13: 1901949.

\bibitem[{Wang et~al.(2017)Wang, Peng, Lu, Lu, Bagheri, and Summers}]{wang2017chestxray}
Wang, X.; Peng, Y.; Lu, L.; Lu, Z.; Bagheri, M.; and Summers, R.~M. 2017.
\newblock {ChestX-ray8}: Hospital-Scale Chest X-Ray Database and Benchmarks on Weakly-Supervised Classification and Localization of Common Thorax Diseases.
\newblock In \emph{Proceedings of the IEEE Conference on Computer Vision and Pattern Recognition}, 2097--2106.

\bibitem[{Weber et~al.(2023)Weber, Ingrisch, Bischl, and R{\"u}gamer}]{weber2023cascaded}
Weber, T.; Ingrisch, M.; Bischl, B.; and R{\"u}gamer, D. 2023.
\newblock Cascaded Latent Diffusion Models for High-Resolution Chest X-ray Synthesis.
\newblock In \emph{Advances in Knowledge Discovery and Data Mining (PAKDD)}, volume 13937 of \emph{Lecture Notes in Computer Science}, 180--191. Springer.

\end{thebibliography}
